\documentclass[10pt,twocolumn,letterpaper]{article}

\usepackage[pagenumbers]{cvpr} % To force page numbers, e.g. for an arXiv version

\usepackage{colortbl}
\usepackage{xcolor}
\usepackage{multirow}
\usepackage{booktabs}
\usepackage{arydshln}
\usepackage{tikz}
\usepackage{pifont}
\usepackage{float}
\newcommand{\greencheck}{{\color{green}\checkmark}}
\newcommand{\redx}{{\color{red}\ding{55}}}
\newcommand{\higher}{$\uparrow$}
\newcommand{\perflower}{$\downarrow$}

\makeatother

\definecolor{Gray}{gray}{0.50}
\newcolumntype{g}{>{\columncolor{Gray}}c}
\definecolor{ffe1da}{RGB}{255,225,218}
\definecolor{F7E0D5}{RGB}{247,224,213}
\definecolor{darkF7E0D5}{RGB}{209,154,128}
\colorlet{Light}{White!0!F7E0D5}

\colorlet{tabfirst}{Green!25}
\definecolor{tabthird}{rgb}{1, 0.85, 0.7}
\definecolor{tabsecond}{rgb}{1, 0.96, 0.7}

\definecolor{cvprblue}{rgb}{0.21,0.49,0.74}

\usepackage[pagebackref,breaklinks,colorlinks,allcolors=cvprblue]{hyperref}

\def\paperID{17566} % *** Enter the Paper ID here
\def\confName{CVPR}
\def\confYear{2025}

\title{FlashSLAM: Accelerated RGB-D SLAM for Real-Time 3D Scene Reconstruction with Gaussian Splatting \\[1pt]
\Large \href{https://flashslam.github.io}{flashslam.github.io}
}

\author{Phu Pham$^{1}$, Damon Conover$^{2}$,  Aniket Bera$^{1}$\\
$^1$Department of Computer Science, Purdue University  $^2$DEVCOM Army Research Laboratory\\
\texttt{\{phupham, aniketbera\}@purdue.edu, damon.m.conover.civ@army.mil}
}

\begin{document}
\maketitle
\begin{abstract}

We present FlashSLAM, a novel SLAM approach that leverages 3D Gaussian Splatting for efficient and robust 3D scene reconstruction. Existing 3DGS-based SLAM methods often fall short in sparse view settings and during large camera movements due to their reliance on gradient descent-based optimization, which is both slow and inaccurate. FlashSLAM addresses these limitations by combining 3DGS with a fast vision-based camera tracking technique, utilizing a pretrained feature matching model and point cloud registration for precise pose estimation in under 80 ms - a 90\% reduction in tracking time compared to SplaTAM - without costly iterative rendering. In sparse settings, our method achieves up to a 92\% improvement in average tracking accuracy over previous methods. Additionally, it accounts for noise in depth sensors, enhancing robustness when using unspecialized devices such as smartphones. Extensive experiments show that FlashSLAM performs reliably across both sparse and dense settings, in synthetic and real-world environments. Evaluations on benchmark datasets highlight its superior accuracy and efficiency, establishing FlashSLAM as a versatile and high-performance solution for SLAM, advancing the state-of-the-art in 3D reconstruction across diverse applications.

\end{abstract}    
\section{Introduction}
\label{sec:intro}

Building accurate 3D maps while simultaneously tracking position has become foundational in advancing autonomous systems across robotics, computer vision, and augmented reality \cite{slam-p1, slam-driving, slam-ar}, leading to the rapid evolution of Simultaneous Localization and Mapping (SLAM) techniques. Recent advancements in SLAM have focused on developing more accurate, efficient, and robust methods for real-time 3D reconstruction and camera tracking.

Integrating neural rendering techniques with SLAM systems has shown promise for joint localization and photorealistic view reconstruction \cite{nice-slam, point-slam, imap, nerf-slam}. However, many of these approaches depend heavily on implicit representations, which are computationally intensive and impractical for portable device deployment \cite{Photo-SLAM}. This constraint has driven research toward more efficient representations that preserve high-quality reconstructions while supporting real-time performance.

One such representation that has gained significant attention is 3D Gaussian Splatting (3DGS) \cite{3dgs}. Originally developed for offline scene reconstruction, 3DGS has recently been adapted for real-time SLAM applications, driving a wave of innovative research in the field. Several pioneering works \cite{monogs, gsslam, splatam} have demonstrated the potential of 3DGS in unifying tracking, mapping, and rendering within a single framework, capable of handling both monocular and RGB-D inputs effectively. These methods have shown significant progress in addressing key challenges in SLAM, such as real-time performance, high-quality reconstruction, and robustness to various input modalities.

Despite these advancements, existing 3DGS-based SLAM methods still face challenges in sparse settings or scenarios with rapid camera movement. Current approaches often rely on gradient descent optimization for localization, which can lead to slow convergence and inaccurate results when dealing with large camera displacements.

Our work addresses these limitations by introducing a novel SLAM approach that leverages 3D Gaussian splatting for robust 3D scene reconstruction while incorporating an efficient and accurate vision-based camera tracking method. Unlike existing approaches that optimize camera pose using gradient descent and require rendering for each parameter update, our approach utilizes a pretrained local feature-matching model and point cloud registration. This results in faster and more accurate camera estimation, particularly in sparse settings or when dealing with large camera movements.

The main contributions of our work are as follows:

\begin{itemize}
    \item \textbf{Efficient Vision-based Camera Tracking}: A fast and accurate camera pose estimation method using pretrained feature matching and point cloud registration, achieving significant speed and accuracy gains, especially in sparse environments and with large camera movements.

    \item \textbf{High-Fidelity Reconstruction}: An optimized 3DGS representation utilizing RGB-D images to produce high-quality 3D reconstructions in real time.

    \item \textbf{Robustness to Depth Sensor Errors}: A truncated depth approach mitigates errors from consumer-grade depth sensors, enabling effective use on devices like smartphones.

    \item \textbf{Comprehensive Evaluation}: Extensive experiments on benchmark datasets highlight our approach’s accuracy, efficiency, and robustness.
\end{itemize}

These contributions advance SLAM and 3D reconstruction, offering a robust, efficient, and adaptable solution for diverse applications and hardware. The following sections detail our methodology, experimental setup, and results, demonstrating our approach's effectiveness across multiple benchmarks.

\section{Related Work}

\subsection{SLAM Systems}

Simultaneous Localization and Mapping (SLAM) has been a central challenge in robotics and computer vision for decades, evolving from traditional sparse feature-based methods to dense volumetric representations. Early approaches like ORB-SLAM \cite{orb-slam} utilized distinctive visual features for mapping and localization, while later techniques such as KinectFusion \cite{kineticfusion} employed depth sensors for richer geometric reconstruction. These methods, however, faced limitations in certain environments and often required specialized hardware.

The recent introduction of neural implicit representations, particularly Neural Radiance Fields (NeRF) \cite{nerf}, has opened new avenues for SLAM research. Systems like iMAP \cite{imap} and NICE-SLAM \cite{nice-slam} have demonstrated the potential of these representations for high-quality scene reconstruction and localization. While promising, these neural approaches face challenges in computational efficiency, scalability, and real-time performance, especially in large-scale or dynamic environments. Current research focuses on addressing these limitations through hybrid approaches, more efficient neural architectures, and multi-modal sensor fusion, aiming to create more robust and versatile SLAM systems for real-world applications \cite{point-slam}.

\subsection{3D Gaussian Splatting}

3D Gaussian Splatting (3DGS), introduced by Kerbl et al.~\cite{3dgs}, is a robust technique for representing and rendering 3D scenes using collections of 3D Gaussian primitives to model both geometry and appearance. Compared to traditional mesh-based methods \cite{PoissonRecon, MarchingCube}, 3DGS offers a more flexible representation that enables smoother surface approximations and better management of partial or noisy data. These properties make it especially suitable for real-time applications like SLAM, where data is often incomplete.

Recent advancements in 3DGS have further expanded its utility, supporting efficient integration of multimodal data such as RGB and depth, which enhances reconstruction accuracy. This method has shown significant improvements in rendering speed and quality over earlier approaches like Neural Radiance Fields (NeRF) \cite{nerf}, establishing it as a promising direction for research in computer graphics and computer vision \cite{3dgs}.

\subsection{3DGS-based SLAM}

The integration of 3D Gaussian Splatting (3DGS) into SLAM systems marks a significant advancement, utilizing 3DGS for real-time mapping and localization. Matsuki et al.~\cite{monogs} introduced a monocular Gaussian Splatting SLAM system that operates live at 3 fps, using Gaussians as the sole 3D representation for efficient tracking, mapping, and rendering.

Similarly, GS-SLAM by Yan et al.~\cite{gsslam} proposed a dense visual SLAM system based on 3DGS, offering competitive reconstruction and localization performance at reduced computational costs. Through a real-time differentiable splatting rendering pipeline, GS-SLAM balances accuracy and efficiency, accelerating map optimization and RGB-D rendering.

SplaTAM by Keetha et al.~\cite{splatam} further advanced 3DGS-based SLAM by introducing a dense RGB-D system that achieves real-time performance and high-quality reconstruction, showcasing the potential of 3DGS in SLAM, especially with depth data.

Despite their strengths, these systems face challenges in sparse or dynamic real-world scenes. All current 3DGS-based SLAM methods estimate camera pose by projecting the previous pose forward using a constant velocity assumption in camera center + quaternion space. This approach struggles with large camera movements, leading to inaccurate poses and cumulative errors that cause \textit{camera drifting}, compromising reconstruction accuracy.

Techniques like loop closure detection and bundle adjustment (BA) could refine global poses but often fail in feature-sparse environments or under lighting, viewpoint, or dynamic changes \cite{BA}. Without a loop in the camera path, loop closure cannot mitigate drift. Additionally, rapid camera movements, repetitive patterns, and limited computational resources exacerbate drift, challenging SLAM systems to maintain real-time precision in large or complex environments.
\section{Method}

\begin{figure*}
    \centering
    \includegraphics[width=0.9\linewidth]{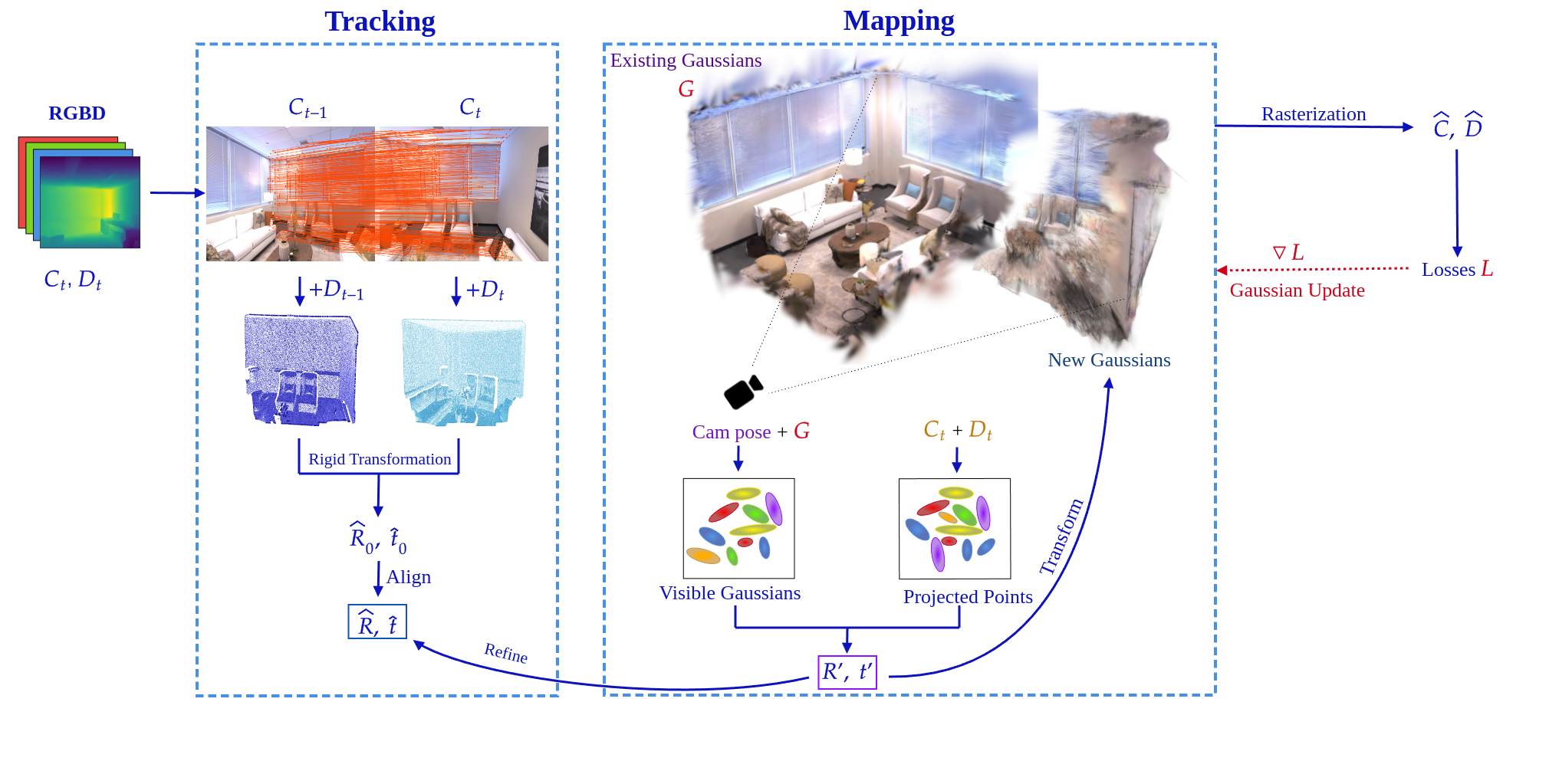}
    \caption{\textbf{Overview of FlashSLAM}: Our approach takes RGB-D inputs to perform accurate 3D scene reconstruction. Initially, precise matches between consecutive frames are detected, which enables tracking of the camera pose through a rigid transformation. This pose is further refined using gradient-based optimization, leveraging Gaussian alignment to ensure accurate registration of new frames with the existing 3D model. The mapping process updates and transforms existing Gaussian splats in the 3D scene, producing high-quality reconstructions with efficient alignment and optimization steps.}
    \label{fig:overview}
\end{figure*}

The core of our method is an efficient tracking system designed to perform accurately in both dense and sparse settings. In this section, we describe our approach in detail. Figure \ref{fig:overview} provides an overview, highlighting how our method integrates precise feature matching, reliable pose estimation, and refinement techniques to achieve high-accuracy tracking. This enables robust SLAM performance across varying levels of data sparsity and environmental complexity.

\subsection{3D Gaussian representation}

Following recent advances, we adopt the 3D Gaussian splatting (3DGS) approach to represent the scene as a continuous field of Gaussian splats. Each splat $G_i$ has a 3D center $\mathbf{c}_i$, covariance $\Sigma_i$, color $\mathbf{v}_i$, and opacity $\alpha_i$. The Gaussian function $G_i(\mathbf{x})$ for a point $\mathbf{x}$ in space is defined as:

\begin{equation}
G_i(\mathbf{x}) = \alpha_i \exp \left( -\frac{1}{2} (\mathbf{x} - \mathbf{c}_i)^T \Sigma_i^{-1} (\mathbf{x} - \mathbf{c}_i) \right),
\end{equation}

where $\Sigma_i$ controls the size and orientation, and $\alpha_i$ modulates opacity.

To render, each splat’s color $\mathbf{v}_i$ is scaled by its intensity and opacity, contributing to the final image. We project each 3D center $\mathbf{c}_i$ to the 2D image plane using the camera intrinsics $K$ and extrinsics $[R | t]$, yielding:

\begin{equation}
\mathbf{p}_i = K [R | t] \mathbf{c}_i,
\end{equation}

with projected covariance $\Sigma_i^{\text{proj}} = J_i \Sigma_i J_i^T$ for adapting the splat’s shape in the image plane.

This differentiable rendering process allows us to optimize Gaussian parameters by comparing rendered images from multiple viewpoints with ground-truth images, refining the splats for accurate scene representation. We use this 3DGS representation in our SLAM framework to achieve high-fidelity, adaptable 3D reconstructions without relying on explicit meshes.

\subsection{Camera Pose Tracking}

Recent Gaussian-based SLAM methods, such as Splatam \cite{monogs, splatam, gsslam}, rely on an assumption of constant velocity and small discrepancies in camera movement, optimizing the camera position based on photometric and depth losses between rendered and ground-truth images. While effective for dense SLAM, this approach struggles in sparse settings where large camera movements and limited frame overlap challenge the model's ability to accurately estimate pose. In these scenarios, the assumption of gradual motion breaks down, leading to failure in aligning frames with significant disparity.

Classic approaches for estimating rigid body transformations, such as the least-squares fitting method for 3D point sets \cite{least-squares-arun, rigid-horn, rigid-eggert}, provide a foundational method for determining camera transformations by aligning corresponding points between frames. In our setup, we utilize LightGlue \cite{lightglue}, a robust feature-matching framework designed to produce highly reliable correspondences, even under challenging conditions. LightGlue generates high-confidence matches between frames $I_{t-1}$ and $I_{t}$, forming a strong basis for accurate transformation estimation, especially in scenarios with significant disparity between frames.

To utilize these correspondences, we first extract keypoints and descriptors from the RGB frames $I_{t-1}$ and $I_{t}$ using SuperPoint \cite{superpoint}, with LightGlue yielding high-quality matched points $(\mathbf{p}_{t-1}, \mathbf{p}_{t})$. After filtering for reliable matches, we back-project these 2D points to 3D space using the camera’s intrinsic parameters, resulting in two point sets, $\mathbf{P}_{t-1}$ and $\mathbf{P}_{t}$, representing frames $I_{t-1}$ and $I_{t}$. We then estimate the optimal rigid transformation, consisting of a rotation matrix $\mathbf{R} \in \text{SO}(3)$ and translation vector $\mathbf{t} \in \mathbb{R}^3$, by minimizing the least-squares error:

\begin{equation}
\min_{\mathbf{R}, \mathbf{t}} \sum_{i=1}^{N} \|\mathbf{p}_{t}^{(i)} - (\mathbf{R} \mathbf{p}_{t-1}^{(i)} + \mathbf{t}) \|^2.
\end{equation}

To estimate the rotation matrix $\mathbf{R}$ and translation vector $\mathbf{t}$, we first calculate the centroids of each point set:

\begin{equation}
\mathbf{c}_{t-1} = \frac{1}{N} \sum_{i=1}^{N} \mathbf{p}_{t-1}^{(i)}, \quad \mathbf{c}_{t} = \frac{1}{N} \sum_{i=1}^{N} \mathbf{p}_{t}^{(i)}.
\end{equation}

Centering each point set around its centroid, we obtain $\mathbf{Q}_{t-1} = \mathbf{P}_{t-1} - \mathbf{c}_{t-1}$ and $\mathbf{Q}_{t} = \mathbf{P}_{t} - \mathbf{c}_{t}$, and compute the covariance matrix $\mathbf{H} = \mathbf{Q}_{t-1}^T \mathbf{Q}_{t}$. 

The optimal rotation matrix $\mathbf{R}$ is given by:

\begin{equation}
\mathbf{R} = \mathbf{V} \mathbf{U}^T, 
\end{equation}

where $\mathbf{U}$ and $\mathbf{V}$ are obtained from the Singular Value Decomposition (SVD) of $\mathbf{H}$, such that $\mathbf{H} = \mathbf{U} \mathbf{\Sigma} \mathbf{V}^T$. If $\det(\mathbf{R}) < 0$, we adjust by negating the last column of $\mathbf{V}$ to ensure $\mathbf{R}$ is a valid rotation. The translation vector is computed as:

\begin{equation}
\mathbf{t} = \mathbf{c}_{t} - \mathbf{R} \mathbf{c}_{t-1}.
\end{equation}

The rigid transformation provides a good initial estimate of the camera pose; however, it can be further refined through gradient-based optimization with significantly fewer iterations. Although the transformation is generally accurate when depth values are reliable, performance can degrade in the presence of depth sensor noise, which may distort the transformation. To mitigate this, we apply a dynamically adjusted depth truncation threshold, set at the 70th percentile of the depth distribution for each frame, to filter out excessively noisy points. This approach avoids setting the threshold too low, thereby retaining valid matches while excluding unreliable depths. By ensuring that only reliable depth values contribute to the final transformation, we improve robustness in challenging conditions.

\subsection{Mapping}

\paragraph{Adding New Gaussians.} Our mapping step dynamically updates the 3D Gaussian map to maintain an accurate representation of the scene. At each frame, we render the scene using the current estimated camera pose and compare the rendered depth values, $D_r$, with the ground-truth depth image, $D_{gt}$. For each pixel $(u, v)$ in $D_{gt}$ where $D_{gt}(u, v) < D_r(u, v)$, we identify the need for additional Gaussians to represent that region of the scene, as existing Gaussians fail to capture closer surfaces.

After identifying regions that require new Gaussians, we project the pixels of the new frame into 3D space using depth values and the camera’s intrinsic parameters. Let $P_o$ denote the set of projected points that have already been mapped and are represented by existing Gaussians, and let $P_n$ represent the set of points corresponding to the newly identified regions that need additional Gaussians. Due to potential noise in the depth sensor, the projected positions in $P_n$ may be inaccurate. To address this, we apply a transformation correction process to align $P_o$ with the current Gaussians before integrating $P_n$.

To align $P_o$ with the existing Gaussians, we first select a subset $G_v \subset G$ of Gaussians that are visible from the current camera position, where $G$ represents all Gaussians in the map. The visibility of a Gaussian $g \in G$ is determined by checking if its center falls within the frustum of the current camera view. We then downsample both $G_v$ and $P_o$ to reduce computational complexity, before applying Iterative Closest Point (ICP) to find the optimal rotation $\mathbf{R}$ and translation $\mathbf{t}$. The transformation $(\mathbf{R}, \mathbf{t})$ is accepted if the ICP alignment achieves a fitness score $f > 0.2$ and an error $e < 0.1$.
If these conditions are not met, the transformation is rejected to prevent the introduction of inaccuracies.

Once a valid transformation $(\mathbf{R}, \mathbf{t})$ is obtained, we apply it to the points in $P_n$ to correct their positions:

\begin{equation}
\mathbf{p}_{n}^{(i)} \leftarrow \mathbf{R} \mathbf{p}_{n}^{(i)} + \mathbf{t},
\end{equation}

where $\mathbf{p}_{n}^{(i)} \in P_n$. The transformed points in $P_n$ are then used as the centers for new Gaussians to be added to the map.

Additionally, we adjust the current estimated camera pose by incorporating the computed transformation, ensuring that the newly added Gaussians are seamlessly integrated with the existing map. This alignment process helps maintain coherence between new and existing Gaussians, bolstering the robustness of subsequent optimizations.

\paragraph{Loss Functions.} To optimize the Gaussian map, we adopt a strategy similar to recent RGB-D SLAM methods by combining photometric and depth losses to ensure both color fidelity and geometric accuracy in the reconstructed scene.

The photometric loss function, $L_{\text{color}}$, measures the $L_1$ difference between the ground-truth image $C$ and the reconstructed image $\hat{C}$ from the same estimated camera pose:

\begin{equation}
L_{\text{color}} = \| C - \hat{C} \|_1.
\end{equation}

In addition to the photometric loss, we apply a depth loss, $L_{\text{depth}}$, to ensure geometric consistency in the scene reconstruction. The depth loss is defined as the $L_1$ distance between the ground-truth depth map $D$ and the reconstructed depth map $\hat{D}$:

\begin{equation}
L_{\text{depth}} = \| D - \hat{D} \|_1,
\end{equation}

where $D$ and $\hat{D}$ represent the depth values for each pixel in the ground-truth and reconstructed images, respectively.

The overall loss function, $L$, combines the photometric and depth losses with a weighting factor $\lambda$ that controls the relative importance of each term:

\begin{equation}
L = \lambda L_{\text{color}} + (1 - \lambda) L_{\text{depth}}.
\end{equation}

The parameter $\lambda$ allows us to balance color and depth consistency according to the specific requirements of the scene reconstruction.

\paragraph{Keyframe Selection.} In dense SLAM settings, where camera movement is minimal, many frames have substantial overlap with their neighboring frames, leading to redundancy. Optimizing the Gaussian map using all these frames would be computationally expensive and unnecessary. To address this, we employ a simple keyframe selection strategy, as proposed in prior work \cite{monogs}. A new frame is added to the keyframe list if it overlaps with the previous keyframe and falls below a specified threshold. This overlap is measured as the intersection over union (IoU) of the visible Gaussians within the camera frustums of the two frames. In sparse settings, we simplify the selection by designating every $k^{th}$ frame as a keyframe.

\subsection{Color Refinement with Priority Sampling}

To enhance the appearance of the reconstructed scene, we employ a refinement step to improve the alignment of Gaussian positions and colors. Unlike initial optimization, this step does not add or remove Gaussians; rather, it adjusts the existing Gaussians for more precise alignment, making the refinement computationally efficient as the number of Gaussians remains constant.

For this process, we use a list of keyframes and apply a priority sampling strategy, loss-weighted sampling, where frames with higher losses are more likely to be sampled for refinement. The probability of sampling a frame $i$ is proportional to its loss magnitude, $L_i$. Specifically, we compute the probability $P(i)$ of selecting frame $i$ as follows:

\begin{equation}
P(i) = \frac{L_i}{\sum_{j=1}^N L_j},
\end{equation}

where $N$ is the total number of keyframes, and $L_j$ represents the loss for each frame $j$. This ensures that frames with greater discrepancies (higher $L_i$) are sampled more frequently, directing optimization efforts toward frames where alignment is weakest.

We also experimented with an alternative strategy, worst-first sampling, which samples only the frame with the highest loss by maintaining a heap of frame losses. However, we found that loss-weighted sampling combined with random sampling yielded better results. Specifically, to avoid `forgetting" well-aligned frames, we apply priority sampling with a probability $p = 0.4$ and use random sampling with probability $1 - p = 0.6$, selecting frames uniformly from the keyframe list. This balance ensures that high-error frames are refined more often while still maintaining optimization across well-aligned frames, preserving global consistency in the reconstructed scene.
\section{Experiments and Results}

\subsection{Experiment Setup}

\paragraph{Datasets.} For performance evaluation, we use two widely recognized datasets in 3D computer vision and robotics: Replica \cite{replica} and TUM RGB-D \cite{tum-rgbd}. The Replica dataset consists of synthetic 3D indoor scenes with dense geometry and high-resolution HDR textures. The TUM RGB-D dataset provides synchronized RGB and depth images from a handheld camera, suitable for real-world indoor environments. For comparison with existing methods, we evaluate on a subset of each dataset, specifically using 8 scenes from Replica \cite{replica} and all 5 scenes from TUM RGB-D \cite{tum-rgbd}.

\paragraph{Evaluation Metrics.} To evaluate rendering quality, we employ three metrics: Peak Signal-to-Noise Ratio (PSNR), Structural Similarity Index (SSIM) \cite{ssim}, and Learned Perceptual Image Patch Similarity (LPIPS) \cite{lpips}. These metrics are computed by comparing the rendered image frames against the corresponding sensor images. For assessing camera tracking precision, we utilize the Root Mean Square Error (RMSE) of the Absolute Trajectory Error (ATE) calculated on the keyframes.

\paragraph{Baseline Methods.} We evaluate our SLAM method in comparison to state-of-the-art SLAM approaches based on NeRF and 3D Gaussian Splatting. For rendering quality, we compare against Point-SLAM \cite{point-slam}, MonoGS \cite{monogs}, SplatAM \cite{splatam}, and GS-SLAM \cite{gsslam}. Furthermore, we assess our tracking performance relative to iMAP \cite{imap}, NICE-SLAM \cite{nice-slam}, and ESSLAM \cite{eslam} across various datasets. In this work, we place greater emphasis on Gaussian-based methods, as NeRF-based approaches have been extensively studied in recent Gaussian-based SLAM methods. It has been demonstrated that Gaussian-based approaches offer significant advantages in terms of both rendering quality and speed.

\paragraph{Implementation Details.} We implement our method using PyTorch and evaluate it on a single NVIDIA A100 GPU with 80GB of memory. Further details will be provided in the supplementary material.

\subsection{Tracking and Mapping Accuracies}

\begin{table}[!htp]
\centering
\scriptsize
\setlength{\tabcolsep}{1.2pt}
\begin{tabular}{@{}l l c c c c c c c c c c@{}}
\toprule
\textbf{Methods} & \textbf{Metrics} & \textbf{r0} & \textbf{r1} & \textbf{r2} & \textbf{o0} & \textbf{o1} & \textbf{o2} & \textbf{o3} & \textbf{o4} & \textbf{Avg.} & \textbf{FPS} \\
\midrule

\multirow{3}{*}{\parbox{1.3cm}{Point-SLAM \\ \cite{point-slam}}}
& PSNR \higher & 32.04 & \cellcolor{tabthird}34.08 & \cellcolor{tabthird}35.50 & \cellcolor{tabthird}38.26 & 39.16 & \cellcolor{tabthird}33.99 & \cellcolor{tabthird}33.48 & \cellcolor{tabthird}33.49 & \cellcolor{tabthird}35.00 & \multirow{3}{*}{1.33}\\ 
& SSIM \higher & \cellcolor{tabsecond}0.974 & \cellcolor{tabfirst}\textbf{0.977} & \cellcolor{tabfirst}\textbf{0.982} & \cellcolor{tabthird}0.983 & \cellcolor{tabsecond}0.986 & 0.960 & \cellcolor{tabthird}0.960 & \cellcolor{tabfirst}\textbf{0.979} & \cellcolor{tabsecond}0.975 \\ 
& LPIPS \perflower & 0.113 & 0.116 & 0.111 & 0.100 & 0.118 & 0.156 & 0.132 & 0.142 & 0.124 \\ 

% \cdashmidrule{1-11}
\cmidrule{1-12}

% \multirow{3}{*}{GS-SLAM~\cite{gsslam}} 
\multirow{3}{*}{\parbox{1.3cm}{GS-SLAM \\ \cite{gsslam}}}
& PSNR \higher & 31.56 & 32.86 & 32.59 & 38.70 & \cellcolor{tabthird}41.17 & 32.36 & 32.03 & 32.92 & 34.27 & \multirow{3}{*}{387}\\ 
& SSIM \higher & 0.968 & \cellcolor{tabthird}0.973 & 0.971 & \cellcolor{tabfirst}\textbf{0.986} & \cellcolor{tabfirst}\textbf{0.993} & \cellcolor{tabfirst}\textbf{0.978} & \cellcolor{tabsecond}0.970 & \cellcolor{tabthird}0.968 & \cellcolor{tabfirst}\textbf{0.976} \\ 
& LPIPS \perflower & 0.094 & \cellcolor{tabsecond}0.075 & 0.093 & \cellcolor{tabsecond}0.050 & \cellcolor{tabsecond}0.033 & \cellcolor{tabthird}0.094 & \cellcolor{tabthird}0.110 & \cellcolor{tabthird}0.112 & \cellcolor{tabthird}0.083 \\ 

% \cdashmidrule{1-12}
\cmidrule{1-12}

% \multirow{3}{*}{SplaTAM~\cite{splatam}} 
\multirow{3}{*}{\parbox{1.3cm}{SplaTAM \\ \cite{splatam}}}
& PSNR \higher & \cellcolor{tabthird}32.86 & 33.89 & 35.25 & \cellcolor{tabthird}38.26 & 39.17 & 31.97 & 29.70 & 31.81 & 34.11 & \multirow{3}{*}{400} \\ 
& SSIM \higher & \cellcolor{tabfirst}\textbf{0.980} & 0.970 & \cellcolor{tabsecond}0.980 & 0.980 & \cellcolor{tabthird}0.980 & \cellcolor{tabthird}0.970 & 0.950 & 0.950 & \cellcolor{tabthird}0.970 \\ 
& LPIPS \perflower & \cellcolor{tabthird}0.070 & 0.100 & \cellcolor{tabthird}0.080 & 0.090 & 0.090 & 0.100 & 0.120 & 0.150 & 0.100 \\ 

% \cdashmidrule{1-12}
\cmidrule{1-12}

% \multirow{3}{*}{MonoGS~\cite{monogs}}
\multirow{3}{*}{\parbox{1.3cm}{MonoGS \\ \cite{monogs}}}
& PSNR \higher & \cellcolor{tabsecond}34.83 & \cellcolor{tabsecond}36.43 & \cellcolor{tabsecond}37.49 & \cellcolor{tabsecond}39.95 & \cellcolor{tabsecond}42.09 & \cellcolor{tabsecond}36.24 & \cellcolor{tabsecond}36.70 & \cellcolor{tabfirst}\textbf{37.50} & \cellcolor{tabsecond}37.65 & \multirow{3}{*}{769}\\ 
& SSIM \higher & 0.954 & 0.959 & 0.965 & \cellcolor{tabthird}0.971 & 0.977 & 0.964 & 0.963 & 0.963 & 0.964 \\ 
& LPIPS \perflower & \cellcolor{tabsecond}0.068 & \cellcolor{tabthird}0.076 & \cellcolor{tabsecond}0.075 & \cellcolor{tabthird}0.072 & \cellcolor{tabthird}0.055 & \cellcolor{tabsecond}0.078 & \cellcolor{tabsecond}0.065 & \cellcolor{tabsecond}0.070 & \cellcolor{tabsecond}0.070 \\ 

% \cdashmidrule{1-12}
\cmidrule{1-12}

\multirow{3}{*}{\textbf{Ours}}
& PSNR \higher & \cellcolor{tabfirst}\textbf{36.78} & \cellcolor{tabfirst}\textbf{38.85} & \cellcolor{tabfirst}\textbf{38.45} & \cellcolor{tabfirst}\textbf{43.32} &  \cellcolor{tabfirst}\textbf{43.73} & \cellcolor{tabfirst}\textbf{37.39} & \cellcolor{tabfirst}\textbf{37.56} & \cellcolor{tabsecond}37.45 & \cellcolor{tabfirst}\textbf{39.21} & \multirow{3}{*}{\textbf{899}}\\ 
& SSIM \higher & \cellcolor{tabthird}0.970 & \cellcolor{tabsecond}0.974 & \cellcolor{tabthird}0.976 & \cellcolor{tabsecond}0.985 & \cellcolor{tabsecond}0.986 & \cellcolor{tabsecond}0.976 & \cellcolor{tabfirst}\textbf{0.973} & \cellcolor{tabsecond}0.970 & \cellcolor{tabfirst}\textbf{0.976} \\ 
& LPIPS \perflower & \cellcolor{tabfirst}\textbf{0.042} & \cellcolor{tabsecond}\cellcolor{tabfirst}\textbf{0.045} & \cellcolor{tabfirst}\textbf{0.057} & \cellcolor{tabfirst}\textbf{0.028} & \cellcolor{tabfirst}\textbf{0.030} & \cellcolor{tabfirst}\textbf{0.040} & \cellcolor{tabfirst}\textbf{0.038} & \cellcolor{tabfirst}\textbf{0.054} & \cellcolor{tabfirst}\textbf{0.042} \\ 

\bottomrule
\end{tabular}
\caption{Comparison of different methods based on PSNR, SSIM, and LPIPS metrics across various scenes. For each scene, the top three performances for each metric are highlighted as \colorbox{tabfirst}{\textbf{first}}, \colorbox{tabsecond}{second}, and \colorbox{tabthird}{third}.}
\label{table:render_replica}
\end{table}

\paragraph{Rendering quality}

Table \ref{table:render_replica} presents a quantitative comparison of our approach on the Replica dataset \cite{replica} against four baseline methods: Point-SLAM \cite{point-slam}, GS-SLAM \cite{gsslam}, SplaTAM \cite{splatam}, and MonoGS \cite{monogs}. The performance metrics—PSNR, SSIM, and LPIPS—are evaluated across eight scenes (\texttt{room 0-2} and \texttt{office 0-4}). For each metric, the top three results per scene are highlighted, with the highest score marked as \colorbox{tabfirst}{\textbf{first}}, \colorbox{tabsecond}{second}, and \colorbox{tabthird}{third}. Results from other methods are sourced directly from their original papers.

Our method consistently surpasses the baselines, achieving the highest average scores across all metrics. Specifically, we achieve an average PSNR of \textbf{38.45}, SSIM of \textbf{0.978}, and LPIPS of \textbf{0.042}, showing notable improvements over the other methods. On a per-scene basis, our method ranks first for all but one PSNR score in the \texttt{office 4} scene. For SSIM, our approach also achieves top scores in most scenes, indicating superior structural similarity in reconstructions compared to baselines. In terms of LPIPS, which reflects perceptual quality, our method consistently produces the lowest scores across all scenes.

Moreover, our approach achieves the highest frame rate (FPS) of \textbf{899}, substantially outperforming the competing methods. This high FPS highlights the computational efficiency of our approach, making it well-suited for real-time applications.

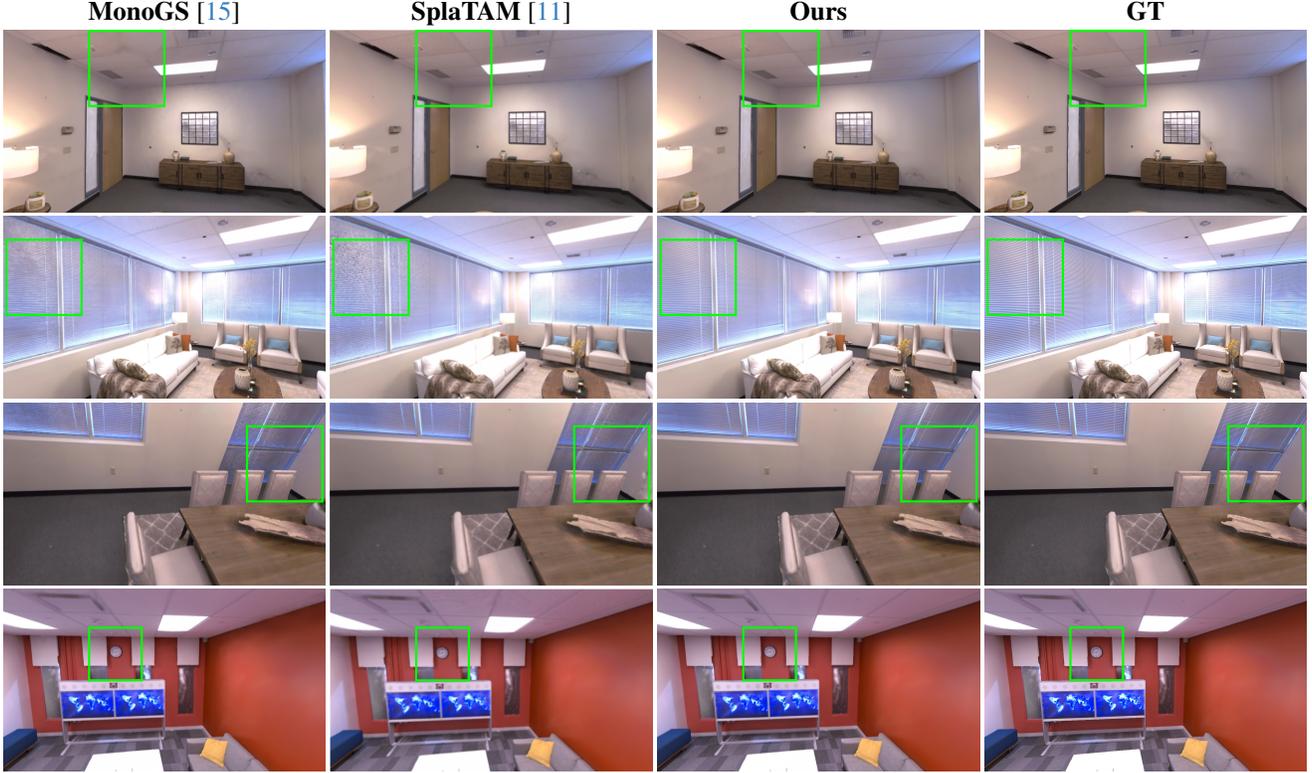
\begin{figure*}[h]
\centering
\setlength{\tabcolsep}{1pt}
\begin{tabular}{c c c c}

\textbf{MonoGS} \cite{monogs} & \textbf{SplaTAM} \cite{splatam} & \textbf{Ours} & \textbf{GT} \\

% Row 1
\begin{tikzpicture}
    \node[inner sep=0pt] (image) at (0.5,-0.7) {\includegraphics[width=0.245\textwidth]{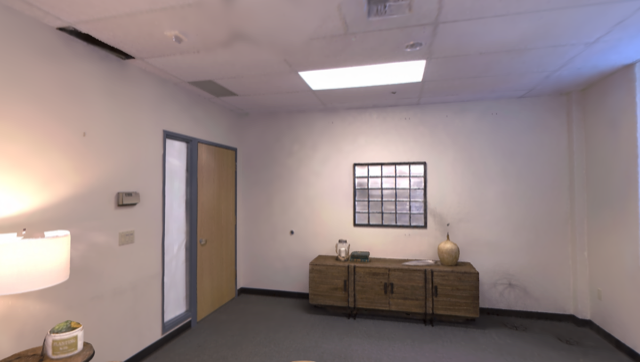}};
    \draw[green, thick] (-0.5, -0.5) rectangle (0.5, 0.5); % Centered box
\end{tikzpicture} &
\begin{tikzpicture}
    \node[inner sep=0pt] (image) at (0.5,-0.7) {\includegraphics[width=0.245\textwidth]{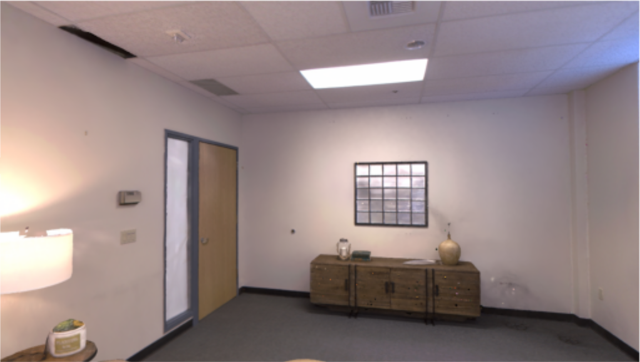}};
    \draw[green, thick] (-0.5, -0.5) rectangle (0.5, 0.5); % Centered box
\end{tikzpicture} &
\begin{tikzpicture}
    \node[inner sep=0pt] (image) at (0.5,-0.7) {\includegraphics[width=0.245\textwidth]{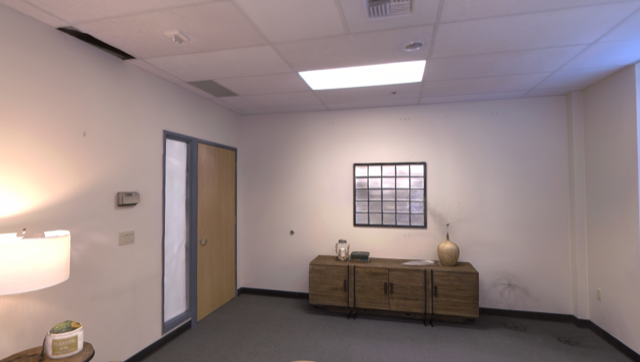}};
    \draw[green, thick] (-0.5, -0.5) rectangle (0.5, 0.5); % Centered box
\end{tikzpicture} &
\begin{tikzpicture}
    \node[inner sep=0pt] (image) at (0.5,-0.7) {\includegraphics[width=0.245\textwidth]{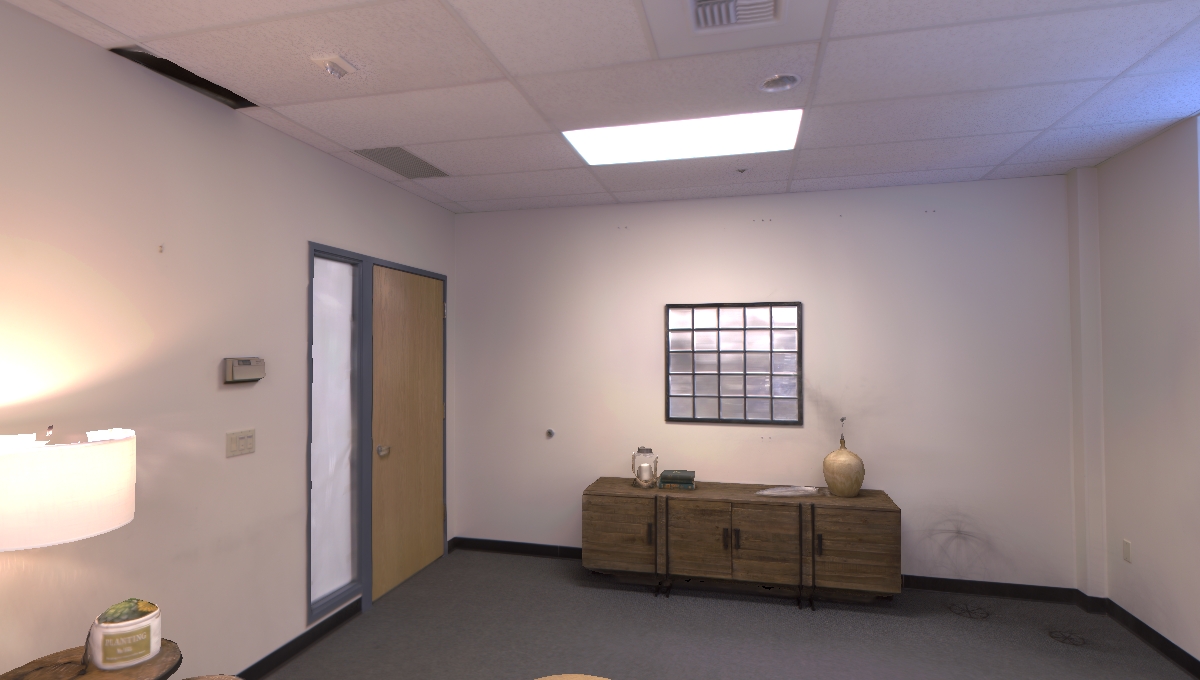}};
    \draw[green, thick] (-0.5, -0.5) rectangle (0.5, 0.5); % Centered box
\end{tikzpicture} \\
 \\[-14pt] % Reduce vertical space

% Row 2
\begin{tikzpicture}
    \node[inner sep=0pt] (image) at (1.6,-0.4) {\includegraphics[width=0.245\textwidth]{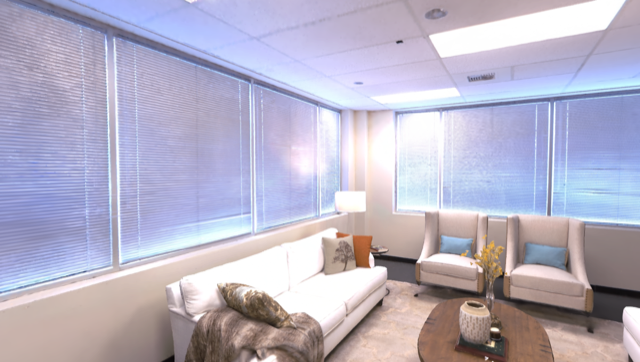}};
    \draw[green, thick] (-0.5, -0.5) rectangle (0.5, 0.5); % Centered box
\end{tikzpicture} &
\begin{tikzpicture}
    \node[inner sep=0pt] (image) at (1.6,-0.4) {\includegraphics[width=0.245\textwidth]{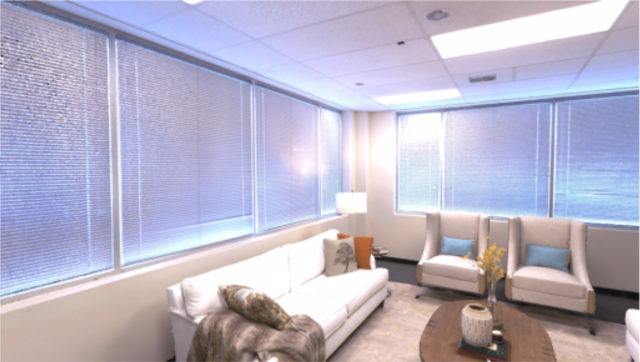}};
    \draw[green, thick] (-0.5, -0.5) rectangle (0.5, 0.5); % Centered box
\end{tikzpicture} &
\begin{tikzpicture}
    \node[inner sep=0pt] (image) at (1.6,-0.4) {\includegraphics[width=0.245\textwidth]{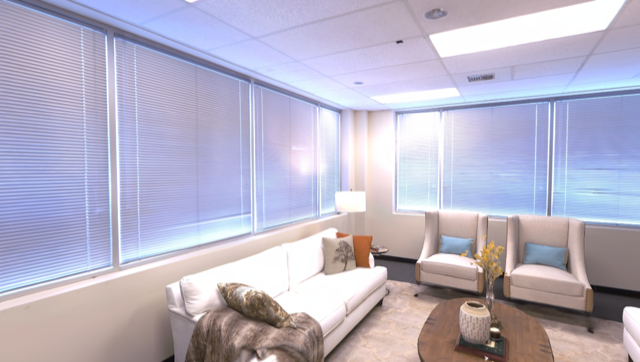}};
    \draw[green, thick] (-0.5, -0.5) rectangle (0.5, 0.5); % Centered box
\end{tikzpicture} &
\begin{tikzpicture}
    \node[inner sep=0pt] (image) at (1.6,-0.4) {\includegraphics[width=0.245\textwidth]{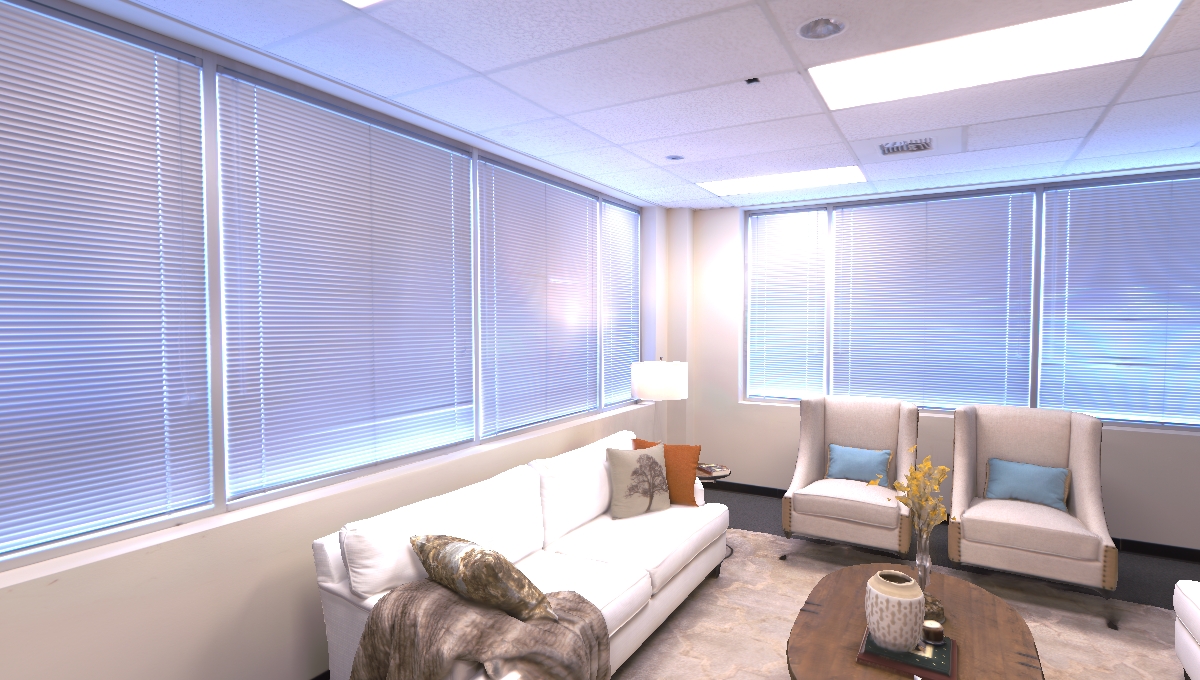}};
    \draw[green, thick] (-0.5, -0.5) rectangle (0.5, 0.5); % Centered box
\end{tikzpicture} \\
 \\[-14pt]

% Row 3
\begin{tikzpicture}
    \node[inner sep=0pt] (image) at (-1.6, -0.4) {\includegraphics[width=0.245\textwidth]{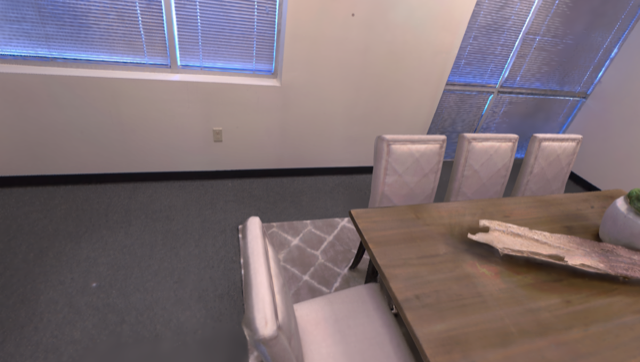}};
    \draw[green, thick] (-0.5, -0.5) rectangle (0.5, 0.5); % Centered box
\end{tikzpicture} &
\begin{tikzpicture}
    \node[inner sep=0pt] (image) at (-1.6, -0.4) {\includegraphics[width=0.245\textwidth]{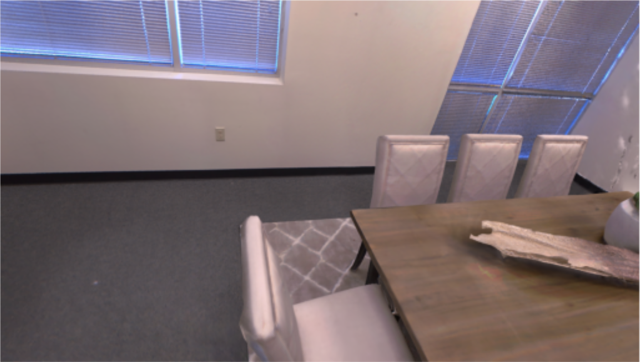}};
    \draw[green, thick] (-0.5, -0.5) rectangle (0.5, 0.5); % Centered box
\end{tikzpicture} &
\begin{tikzpicture}
    \node[inner sep=0pt] (image) at (-1.6, -0.4) {\includegraphics[width=0.245\textwidth]{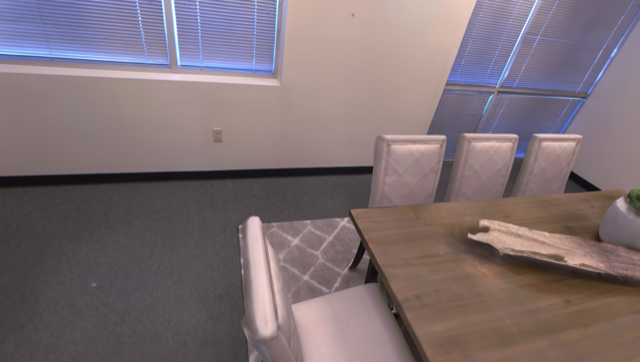}};
    \draw[green, thick] (-0.5, -0.5) rectangle (0.5, 0.5); % Centered box
\end{tikzpicture} &
\begin{tikzpicture}
    \node[inner sep=0pt] (image) at (-1.6, -0.4) {\includegraphics[width=0.245\textwidth]{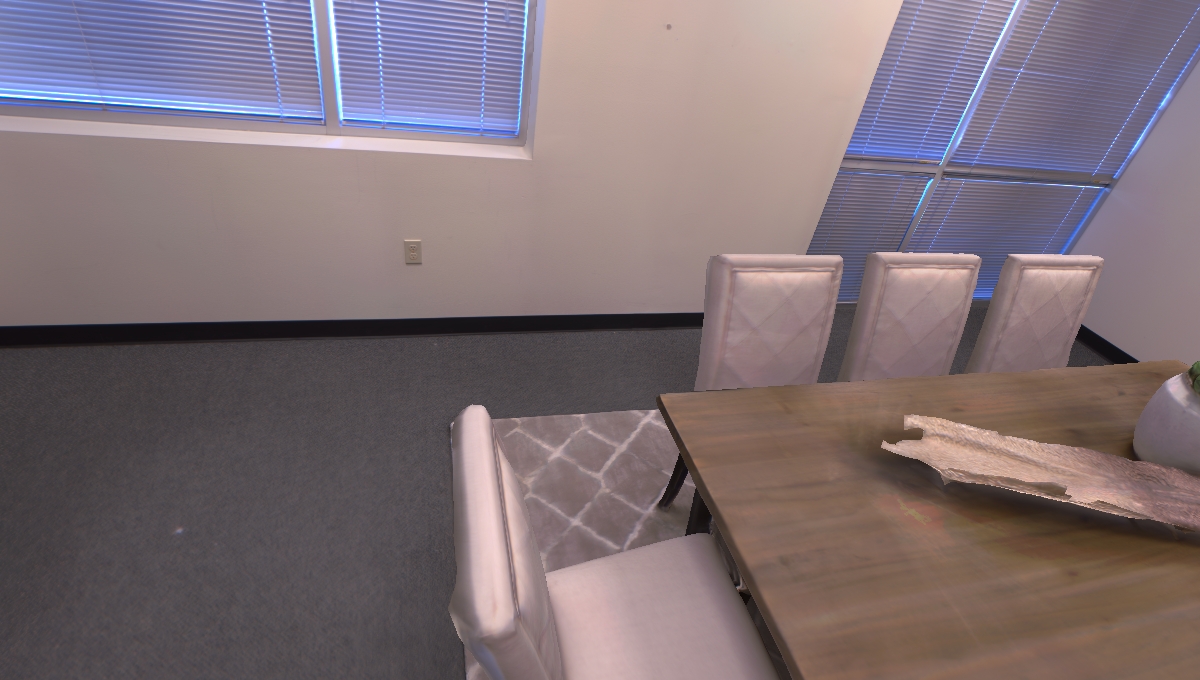}};
    \draw[green, thick] (-0.5, -0.5) rectangle (0.5, 0.5); % Centered box
\end{tikzpicture} \\
 \\[-14pt]

% Row 4
\begin{tikzpicture}
    \node[inner sep=0pt] (image) at (0.5,-0.5) {\includegraphics[width=0.245\textwidth]{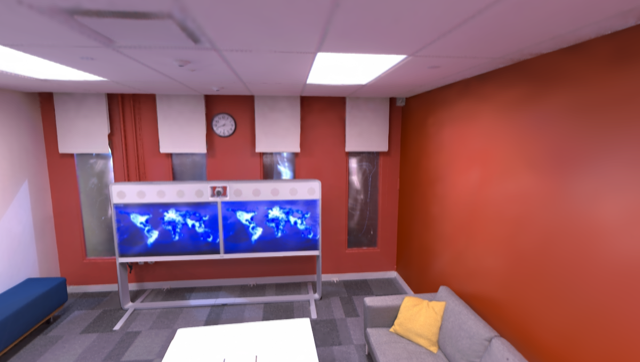}};
    \draw[green, thick] (-0.5, -0.5) rectangle (0.2, 0.2); % Centered box
\end{tikzpicture} &
\begin{tikzpicture}
    \node[inner sep=0pt] (image) at (0.5,-0.5) {\includegraphics[width=0.245\textwidth]{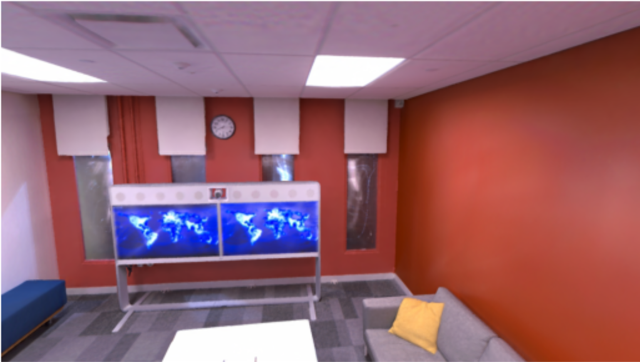}};
    \draw[green, thick] (-0.5, -0.5) rectangle (0.2, 0.2); % Centered box
\end{tikzpicture} &
\begin{tikzpicture}
    \node[inner sep=0pt] (image) at (0.5,-0.5) {\includegraphics[width=0.245\textwidth]{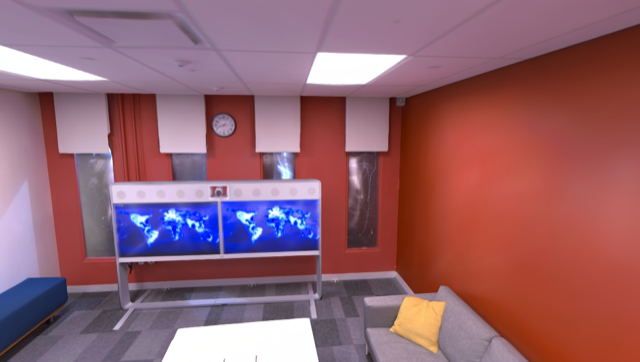}};
    \draw[green, thick] (-0.5, -0.5) rectangle (0.2, 0.2); % Centered box
\end{tikzpicture} &
\begin{tikzpicture}
    \node[inner sep=0pt] (image) at (0.5,-0.5) {\includegraphics[width=0.245\textwidth]{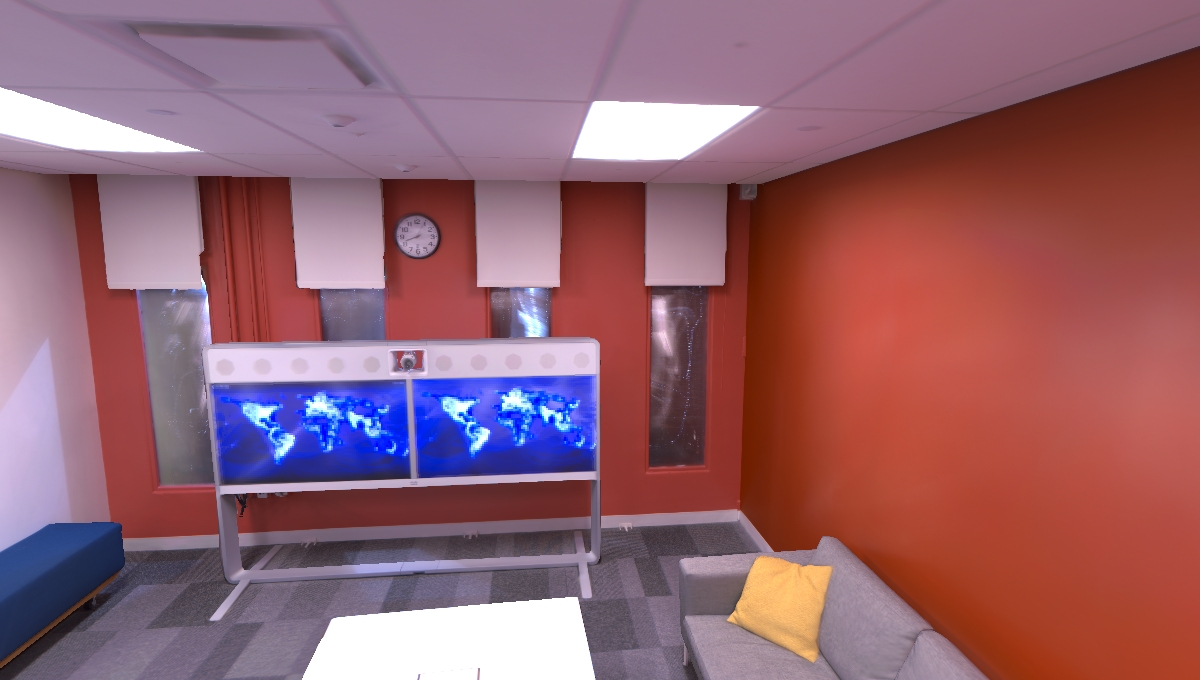}};
    \draw[green, thick] (-0.5, -0.5) rectangle (0.2, 0.2); % Centered box
\end{tikzpicture} \\
 \\[-14pt]

\end{tabular}
\caption{Rendering comparison on Replica dataset.}
\label{fig:qualitative_comparison}
\end{figure*}

The qualitative results shown in Figure \ref{fig:qualitative_comparison} further illustrate the effectiveness of our approach compared to baseline methods on the Replica dataset. Across various scenes, our method demonstrates superior rendering fidelity, accurately capturing fine details and structure in challenging areas, as seen in the boxed regions. For example, in complex textures like window blinds and furniture contours, our approach produces clearer and sharper reconstructions than the baselines. Competing methods exhibit noticeable blurring, artifacts, or misalignment in these regions, while our results more closely resemble the ground truth.

\begin{table}[h!]
\centering
\resizebox{\linewidth}{!}{
\begin{tabular}{lccccccccc}
\toprule
\textbf{Methods} & \textbf{r0} & \textbf{r1} & \textbf{r2} & \textbf{o0} & \textbf{o1} & \textbf{o2} & \textbf{o3} & \textbf{o4} & \textbf{Avg.} \\ 
\midrule
iMAP \cite{imap} & 3.12 & 2.54 & 2.31 & 1.69 & 1.03 & 3.99 & 4.05 & 1.93 & 2.58 \\ 
Vox-Fusion \cite{vox-fusion} & 1.37 & 4.70 & 1.47 & 8.48 & 2.04 & 2.58 & 1.11 & 2.94 & 3.09 \\
NICE-SLAM \cite{nice-slam} & 0.97 & 1.31 & 1.07 & 0.88 & 1.00 & 1.06 & 1.10 & 1.13 & 1.07 \\ 
ESLAM \cite{eslam} & 0.71 & 0.70 & 0.52 & 0.57 & 0.55 & 0.58 & 0.72 & \cellcolor{tabsecond}0.63 & 0.63 \\ 
Point-SLAM \cite{point-slam} & 0.61 & 0.41 & 0.37 & \cellcolor{tabfirst}\textbf{0.38} & 0.48 & 0.54 & 0.69 & 0.72 & \cellcolor{tabthird}0.53 \\
GS-SLAM \cite{gsslam} & \cellcolor{tabthird}0.48 & 0.53 & 0.33 & 0.52 & \cellcolor{tabthird}0.41 & 0.59 & 0.46 & \cellcolor{tabthird}0.70 & \cellcolor{tabsecond}0.50 \\
SplaTAM \cite{splatam} & \cellcolor{tabfirst}\textbf{0.31} & \cellcolor{tabthird}0.40 & \cellcolor{tabsecond}0.29 & 0.47 & \cellcolor{tabfirst}\textbf{0.27} & \cellcolor{tabthird}0.29 & \cellcolor{tabthird}0.32 & \cellcolor{tabfirst}\textbf{0.55} & \cellcolor{tabfirst}\textbf{0.36} \\
MonoGS \cite{monogs} & \cellcolor{tabsecond}0.44 & \cellcolor{tabfirst}\textbf{0.32} & \cellcolor{tabthird}0.31 & \cellcolor{tabthird}0.44 & 0.52 & \cellcolor{tabfirst}\textbf{0.23} & \cellcolor{tabfirst}\textbf{0.17} & 2.25 & 0.58 \\
\textbf{Ours} & \cellcolor{tabfirst}\textbf{0.31} & \cellcolor{tabsecond}0.34	& \cellcolor{tabfirst}\textbf{0.28} & \cellcolor{tabsecond}0.41 & \cellcolor{tabsecond}0.40 & \cellcolor{tabsecond}0.26 & \cellcolor{tabsecond}0.22 & 2.18 & 0.55 \\

\bottomrule
\end{tabular}
}
\caption{Tracking performance on the Replica dataset \cite{replica}, measured by Absolute Trajectory Error (ATE) in centimeters.}
\label{tab:ate_replica}
\end{table}

\paragraph{Tracking Performance.} Table \ref{tab:ate_replica} compares the tracking performance of our SLAM method with several state-of-the-art methods on the Replica dataset \cite{replica}. The metrics presented are tracking errors measured by the average absolute trajectory error (ATE). For each scene, the three best-performing methods are highlighted.

Our method consistently ranks among the top performers, achieving an average tracking error of \textbf{0.55}, which places it alongside the leading methods in terms of tracking accuracy. Specifically, our approach is either the first or second-best in all room scenes (r0, r1, r2), and it performs competitively across office scenes. The performance drop in the Office 4 scene can be attributed to the lack of texture in consecutive frames, which results in poor correspondence detection and matching. This limitation highlights the challenges of maintaining high tracking accuracy in low-texture environments, where reliable feature matching becomes more difficult.

Table \ref{tab:ate_tum} further demonstrates the robustness of our method on the TUM-RGBD dataset \cite{tum-rgbd}, where it achieves competitive results across sequences with an average tracking error of \textbf{4.17}, consistently ranking among the top three performers.

\begin{table}[h!]
\centering
\resizebox{\linewidth}{!}{
\begin{tabular}{lcccccc}
\toprule
\textbf{Methods} & \textbf{fr1/} & \textbf{fr1/} & \textbf{fr1/} & \textbf{fr2/} & \textbf{fr3/} & \textbf{Avg.} \\
& \textbf{desk} & \textbf{desk2} & \textbf{room} & \textbf{xyz} & \textbf{office} & \\
\midrule
ElasticFusion~\cite{elastic-fusion} & 2.53 & 6.83 & 21.49 & 1.17 & 2.52 & 6.91 \\
ORB-SLAM2~\cite{orb-slam2} & \cellcolor{tabthird}1.60 & \cellcolor{tabfirst}\textbf{2.20} & \cellcolor{tabfirst}\textbf{4.70} & \cellcolor{tabfirst}\textbf{0.40} & \cellcolor{tabfirst}\textbf{1.00} & \cellcolor{tabfirst}\textbf{1.98} \\
NICE-SLAM~\cite{nice-slam} & 4.26 & 4.99 & 34.49 & 31.73 & 3.87 & 15.87 \\
Vox-Fusion~\cite{vox-fusion} & 3.52 & 6.00 & \cellcolor{tabthird}9.43 & 1.59 & 26.01 & 11.31 \\
Point-SLAM~\cite{point-slam} & 4.34 & \cellcolor{tabthird}4.54 & 30.92 & 3.23 & 1.84 & 8.92 \\
MonoGS * \cite{monogs} & \cellcolor{tabsecond}1.50 & 6.57 & \cellcolor{tabsecond}5.21 & \cellcolor{tabthird}1.44 & \cellcolor{tabsecond}1.49 & \cellcolor{tabsecond}2.70 \\
SplaTAM \cite{splatam} & 3.35 & 5.64 & 11.13 & \cellcolor{tabsecond}1.24 & 5.16 & 5.48 \\
\textbf{Ours} & \cellcolor{tabfirst}\textbf{1.48} & \cellcolor{tabsecond}3.85 & 11.75 & 1.81 & \cellcolor{tabthird}1.98 & \cellcolor{tabthird}4.17 \\
\bottomrule
\end{tabular}
}
\caption{Tracking performance on TUM-RGBD~\cite{tum-rgbd} (ATE [cm]). Results of the based lines are reported directly from their paper. For MonoGS \cite{monogs}, since perfrmance on \textbf{fr1/ desk2} and \textbf{fr1/ room} are not reported, we reproduced the numbers by running their official released code.}
\label{tab:ate_tum}
\end{table}

\paragraph{Tracking in sparse setting}

Our approach demonstrates high tracking accuracy in sparse environments, where reduced frame density and rapidly changing scenes pose significant challenges. Tables \ref{tab:sparse_tracking_replica} and \ref{tab:sparse_tracking_tum} show tracking performance results on the Replica and TUM-RGBD datasets, where our method consistently achieves the lowest average absolute trajectory error (ATE) across various strides compared to baseline methods such as SplaTAM \cite{splatam} and MonoGS \cite{monogs}. These baseline results were obtained by running their publicly released code.

In these experiments, the stride represents the number of frames skipped in the original dataset. For example, if a dataset contains 2000 frames and a stride of 20 is used, only the $1^{st}$, $20^{th}$, $40^{th}$ frames, and so on are selected, effectively reducing the frame count and increasing sparsity. This setup tests the ability of SLAM methods to handle scenarios with limited frame data.

On the Replica dataset, our method demonstrates superior accuracy, achieving an average ATE of \textbf{0.84} with a stride of 10 and maintaining low ATE values as stride increases, showing minimal performance degradation even at high sparsity levels. Similarly, on the TUM-RGBD dataset, our method maintains low ATE values across all strides, achieving an average ATE of \textbf{3.07} with a stride of 20 and \textbf{3.49} with a stride of 40, significantly outperforming MonoGS \cite{monogs} and SplaTAM \cite{splatam}, which exhibit higher errors under the same conditions. These baseline methods fail in most cases, with the exception of the \textbf{fr2/xyz} scene on TUM-RGBD \cite{tum-rgbd}, where the higher frame count and simpler camera motion make tracking easier.

The key strength of our approach lies in its ability to maintain precise tracking in sparse settings, unlike competing methods that experience significant accuracy degradation as frame sparsity increases. This capability makes our SLAM method highly suitable for real-time applications in bandwidth-constrained environments, where computational resources may limit the frequency of frame capture. The consistent performance across both the Replica and TUM-RGBD datasets confirms our approach as a reliable solution for sparse environments in SLAM applications.

\begin{table}[h]
\centering
\scriptsize
\resizebox{\linewidth}{!}{
\begin{tabular}{lcccccccccc}
\toprule
\textbf{Methods} & \textbf{Stride} & \textbf{r0} & \textbf{r1} & \textbf{r2} & \textbf{o0} & \textbf{o1} & \textbf{o2} & \textbf{o3} & \textbf{o4} & \textbf{Avg.} \\
\midrule
\multirow{3}{*}{\parbox{0.5cm}{SplaTAM \\ \cite{splatam}}} & 10 & \cellcolor{tabthird}166.4 & \cellcolor{tabthird}147.9 & \cellcolor{tabthird}78.4 & \cellcolor{tabthird}168.2 & \cellcolor{tabthird}63.8 & \cellcolor{tabthird}68.9 & \cellcolor{tabthird}962.6 & \cellcolor{tabthird}466.0 & \cellcolor{tabthird}265.3 \\
 & 20 & \cellcolor{tabthird}195.0 & \cellcolor{tabthird}3119 & \cellcolor{tabthird}231.2 & \cellcolor{tabthird}400.4 & \cellcolor{tabthird}90.9 & \cellcolor{tabthird}1129 & \cellcolor{tabthird}170.1 & \cellcolor{tabthird}618.5 & \cellcolor{tabthird}744.3 \\
 & 40 & \cellcolor{tabthird}767.9 & \cellcolor{tabthird}430.3 & \cellcolor{tabthird}117.3 & \cellcolor{tabthird}815.8 & \cellcolor{tabthird}870.0 & \cellcolor{tabthird}506.8 & \cellcolor{tabthird}508.7 & \cellcolor{tabthird}323.8 & \cellcolor{tabthird}542.6 \\
% \multirow{3}{*}{\parbox{0.5cm}{SplaTAM \\ \cite{splatam}}} & 10 & \cellcolor{tabthird}166.39 & \cellcolor{tabthird}147.88 & \cellcolor{tabthird}78.39 & \cellcolor{tabthird}168.21 & \cellcolor{tabthird}63.75 & \cellcolor{tabthird}68.94 & \cellcolor{tabthird}962.64 & \cellcolor{tabthird}465.96 & \cellcolor{tabthird}265.27 \\
%  & 20 & \cellcolor{tabthird}195.00 & \cellcolor{tabthird}3118.61 & \cellcolor{tabthird}231.18 & \cellcolor{tabthird}400.38 & \cellcolor{tabthird}90.86 & \cellcolor{tabthird}1129.13 & \cellcolor{tabthird}170.05 & \cellcolor{tabthird}618.49 & \cellcolor{tabthird}744.21 \\
%  & 40 & \cellcolor{tabthird}767.94 & \cellcolor{tabthird}430.27 & \cellcolor{tabthird}117.31 & \cellcolor{tabthird}815.82 & \cellcolor{tabthird}870.00 & \cellcolor{tabthird}506.78 & \cellcolor{tabthird}508.67 & \cellcolor{tabthird}323.78 & \cellcolor{tabthird}542.57 \\
\midrule
\multirow{3}{*}{\parbox{0.5cm}{MonoGS \\ \cite{monogs}}} & 10 & \cellcolor{tabsecond}37.86 & \cellcolor{tabsecond}25.02 & \cellcolor{tabsecond}15.49 & \cellcolor{tabsecond}14.44 & \cellcolor{tabsecond}9.68 & \cellcolor{tabsecond}33.58 & \cellcolor{tabsecond}29.63 & \cellcolor{tabsecond}36.05 & \cellcolor{tabsecond}25.22 \\
 & 20 & \cellcolor{tabsecond}92.07 & \cellcolor{tabsecond}80.95 & \cellcolor{tabsecond}53.34 & \cellcolor{tabsecond}67.80 & \cellcolor{tabsecond}20.93 & \cellcolor{tabsecond}101.69 & \cellcolor{tabsecond}89.44 & \cellcolor{tabsecond}113.69 & \cellcolor{tabsecond}77.49 \\
 & 40 & \cellcolor{tabsecond}93.14 & \cellcolor{tabsecond}84.24 & \cellcolor{tabsecond}94.03 & \cellcolor{tabsecond}78.44 & \cellcolor{tabsecond}42.13 & \cellcolor{tabsecond}123.12 & \cellcolor{tabsecond}121.31 & \cellcolor{tabsecond}109.36 & \cellcolor{tabsecond}93.22 \\
\midrule
\multirow{3}{*}{\textbf{Ours}} & 10 & \cellcolor{tabfirst}\textbf{0.46} & \cellcolor{tabfirst}\textbf{2.17} & \cellcolor{tabfirst}\textbf{0.88} & \cellcolor{tabfirst}\textbf{0.56} & \cellcolor{tabfirst}\textbf{0.84} & \cellcolor{tabfirst}\textbf{0.46} & \cellcolor{tabfirst}\textbf{0.52}& \cellcolor{tabfirst}\textbf{0.85} & \cellcolor{tabfirst}\textbf{0.84} \\
 & 20 & \cellcolor{tabfirst}\textbf{0.49} & \cellcolor{tabfirst}\textbf{8.14} & \cellcolor{tabfirst}\textbf{1.57} & \cellcolor{tabfirst}\textbf{2.46} & \cellcolor{tabfirst}\textbf{2.80} & \cellcolor{tabfirst}\textbf{5.21} & \cellcolor{tabfirst}\textbf{0.71} & \cellcolor{tabfirst}\textbf{4.96} & \cellcolor{tabfirst}\textbf{3.29} \\
 & 40 & \cellcolor{tabfirst}\textbf{0.45} & \cellcolor{tabfirst}\textbf{9.53} & \cellcolor{tabfirst}\textbf{1.25} & \cellcolor{tabfirst}\textbf{23.58} & \cellcolor{tabfirst}\textbf{6.10} & \cellcolor{tabfirst}\textbf{1.04} & \cellcolor{tabfirst}\textbf{0.85} & \cellcolor{tabfirst}\textbf{6.29} & \cellcolor{tabfirst}\textbf{6.14} \\
\bottomrule
\end{tabular}
}
\caption{Tracking performance with different strides on various scenes from the Replica dataset.}
\label{tab:sparse_tracking_replica}
\end{table}

\begin{table}[h]
\centering
\resizebox{\linewidth}{!}{
\begin{tabular}{llllllll}
\toprule
\textbf{Methods} & \textbf{Stride} & \textbf{fr1/ desk} & \textbf{fr1/ desk2} & \textbf{fr1/ room} & \textbf{fr2/ xyz} & \textbf{fr3/ office} & \textbf{Avg.} \\
\midrule
\multirow{3}{*}{SplaTAM \cite{splatam}} & 10 & \cellcolor{tabthird}295.40 & \cellcolor{tabthird}593.82 & \cellcolor{tabthird}2206.30 & \cellcolor{tabthird}1.38 & \cellcolor{tabthird}187.95 & \cellcolor{tabthird}656.97 \\
 & 20 & \cellcolor{tabthird}194.59 & \cellcolor{tabthird}328.94 & \cellcolor{tabthird}1091.26 & \cellcolor{tabsecond}1.40 & \cellcolor{tabthird}3116.74 & \cellcolor{tabthird}946.59 \\
 & 40 & \cellcolor{tabthird}185.69 & \cellcolor{tabthird}250.32 & \cellcolor{tabthird}439.97 & \cellcolor{tabfirst}\textbf{1.44} & \cellcolor{tabthird}1074.38 & \cellcolor{tabthird}390.36 \\
\midrule
\multirow{3}{*}{MonoGS \cite{monogs}} & 10 & \cellcolor{tabsecond}73.75 & \cellcolor{tabsecond}90.90 & \cellcolor{tabsecond}97.29 & \cellcolor{tabfirst}\textbf{0.95} & \cellcolor{tabsecond}117.88 & \cellcolor{tabsecond}76.15 \\
 & 20 & \cellcolor{tabsecond}81.34 & \cellcolor{tabsecond}85.08 & \cellcolor{tabsecond}96.98 & \cellcolor{tabfirst}\textbf{0.93} & \cellcolor{tabsecond}189.52 & \cellcolor{tabsecond}90.77 \\
 & 40 & \cellcolor{tabsecond}84.05 & \cellcolor{tabsecond}84.65 & \cellcolor{tabsecond}89.76 & \cellcolor{tabthird}9.13 & \cellcolor{tabsecond}191.63 & \cellcolor{tabsecond}91.84 \\
\midrule
\multirow{3}{*}{\textbf{Ours}} & 10 & \cellcolor{tabfirst}\textbf{1.48} & \cellcolor{tabfirst}\textbf{6.87} & \cellcolor{tabfirst}\textbf{7.24} & \cellcolor{tabthird}1.86 & \cellcolor{tabfirst}\textbf{1.76} & \cellcolor{tabfirst}\textbf{3.84} \\
 & 20 & \cellcolor{tabfirst}\textbf{1.44} & \cellcolor{tabfirst}\textbf{3.60} & \cellcolor{tabfirst}\textbf{6.76} & \cellcolor{tabthird}1.77 & \cellcolor{tabfirst}\textbf{1.78} & \cellcolor{tabfirst}\textbf{3.07} \\
 & 40 & \cellcolor{tabfirst}\textbf{1.48} & \cellcolor{tabfirst}\textbf{4.25} & \cellcolor{tabfirst}\textbf{8.07} & \cellcolor{tabsecond}1.68 & \cellcolor{tabfirst}\textbf{1.98} & \cellcolor{tabfirst}\textbf{3.49} \\

\bottomrule
\end{tabular}
}
\caption{Tracking performance with sparse setting on the TUM-RGBD \cite{tum-rgbd} dataset.}
\label{tab:sparse_tracking_tum}
\end{table}

%%%%%%%%%%%%%%

\section{Ablation Studies}
\label{sec:ablation}

\paragraph{Gradient-based pose refinement.} We evaluate gradient-based camera tracking refinement by varying the number of tracking iterations, where each iteration involves rendering the scene to optimize the camera pose. Results in Table \ref{tab:tracking_ablation} show that increasing iterations improves accuracy (lower ATE and higher PSNR) but also increases tracking time per frame.

Without refinement, we achieve an ATE of 0.65 cm and PSNR of 35.32 dB in 78 ms per frame. The optimal balance between accuracy and speed occurs at \textbf{50} iterations, achieving the lowest ATE (0.55 cm) and highest PSNR (39.21 dB) in 485 ms per frame. Increasing to 70 iterations offers minimal gains in accuracy while significantly raising processing time. Notably, other 3DGS-based methods require 100-200 tracking iterations to reach similar quality in dense settings but still fail completely in sparse settings.

\begin{table}[t]
\centering

\scriptsize
\setlength{\tabcolsep}{5.5pt}
\begin{tabular}{ccccccc}
\toprule

\textbf{Pose} & \textbf{Track.} & \textbf{ATE} & \textbf{PSNR} & \textbf{Track. time} \\
\textbf{refinement} & \textbf{iter} & [cm]$\downarrow$ & [dB]$\uparrow$ & [ms / frame] \\

\midrule

\redx & 0 & 0.65 & 35.32 & \cellcolor{tabfirst}\textbf{78}  \\

\greencheck & 10 & 0.62 & 36.25 & 310\\

\greencheck & 30 & 0.59 & 38.15 & 401\\

\greencheck & 50 & \cellcolor{tabfirst}\textbf{0.55} & \cellcolor{tabfirst}\textbf{39.21} & 485\\
\greencheck & 70 & 0.56 & 38.10 & 512\\

\bottomrule
\end{tabular}
\caption{Camera tracking refinement ablations on Replica \cite{replica}. The results are averaged.}
\label{tab:tracking_ablation}
\end{table}

\paragraph{Priority sampling strategies.}

Table \ref{tab:sampling_ablation} presents an ablation study on color refinement using different sampling strategies. The three strategies evaluated are Random Sampling, Worst-First Sampling, and Loss-Weighted Sampling. Results show that the Loss-Weighted Sampling strategy achieves the highest PSNR on both the Replica and TUM datasets, with values of 39.21 and 22.85 dB, respectively. This indicates that prioritizing frames based on their loss magnitude improves color alignment and overall reconstruction quality more effectively than random or worst-first approaches.

\begin{table}[t]
\centering

\scriptsize
\setlength{\tabcolsep}{5.5pt}
\begin{tabular}{ccccccc}
\toprule

\textbf{Sampling} & \textbf{Replica} & \textbf{TUM} \\

\midrule

Random & 37.62 & 22.15 \\

Worst first & 38.56 & 22.64\\

Loss-weighted & 39.21 & 22.85\\

\bottomrule
\end{tabular}
\caption{Ablation on color refinement with sampling strategies.}
\label{tab:sampling_ablation}
\end{table}

\section{Conclusion}

We propose a novel SLAM approach that utilizes 3D Gaussian Splatting (3DGS) for real-time 3D scene reconstruction, paired with an efficient vision-based camera tracking strategy. This method delivers high-fidelity reconstructions and robust performance against noisy depth sensor data, making it suitable for consumer-grade cameras. By incorporating pretrained feature matching and point cloud registration, FlashSLAM achieves significant improvements in tracking speed and accuracy, excelling in challenging scenarios such as sparse settings and rapid camera movements. Comprehensive evaluations on benchmark datasets confirm FlashSLAM's superior accuracy and efficiency compared to existing methods.

\paragraph{Limitations.} Our method may fail in scenarios with excessive noise in depth sensor data or non-textured images, as these conditions hinder our vision-based camera tracking approach.

% \newpage

{
    \small
    \bibliographystyle{ieeenat_fullname}
    \bibliography{main}
}

\clearpage
\maketitlesupplementary

\section{Novel View Synthesis}

\begin{figure*}[htbp]
\centering
{\footnotesize
\setlength{\tabcolsep}{1pt}
\renewcommand{\arraystretch}{0.99}
\newcommand{\sz}{0.16}
\begin{tabular}{ccccccc}
\rotatebox[origin=c]{90}{\tiny SplaTAM~\cite{splatam}} & 
\raisebox{-0.5\height}{\includegraphics[width=\sz\linewidth]{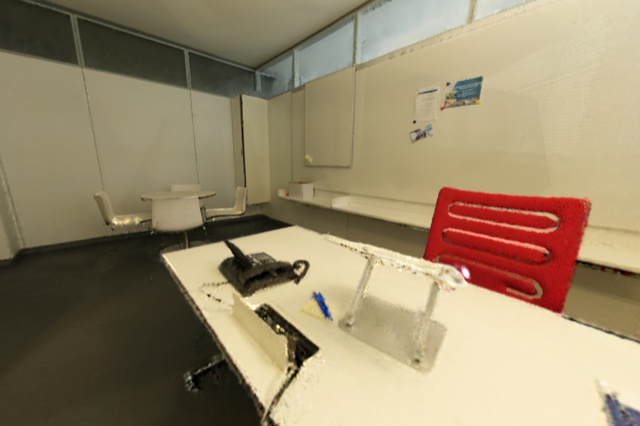}} & 
\raisebox{-0.5\height}{\includegraphics[width=\sz\linewidth]{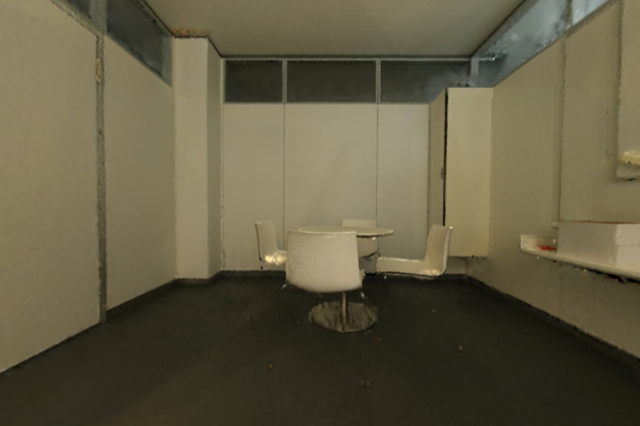}} &
\raisebox{-0.5\height}{\includegraphics[width=\sz\linewidth]{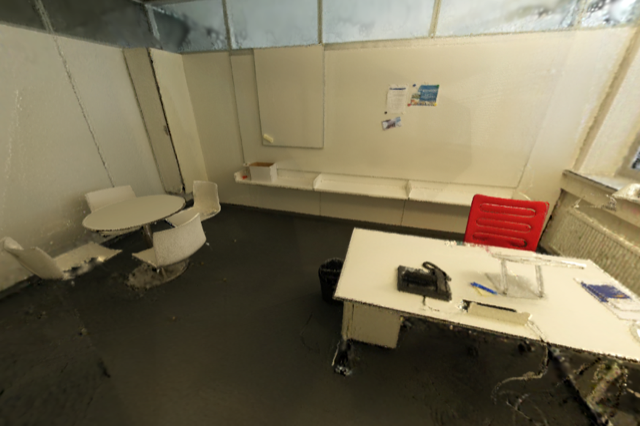}} &
\raisebox{-0.5\height}{\includegraphics[width=\sz\linewidth]{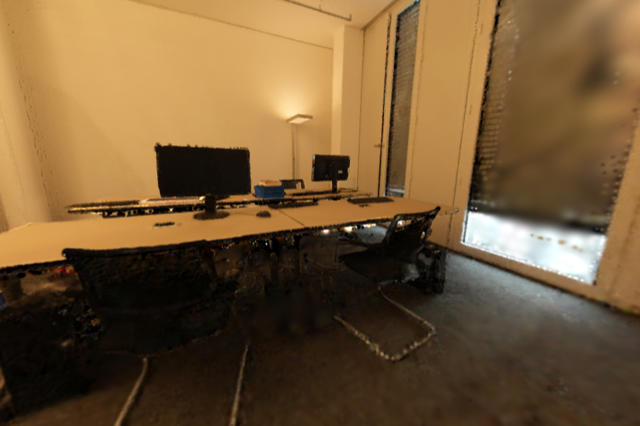}} & 
\raisebox{-0.5\height}{\includegraphics[width=\sz\linewidth]{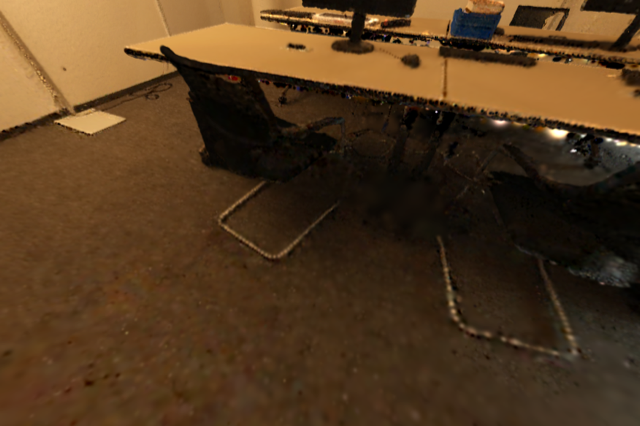}} & 
\raisebox{-0.5\height}{\includegraphics[width=\sz\linewidth]{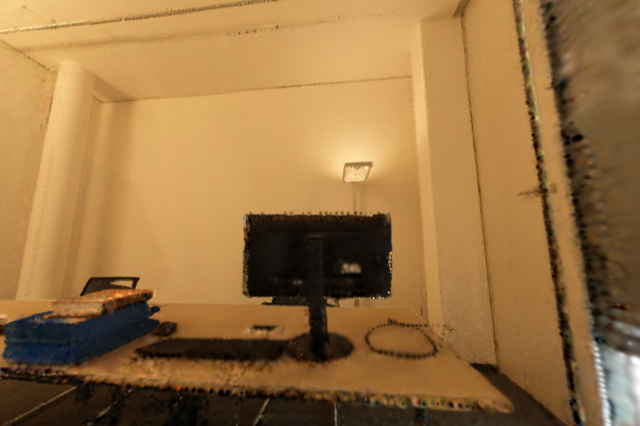}}
\\
\rotatebox[origin=c]{90}{\textbf{\tiny Ours}} & 
\raisebox{-0.5\height}{\includegraphics[width=\sz\linewidth]{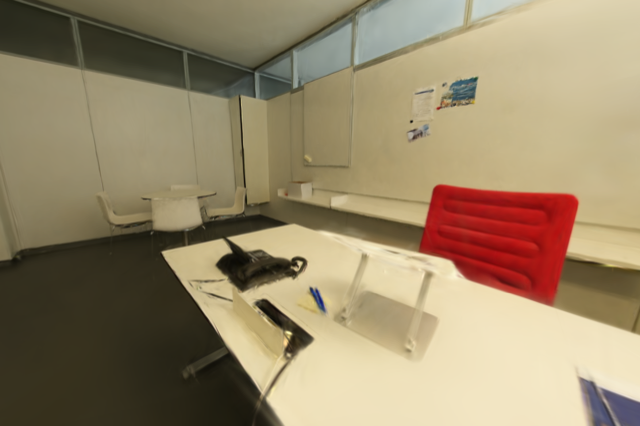}} & 
\raisebox{-0.5\height}{\includegraphics[width=\sz\linewidth]{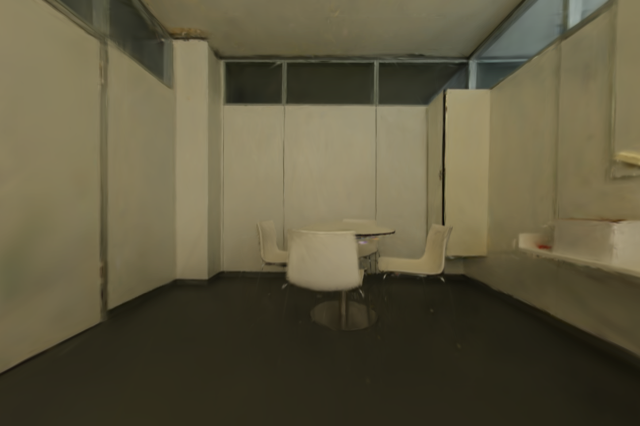}} &
\raisebox{-0.5\height}{\includegraphics[width=\sz\linewidth]{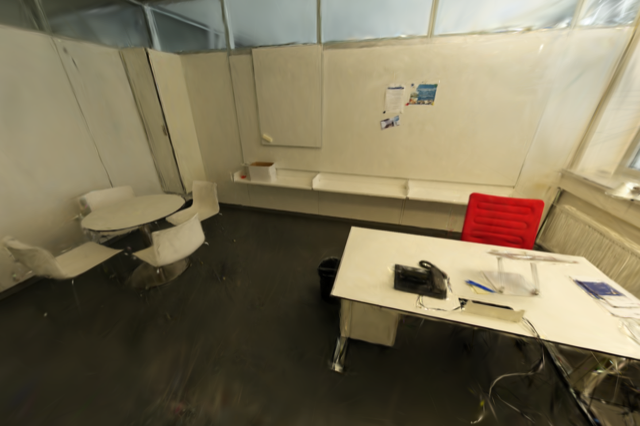}} &
\raisebox{-0.5\height}{\includegraphics[width=\sz\linewidth]{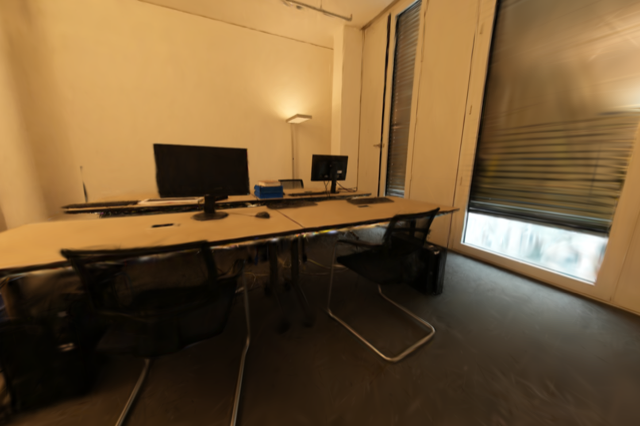}} & 
\raisebox{-0.5\height}{\includegraphics[width=\sz\linewidth]{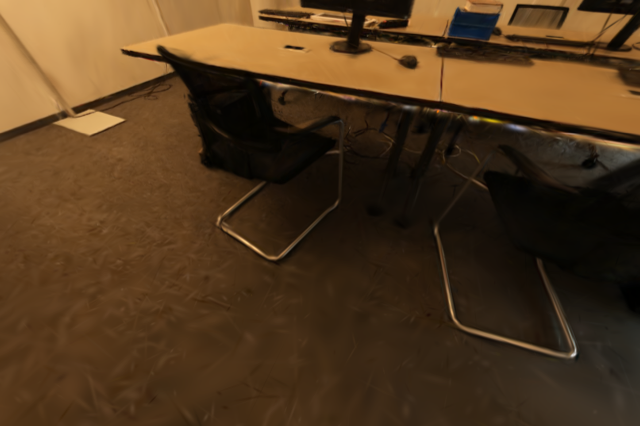}} & 
\raisebox{-0.5\height}{\includegraphics[width=\sz\linewidth]{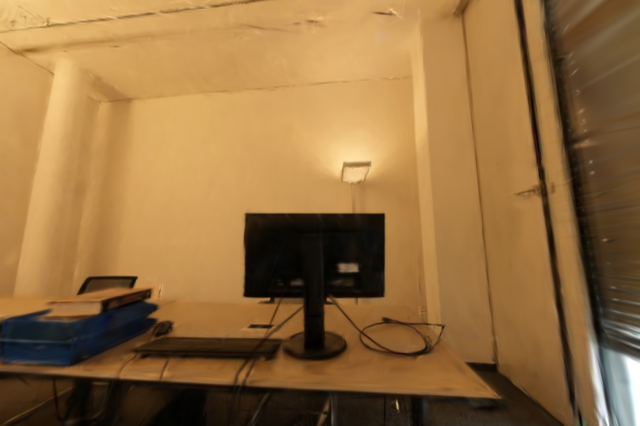}}
\\
\rotatebox[origin=c]{90}{\tiny GT} & 
\raisebox{-0.5\height}{\includegraphics[width=\sz\linewidth]{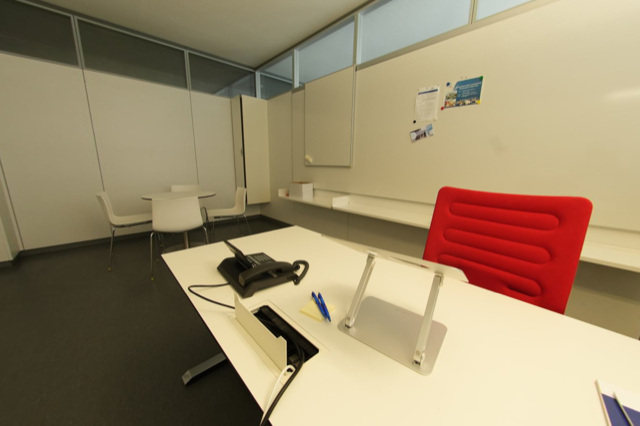}} & 
\raisebox{-0.5\height}{\includegraphics[width=\sz\linewidth]{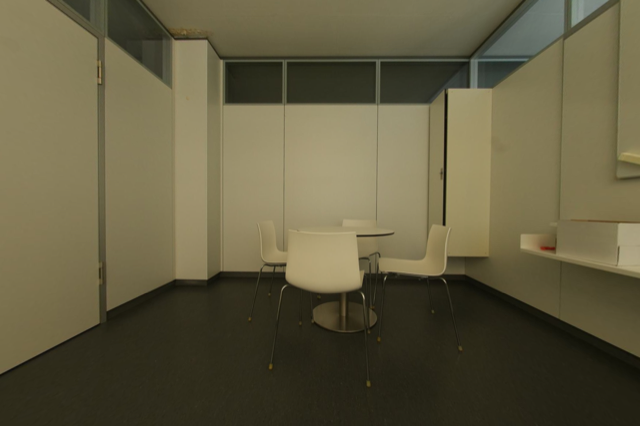}} &
\raisebox{-0.5\height}{\includegraphics[width=\sz\linewidth]{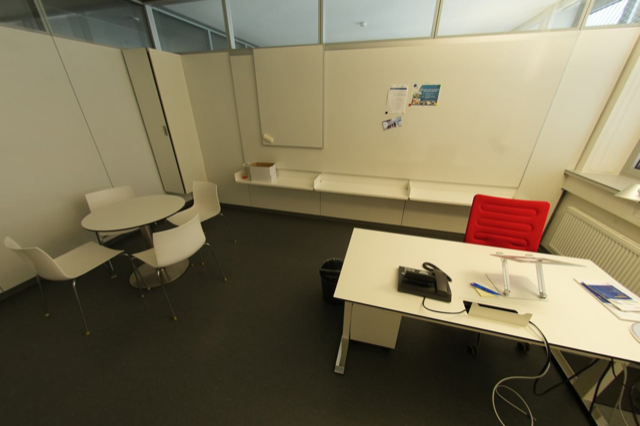}} &
\raisebox{-0.5\height}{\includegraphics[width=\sz\linewidth]{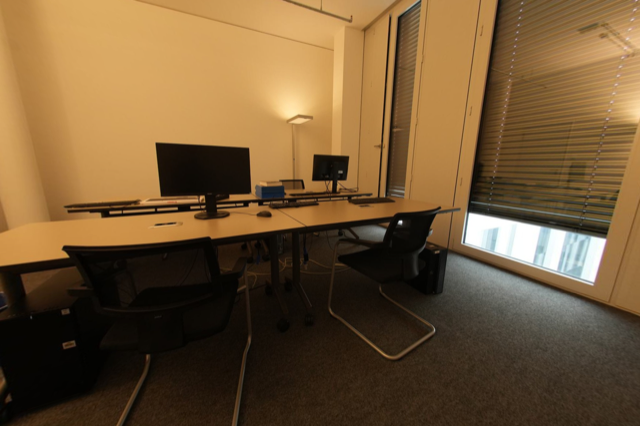}} & 
\raisebox{-0.5\height}{\includegraphics[width=\sz\linewidth]{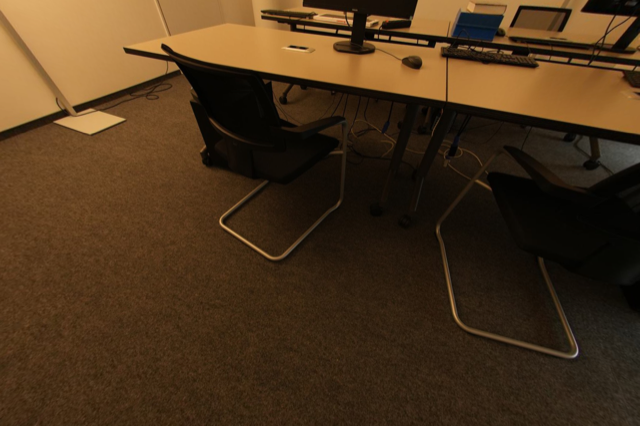}} & 
\raisebox{-0.5\height}{\includegraphics[width=\sz\linewidth]{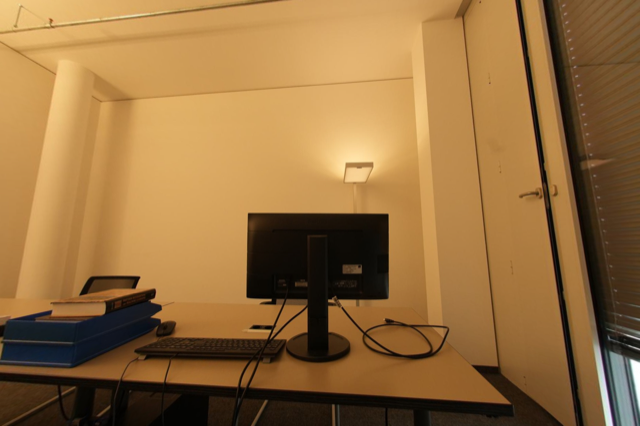}}
\\
& \multicolumn{3}{c}{\texttt{8b5caf3398}} & \multicolumn{3}{c}{\texttt{b20a261fdf}}
\end{tabular}
}
\caption{Novel view synthesis results for two scenes from the ScanNet++ dataset \cite{scannet++}. The first three columns show the results for scene ID \texttt{8b5caf3398}, and the last three columns correspond to scene ID \texttt{b20a261fdf}.}
\label{fig:novel_view1}
\end{figure*}

\begin{figure*}[htbp]
\centering
{\footnotesize
\setlength{\tabcolsep}{1pt}
\renewcommand{\arraystretch}{0.9}
\newcommand{\sz}{0.12}
\begin{tabular}{ccccccccc}
\rotatebox[origin=c]{90}{\tiny SplaTAM~\cite{splatam}} & 
\raisebox{-0.5\height}{\includegraphics[width=\sz\linewidth]{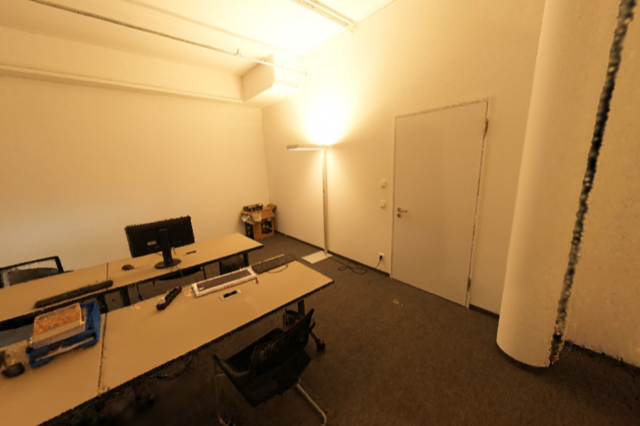}} & 
\raisebox{-0.5\height}{\includegraphics[width=\sz\linewidth]{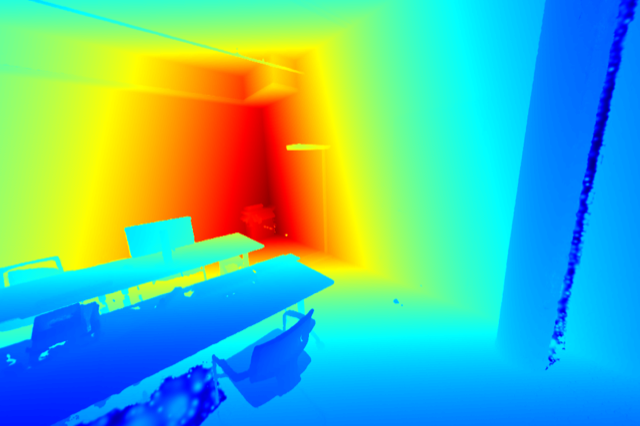}} &
\raisebox{-0.5\height}{\includegraphics[width=\sz\linewidth]{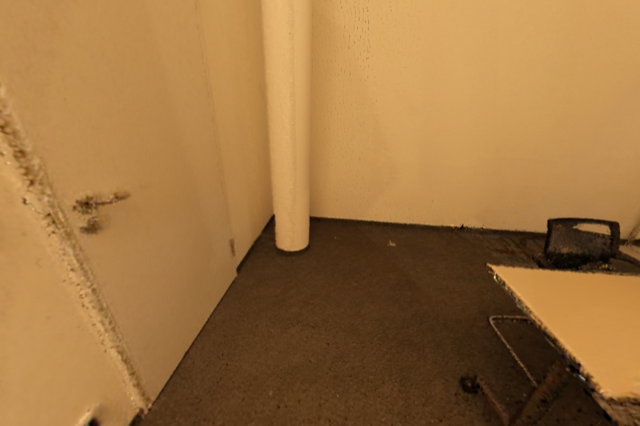}} & 
\raisebox{-0.5\height}{\includegraphics[width=\sz\linewidth]{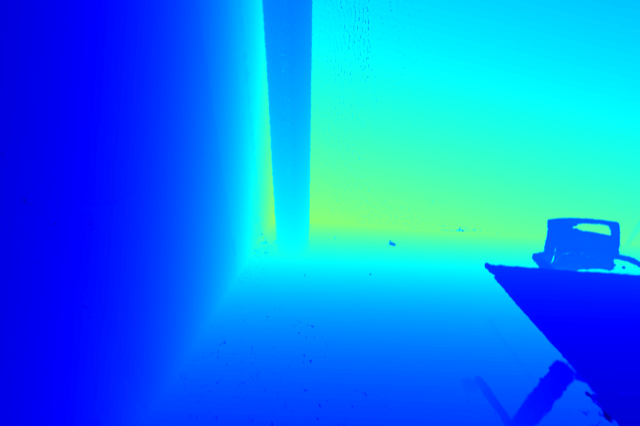}} &
\raisebox{-0.5\height}{\includegraphics[width=\sz\linewidth]{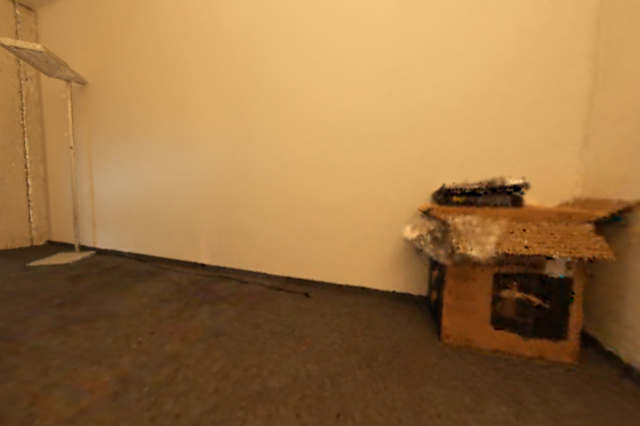}} & 
\raisebox{-0.5\height}{\includegraphics[width=\sz\linewidth]{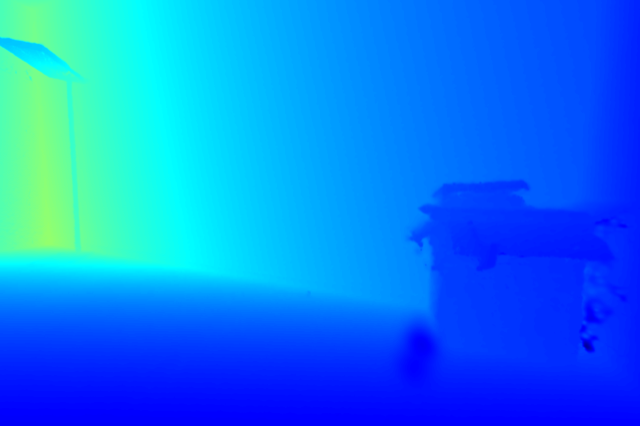}} &
\raisebox{-0.5\height}{\includegraphics[width=\sz\linewidth]{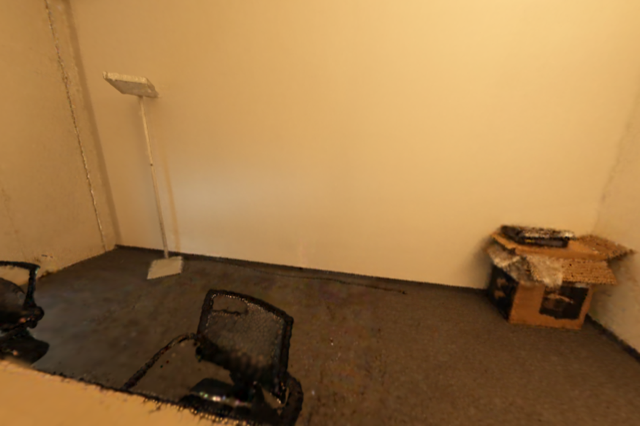}} & 
\raisebox{-0.5\height}{\includegraphics[width=\sz\linewidth]{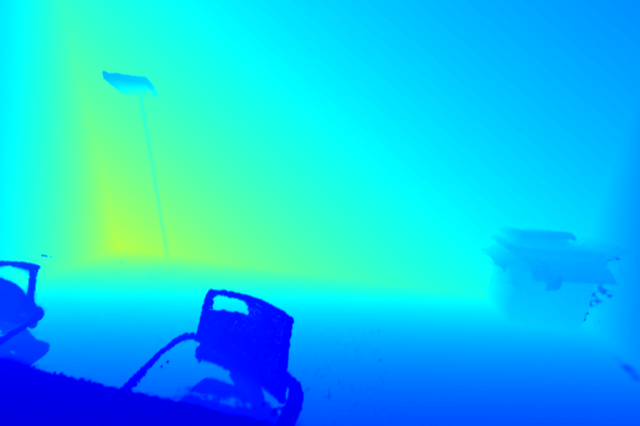}} 
\\
\rotatebox[origin=c]{90}{\textbf{\tiny Ours}} & 
\raisebox{-0.5\height}{\includegraphics[width=\sz\linewidth]{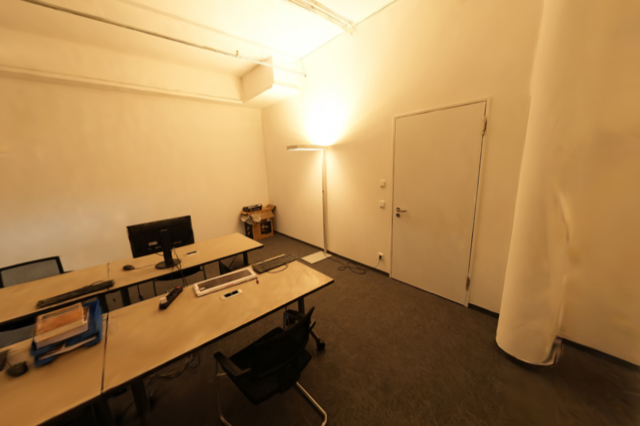}} & 
\raisebox{-0.5\height}{\includegraphics[width=\sz\linewidth]{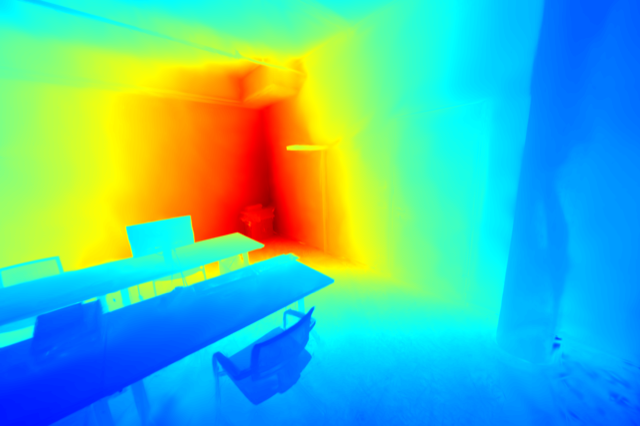}} &
\raisebox{-0.5\height}{\includegraphics[width=\sz\linewidth]{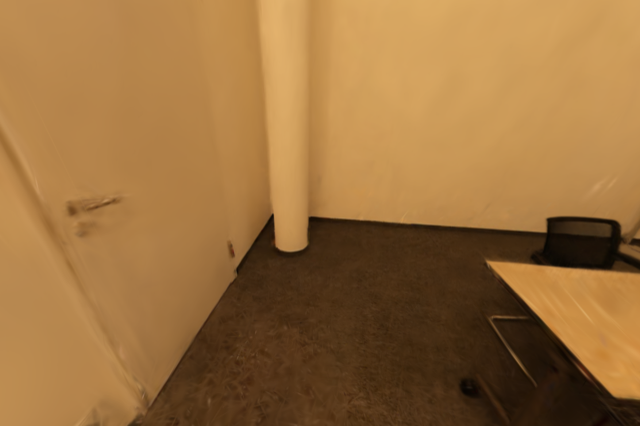}} & 
\raisebox{-0.5\height}{\includegraphics[width=\sz\linewidth]{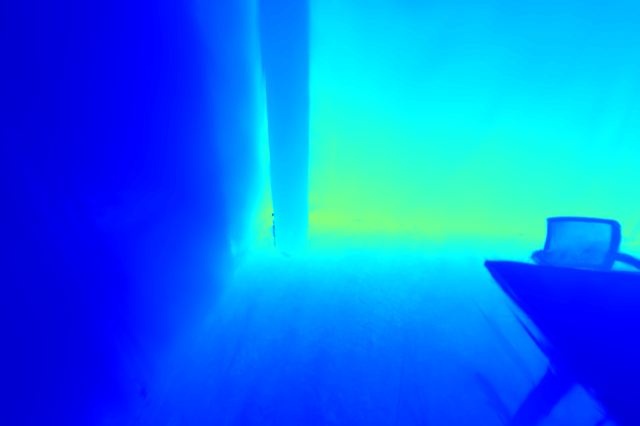}} &
\raisebox{-0.5\height}{\includegraphics[width=\sz\linewidth]{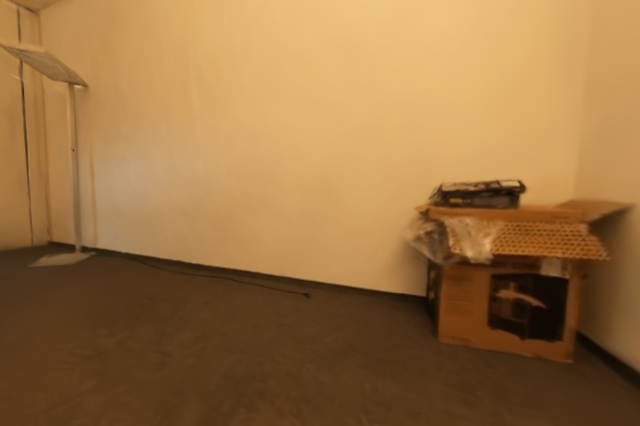}} & 
\raisebox{-0.5\height}{\includegraphics[width=\sz\linewidth]{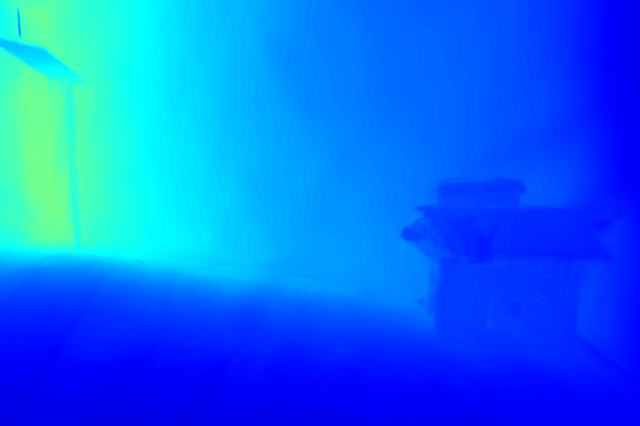}} &
\raisebox{-0.5\height}{\includegraphics[width=\sz\linewidth]{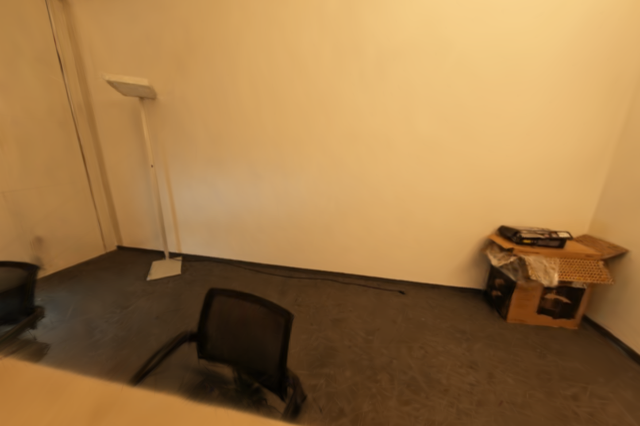}} & 
\raisebox{-0.5\height}{\includegraphics[width=\sz\linewidth]{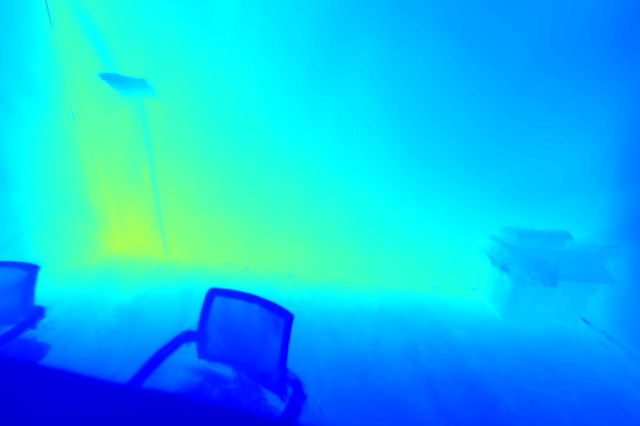}} 
\\
\rotatebox[origin=c]{90}{\tiny GT} & 
\raisebox{-0.5\height}{\includegraphics[width=\sz\linewidth]{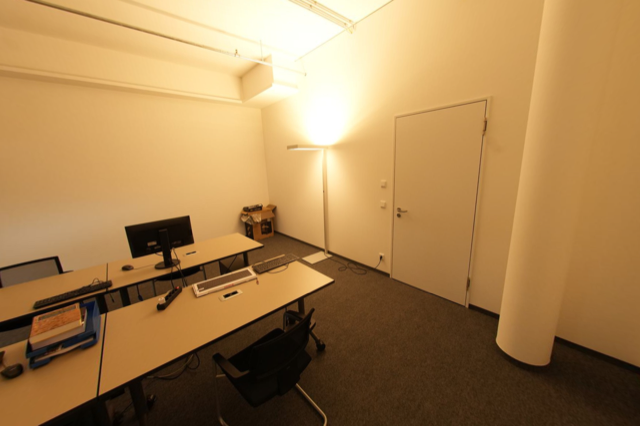}} & 
\raisebox{-0.5\height}{\includegraphics[width=\sz\linewidth]{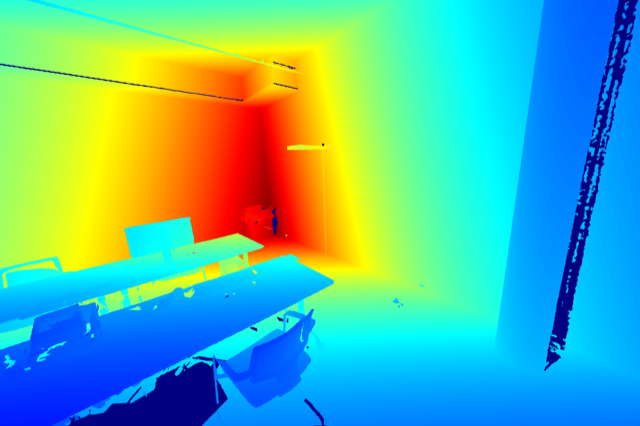}} &
\raisebox{-0.5\height}{\includegraphics[width=\sz\linewidth]{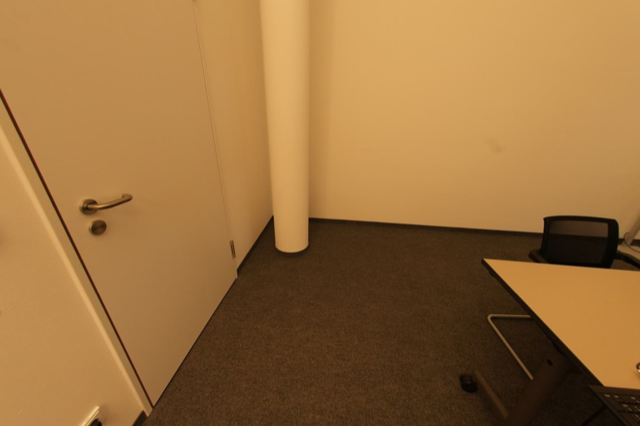}} & 
\raisebox{-0.5\height}{\includegraphics[width=\sz\linewidth]{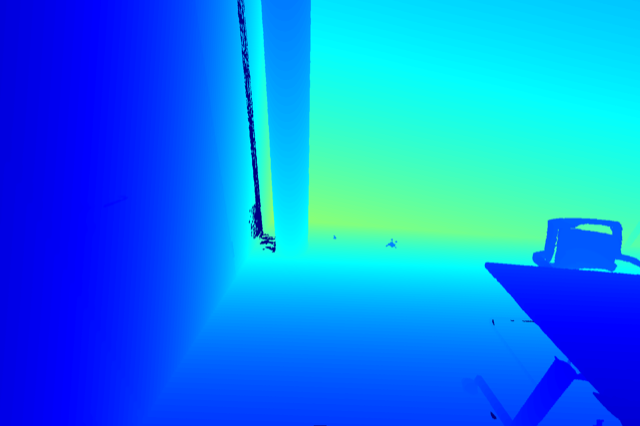}} &
\raisebox{-0.5\height}{\includegraphics[width=\sz\linewidth]{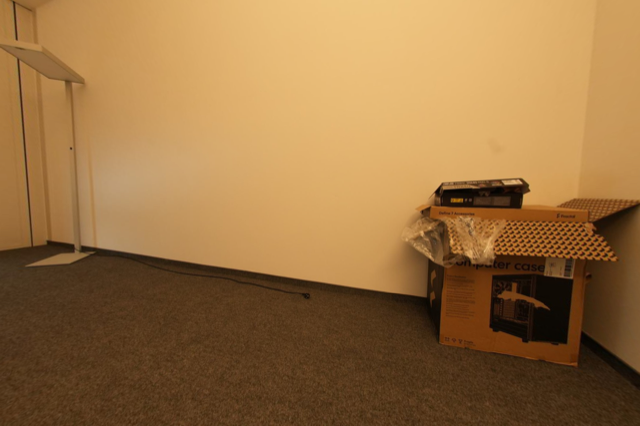}} & 
\raisebox{-0.5\height}{\includegraphics[width=\sz\linewidth]{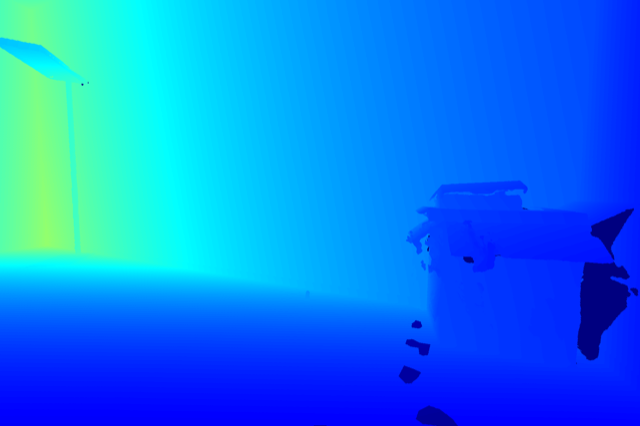}} &
\raisebox{-0.5\height}{\includegraphics[width=\sz\linewidth]{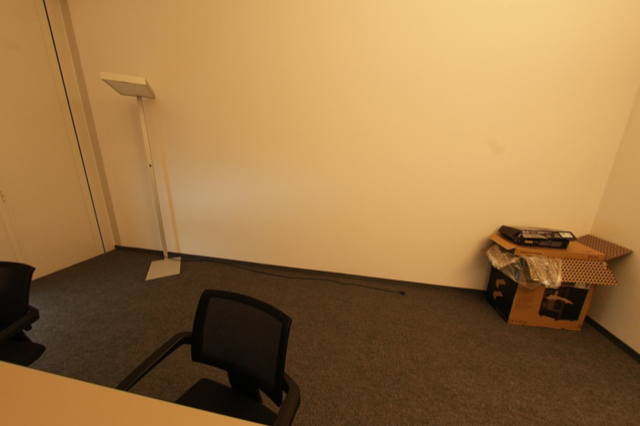}} & 
\raisebox{-0.5\height}{\includegraphics[width=\sz\linewidth]{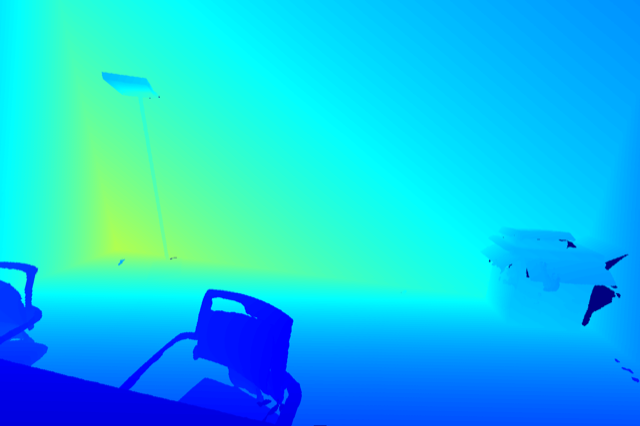}} 
\\
% & \multicolumn{8}{c}{\texttt{\tiny b20a261fdf}}
\end{tabular}
}
\caption{Novel view synthesis results with depth for scene \texttt{b20a261fdf} from the ScanNet++ dataset \cite{scannet++}. The left columns display RGB images, and the right columns show the corresponding depth maps.}
\label{fig:novel_view_depth}
\end{figure*}

Figures \ref{fig:novel_view1} and \ref{fig:novel_view_depth} showcase the novel view synthesis results on the ScanNet++ dataset \cite{scannet++}, highlighting both RGB and depth outputs for two representative scenes. In Figure \ref{fig:novel_view1}, we present qualitative comparisons for scene IDs \texttt{8b5caf3398} (first three columns) and \texttt{b20a261fdf} (last three columns). The top, middle, and bottom rows correspond to results from the baseline method (SplaTAM \cite{splatam}), our proposed method, and the ground truth (GT), respectively. Compared to SplaTAM, our method achieves enhanced fidelity in rendering fine details, including edges, textures, and object boundaries, resulting in closer alignment with the ground truth. For instance, in scene \texttt{8b5caf3398}, our results accurately reconstruct the textures of the red chair and white table, while the baseline suffers from noticeable artifacts and blurring. Similarly, for scene \texttt{b20a261fdf}, our method maintains the structural integrity of the monitors and desks, providing better spatial consistency.

Figure \ref{fig:novel_view_depth} further evaluates novel view synthesis performance by including depth map comparisons for scene \texttt{b20a261fdf}. The left columns depict RGB images, while the right columns show the corresponding depth maps for the same viewpoints. Our method demonstrates robustness to depth noise present in the ground truth. For example, in the first view, there is a discontinuity in the ground truth depth at the pillar, and in the second and last views, noisy depth values are observed on top of the chair. These artifacts result in missing details in SplaTAM \cite{splatam}, as evident in the incomplete reconstruction of these regions. In contrast, our method retains the integrity of the reconstruction, producing depth maps with smoother transitions and fewer missing regions. Additionally, our depth predictions better capture object boundaries and spatial coherence, as evident in regions such as the edges of desks and chairs. These results highlight the effectiveness and reliability of our approach in generating both high-fidelity RGB images and accurate depth maps, making it well-suited for novel view synthesis in 3D scene reconstruction.

\section{Additional Tracking Results}

A comparative evaluation of tracking performance for various SLAM methods is provided on the ScanNet \cite{scannet} and ScanNet++ \cite{scannet++} datasets, as shown in Tables \ref{tab:ate_scannet} and \ref{tab:ate_scannet++}. These tables highlight the accuracy and robustness of our approach in challenging scenarios, demonstrating its ability to outperform baseline methods under varying conditions.

Table \ref{tab:ate_scannet} reports tracking performance on the ScanNet dataset across multiple sequences, including \texttt{0000}, \texttt{0059}, \texttt{0106}, \texttt{0169}, \texttt{0181}, and \texttt{0207}. Our approach achieves the best average ATE ($10.33$ cm), outperforming competing methods such as NICE-SLAM ($10.73$ cm) and SplaTAM ($11.88$ cm). Moreover, our method consistently ranks in the top three for all sequences, underlining its robustness and generalizability across diverse scenes with varying complexities.

Table \ref{tab:ate_scannet++} presents the tracking performance on the ScanNet++ dataset, focusing on two scenes (\texttt{8b5caf3398} and \texttt{b20a261fdf}). SplaTAM only reports results for the first 360 frames of \texttt{b20a261fdf}, and we include these results here for reference. In addition to evaluating the first 360 frames of \texttt{b20a261fdf}, we also report results on the full dataset to demonstrate our method's robustness in handling scenarios with large jumps in camera movement between consecutive frames. 

To account for the differing sequence lengths, the average performance is calculated using results for scene \texttt{8b5caf3398} combined with the corresponding setting (``360" or ``full") for \texttt{b20a261fdf}. Across both settings, our method consistently delivers strong results, achieving an average ATE of $1.56$ cm in the ``360" setting and $1.60$ cm in the ``full" setting. In contrast, SplaTAM's performance degrades significantly when using the full dataset, as it fails to maintain accurate camera tracking under large camera movements. These results underscore our method's ability to handle both limited-frame scenarios and more complex cases involving irregular camera motion, outperforming SplaTAM \cite{splatam}, ORB-SLAM3 \cite{orb-slam3}, and Point-SLAM \cite{point-slam} in both settings.

\begin{table}[htbp]
\centering
\resizebox{\linewidth}{!}{
\begin{tabular}{lccccccc}
\toprule
\textbf{Methods} & \textbf{0000} & \textbf{0059} & \textbf{0106} & \textbf{0169} & \textbf{0181} & \textbf{0207} & \textbf{Avg.} \\ 
\midrule
Vox-Fusion \cite{vox-fusion} & 68.84 &  24.18 &  \cellcolor{tabthird}8.41  &  27.28  & 23.30 &  9.41  &  26.90 \\
NICE-SLAM \cite{nice-slam} & \cellcolor{tabthird}12.00  & 14.00 &  \cellcolor{tabfirst}\textbf{7.90} & \cellcolor{tabfirst}\textbf{10.90} &   \cellcolor{tabthird}13.40 & \cellcolor{tabfirst}\textbf{6.20} & \cellcolor{tabsecond}10.73 \\
Point-SLAM \cite{point-slam} & \cellcolor{tabsecond}10.24  & \cellcolor{tabsecond}7.81 & 8.65 & 22.16 & 14.77 & 9.54 & 12.20 \\
SplaTAM \cite{splatam} & 12.83 & \cellcolor{tabthird}10.10 & 17.72 & \cellcolor{tabsecond}12.08 & \cellcolor{tabfirst}\textbf{11.10} &   \cellcolor{tabsecond}7.46 & \cellcolor{tabthird}11.88 \\
Ours & \cellcolor{tabfirst}\textbf{9.86} & \cellcolor{tabfirst}\textbf{7.80} & \cellcolor{tabsecond}8.02 & \cellcolor{tabthird}14.25 & \cellcolor{tabsecond}13.36 & \cellcolor{tabthird}8.71 & \cellcolor{tabfirst}\textbf{10.33} \\

\bottomrule
\end{tabular}
}
\caption{Tracking performance on the ScanNet dataset \cite{scannet}, reported using the Absolute Trajectory Error (ATE) metric (cm).}
\label{tab:ate_scannet}
\end{table}

\begin{table}[htbp]
\centering
\resizebox{\linewidth}{!}{
\begin{tabular}{lccccc}
\toprule
\textbf{Methods} & \textbf{8b5caf3398} & \textbf{b20a261fdf} & \textbf{b20a261fdf} & \multicolumn{2}{c}{\textbf{Avg.}} \\
& \textbf{full} & \textbf{360} & \textbf{full} & \textbf{360} & \textbf{full} \\
\midrule
Point-SLAM \cite{point-slam} & 296.7 & 390.8 & - & 343.75 & - \\
ORB-SLAM3 \cite{orb-slam3} & 156.8 & 159.7 & - & 158.25 & - \\
Splatam \cite{splatam} & \cellcolor{tabfirst}\textbf{0.62} & \cellcolor{tabsecond}1.94 & \cellcolor{tabsecond}16.09 & \cellcolor{tabfirst}\textbf{1.28} & \cellcolor{tabsecond}8.36 \\
Ours & \cellcolor{tabsecond}1.74 & \cellcolor{tabfirst}\textbf{1.38} & \cellcolor{tabfirst}\textbf{1.46} & \cellcolor{tabsecond}1.56 & \cellcolor{tabfirst}\textbf{1.60} \\
\bottomrule
\end{tabular}
}
\caption{Tracking performance on ScanNet++ \cite{scannet++}, measured by ATE [cm]. Baseline numbers are taken from \cite{splatam}. For SplaTAM, results on the full dataset for scene \texttt{b20a261fdf} were computed using their released code. ``-" indicates missing or unreported values.}
\label{tab:ate_scannet++}
\end{table}

\section{Tracking Time Analysis}

The results in Tables \ref{tab:tracktime_replica}, \ref{tab:tracktime_tum}, and \ref{tab:sparse_tracktime} highlight the efficiency and robustness of our method in reducing tracking time. On the Replica dataset (Table \ref{tab:tracktime_replica}), our method reduces the average tracking time by up to 56\% compared to MonoGS \cite{monogs} and maintains a consistent low tracking time across all scenes. On the TUM dataset (Table \ref{tab:tracktime_tum}), our approach achieves an average time reduction of 87\% compared to SplaTAM and an impressive 90\% reduction in the \texttt{fr3 office} scene.

In sparse settings (Table \ref{tab:sparse_tracktime}), our method outperforms SplaTAM \cite{splatam}, maintaining low and stable tracking times across varying frame strides while accurately tracking the camera pose. This demonstrates the scalability and reliability of our method in challenging scenarios, as further evidenced by its ability to maintain accuracy in pose tracking (see Tables \ref{tab:sparse_tracking_replica} and \ref{tab:sparse_tracking_tum}).

\begin{table}[htbp]
\centering
\resizebox{\linewidth}{!}{
\begin{tabular}{lccccccccc}
\toprule
\textbf{Methods} & \textbf{r0} & \textbf{r1} & \textbf{r2} & \textbf{o0} & \textbf{o1} & \textbf{o2} & \textbf{o3} & \textbf{o4} & \textbf{Avg.} \\ 
\midrule

SplaTAM \cite{splatam} & \cellcolor{tabthird}2.07 & \cellcolor{tabthird}2.03 & \cellcolor{tabthird}1.68 & \cellcolor{tabthird}1.87 & \cellcolor{tabthird}1.70 & \cellcolor{tabthird}1.36 & \cellcolor{tabthird}2.03 & \cellcolor{tabthird}2.29 & \cellcolor{tabthird}1.88 \\
MonoGS \cite{monogs} & \cellcolor{tabsecond}1.27 & \cellcolor{tabsecond}1.16 & \cellcolor{tabsecond}1.16 & \cellcolor{tabsecond}1.07 & \cellcolor{tabsecond}0.83 & \cellcolor{tabsecond}1.00 & \cellcolor{tabsecond}1.10 & \cellcolor{tabsecond}1.06 & \cellcolor{tabsecond}1.08 \\
\textbf{Ours} & \cellcolor{tabfirst}\textbf{0.54} & \cellcolor{tabfirst}\textbf{0.48} & \cellcolor{tabfirst}\textbf{0.52} & \cellcolor{tabfirst}\textbf{0.43} & \cellcolor{tabfirst}\textbf{0.42} & \cellcolor{tabfirst}\textbf{0.49} & \cellcolor{tabfirst}\textbf{0.54} & \cellcolor{tabfirst}\textbf{0.46} & \cellcolor{tabfirst}\textbf{0.48} \\

\bottomrule
\end{tabular}
}
\caption{Tracking time on the Replica dataset \cite{replica}, measured in seconds.}
\label{tab:tracktime_replica}
\end{table}

\begin{table}[htbp]
\centering
\scriptsize
\resizebox{\linewidth}{!}{
\begin{tabular}{lcccccc}
\toprule
\textbf{Methods} & \textbf{fr1/} & \textbf{fr1/} & \textbf{fr1/} & \textbf{fr2/} & \textbf{fr3/} & \textbf{Avg.} \\
& \textbf{desk} & \textbf{desk2} & \textbf{room} & \textbf{xyz} & \textbf{office} & \\
\midrule

SplaTAM \cite{splatam} & \cellcolor{tabthird}3.46 & \cellcolor{tabthird}3.02 & \cellcolor{tabthird}3.57 & \cellcolor{tabthird}3.85 & \cellcolor{tabthird}5.16 & \cellcolor{tabthird}3.81 \\
MonoGS \cite{monogs} & \cellcolor{tabsecond}0.89 & \cellcolor{tabsecond}0.93 & \cellcolor{tabsecond}0.89 & \cellcolor{tabsecond}0.59 & \cellcolor{tabsecond}0.81 & \cellcolor{tabsecond}0.82 \\
\textbf{Ours} & \cellcolor{tabfirst}\textbf{0.50} & \cellcolor{tabfirst}\textbf{0.48} & \cellcolor{tabfirst}\textbf{0.50} & \cellcolor{tabfirst}\textbf{0.51} & \cellcolor{tabfirst}\textbf{0.52} & \cellcolor{tabfirst}\textbf{0.50} \\
\bottomrule
\end{tabular}
}
\caption{Tracking time on the TUM dataset \cite{tum-rgbd}, measured in seconds.}
\label{tab:tracktime_tum}
\end{table}

\begin{table}[htbp]
\centering
\scriptsize
\setlength{\tabcolsep}{5pt}
\begin{tabular}{lccc}
\toprule
\textbf{Methods} & \textbf{Stride} & \textbf{Replica} & \textbf{TUM} \\
\midrule
\multirow{3}{*}{ SplaTAM \cite{splatam}} & 10 & \cellcolor{tabsecond}1.77 & \cellcolor{tabsecond}6.09 \\
        & 20 & \cellcolor{tabsecond}2.51 & \cellcolor{tabsecond}5.35 \\
        & 40 & \cellcolor{tabsecond}1.88 & \cellcolor{tabsecond}3.09 \\
\midrule
\multirow{3}{*}{\textbf{Ours}} & 10 & \cellcolor{tabfirst}\textbf{0.45} & \cellcolor{tabfirst}\textbf{0.51} \\
        & 20 & \cellcolor{tabfirst}\textbf{0.49} & \cellcolor{tabfirst}\textbf{0.51} \\
        & 40 & \cellcolor{tabfirst}\textbf{0.51} & \cellcolor{tabfirst}\textbf{0.50} \\

\bottomrule
\end{tabular}
\caption{Comparison of average tracking time of all scenes in \textbf{\textit{sparse settings}} on the Replica \cite{replica} and TUM \cite{tum-rgbd} datasets.}
\label{tab:sparse_tracktime}
\end{table}

\begin{figure}[!htp]
    \centering
    \includegraphics[width=\linewidth]{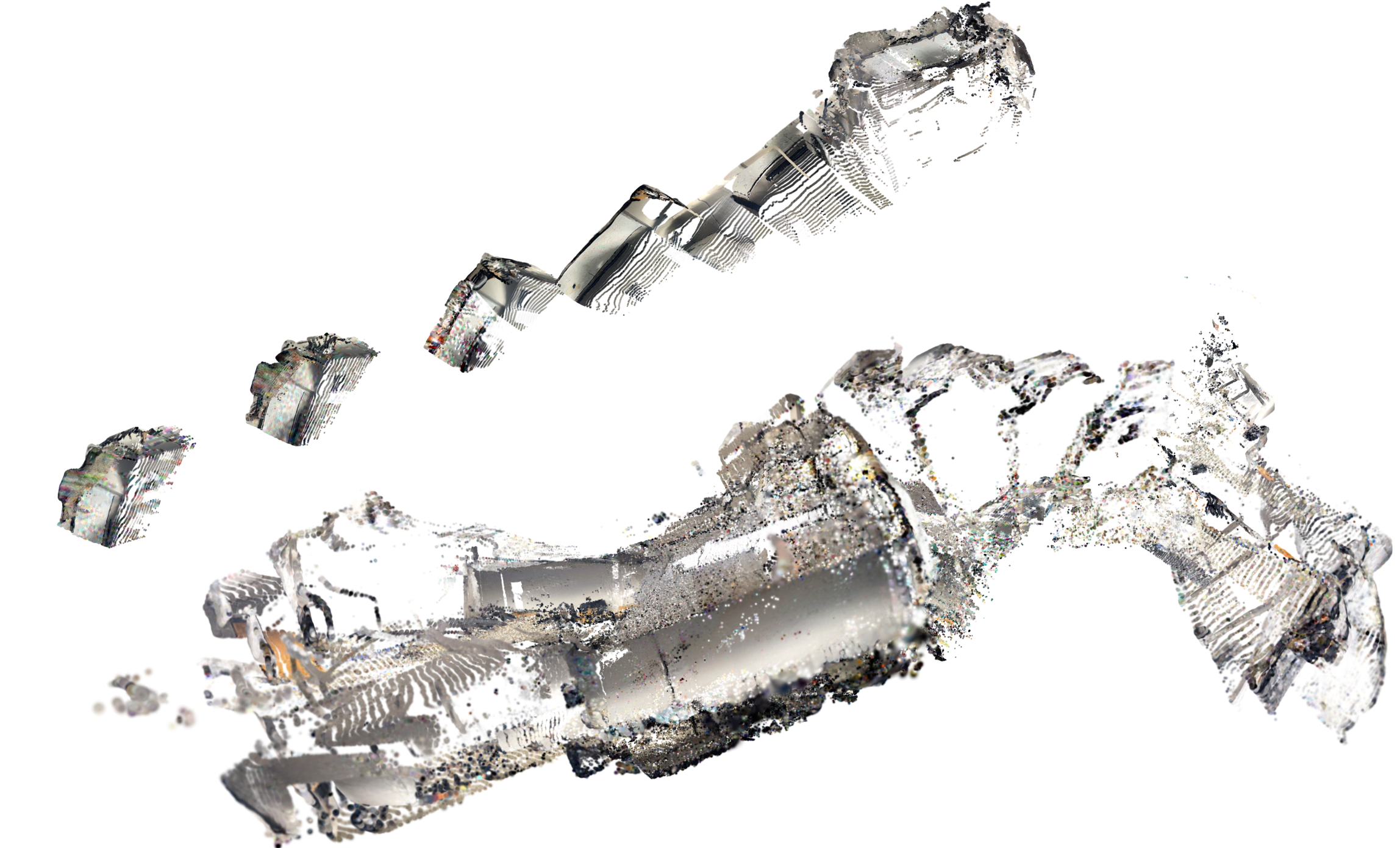}
    \caption{Example of a failed reconstruction by SplaTAM \cite{splatam} on our custom dataset, showing severe distortion and misaligned geometry due to tracking failures.}
    \label{fig:failed_splatam}
\end{figure}

\section{Further Implementation Details}

All experiments are conducted on an NVIDIA A100 80GB PCIe GPU to ensure quick and efficient experimentation. However, our method is also compatible with a standard desktop setup equipped with an Intel Core i7-12700K processor and an NVIDIA GeForce RTX 3050 GPU.

For the final reconstruction, we perform $30,000$ refinement iterations, which add approximately 3-8 minutes of computation time depending on the number of Gaussians in the scene. The tracking process involves 30-70 steps, while the mapping process requires 100-150 steps, ensuring robust and accurate scene reconstruction.

To facilitate further research and reproducibility, we will release the code along with all configuration details.

\section{Rendering Comparisons}

\begin{figure*}[h]
\centering
\setlength{\tabcolsep}{1pt}
\newcommand{\sz}{0.245}
\resizebox{\linewidth}{!}{
\begin{tabular}{c c c c}

\textbf{SplaTAM \cite{splatam}} & \textbf{MonoGS \cite{monogs}} & \textbf{Ours} & \textbf{GT} \\

\includegraphics[width=\sz\textwidth]{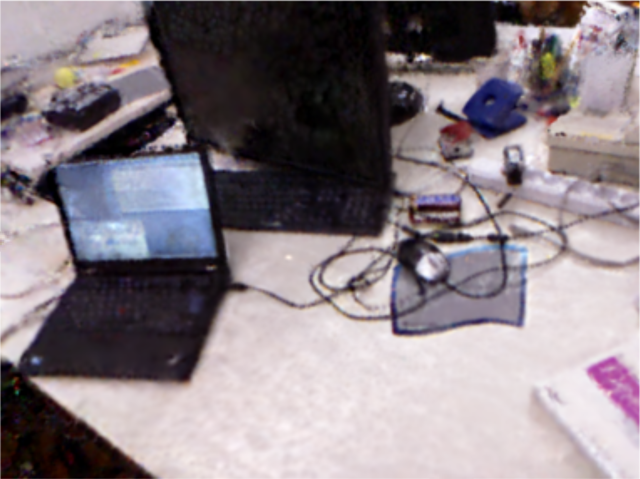} &
\includegraphics[width=\sz\textwidth]{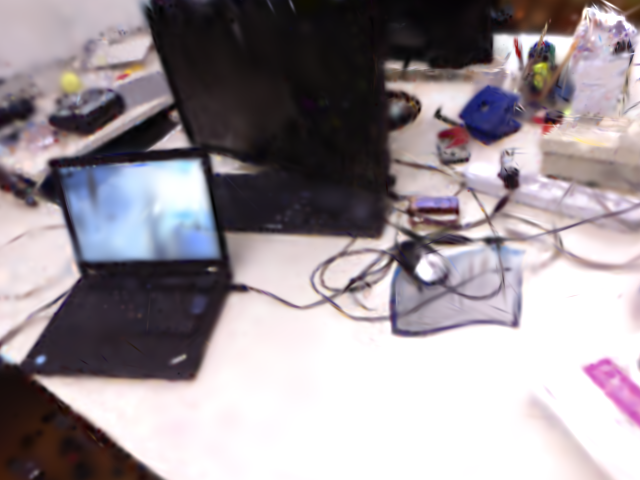} &
\includegraphics[width=\sz\textwidth]{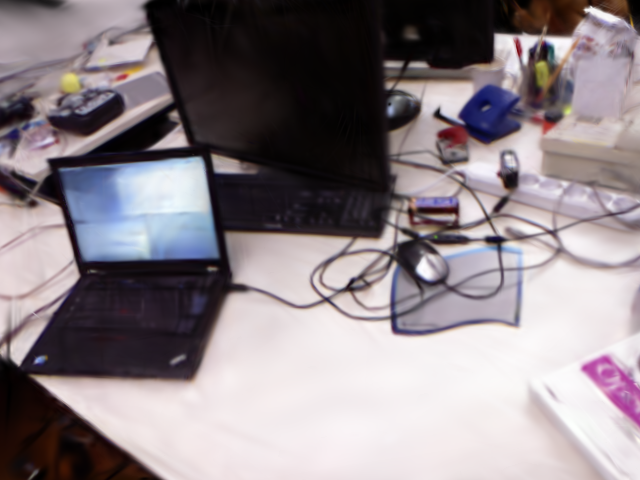} &
\includegraphics[width=\sz\textwidth]{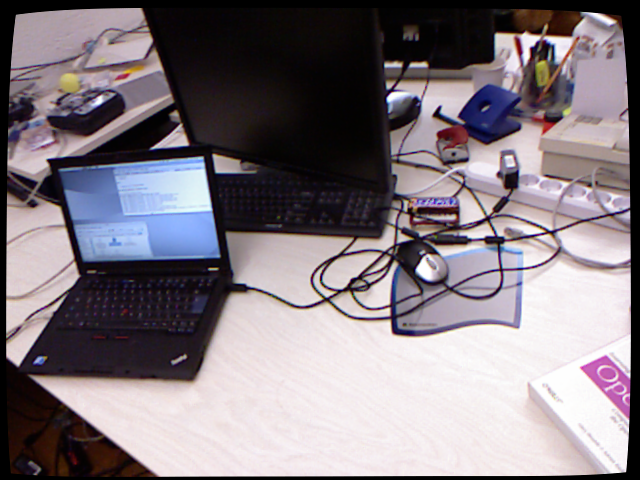} \\
[-2pt]
\includegraphics[width=\sz\textwidth]{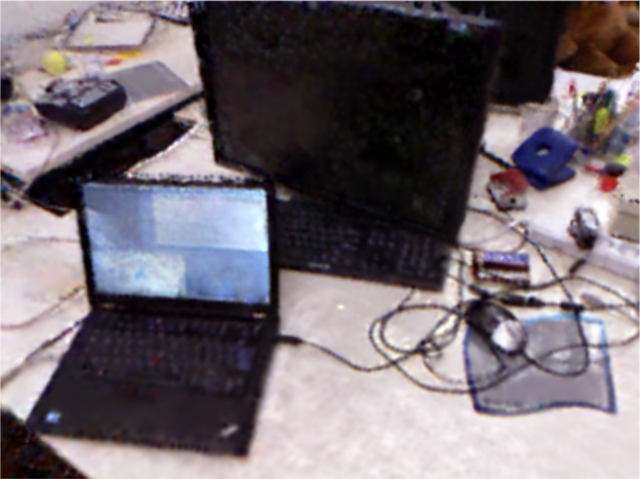} &
\includegraphics[width=\sz\textwidth]{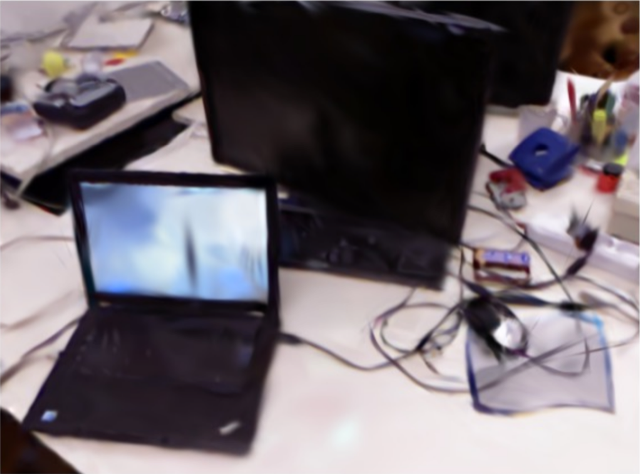} &
\includegraphics[width=\sz\textwidth]{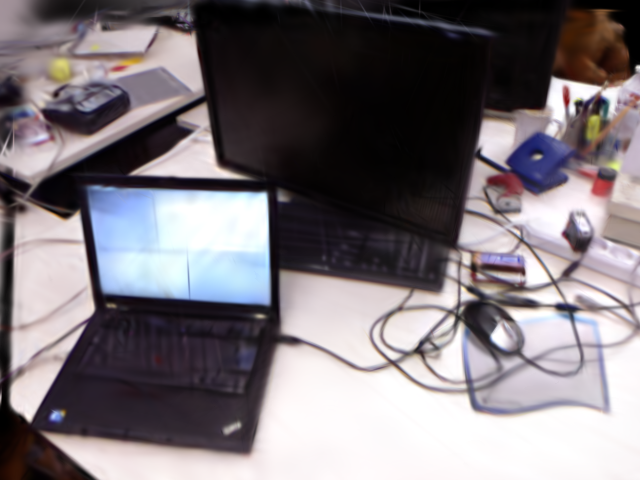} &
\includegraphics[width=\sz\textwidth]{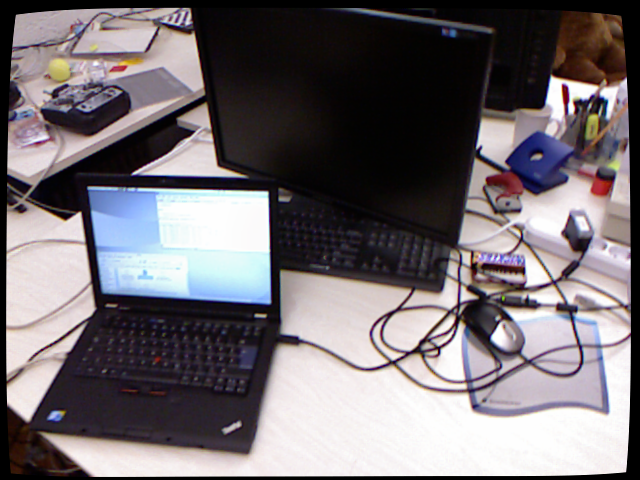} \\
[-2pt]
\includegraphics[width=\sz\textwidth]{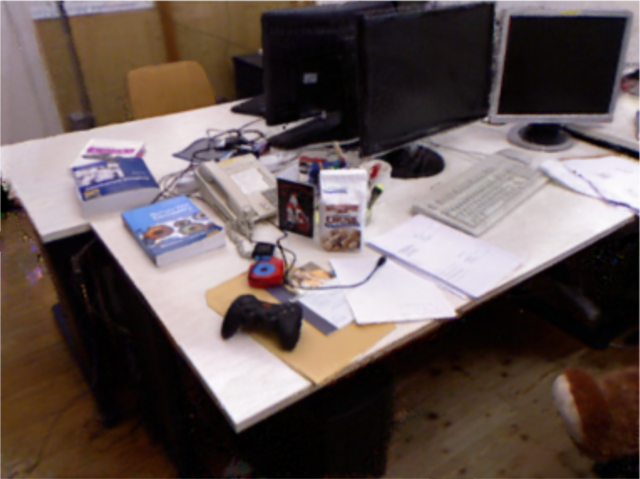} &
\includegraphics[width=\sz\textwidth]{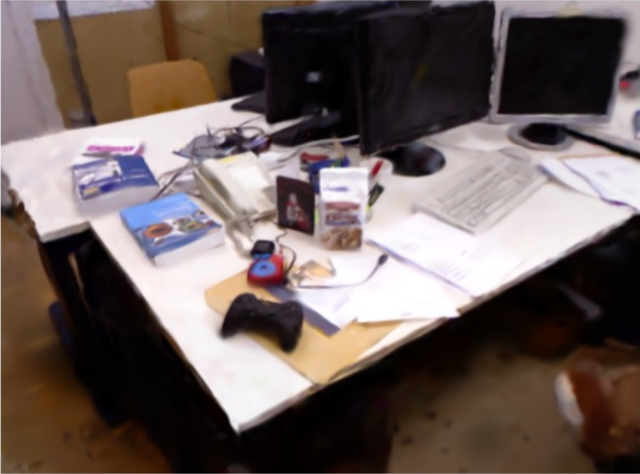} &
\includegraphics[width=\sz\textwidth]{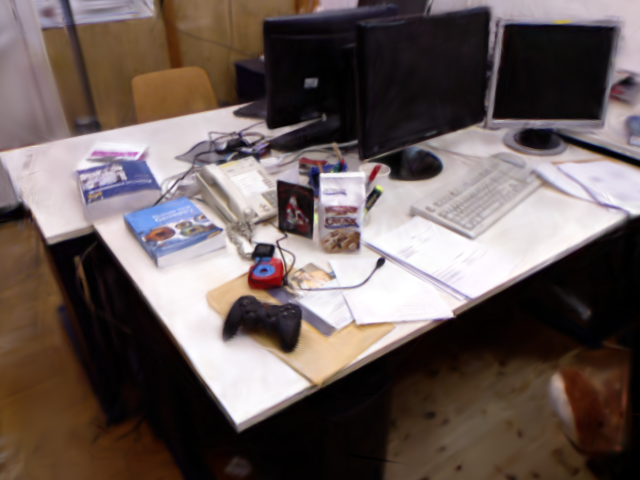} &
\includegraphics[width=\sz\textwidth]{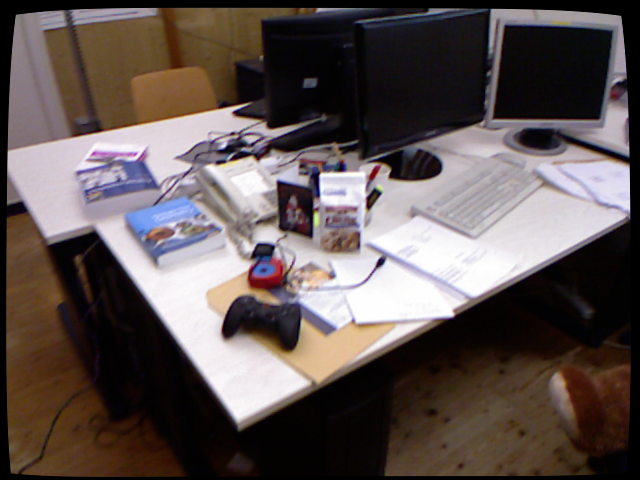} \\
[-2pt]

\end{tabular}
}
\caption{Rendering comparison on TUM \texttt{fr1/desk}.}
\label{fig:rendering_comparison_desk}
\end{figure*}

\begin{figure*}[h]
\centering
\setlength{\tabcolsep}{1pt}
\newcommand{\sz}{0.245}
\resizebox{\linewidth}{!}{
\begin{tabular}{c c c c}

\textbf{SplaTAM \cite{splatam}} & \textbf{MonoGS \cite{monogs}} & \textbf{Ours} & \textbf{GT} \\

\includegraphics[width=\sz\textwidth]{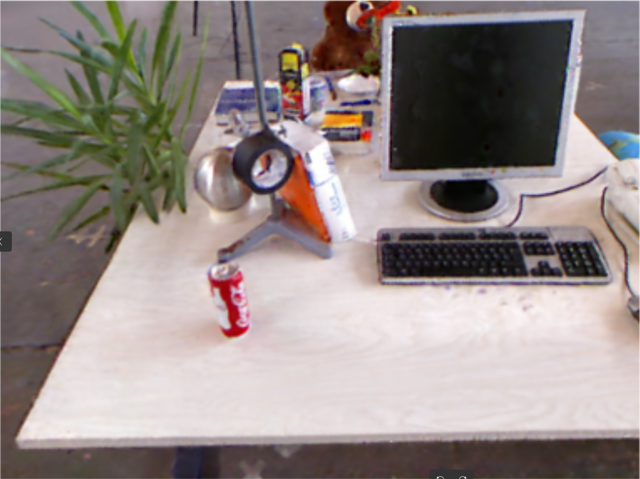} &
\includegraphics[width=\sz\textwidth]{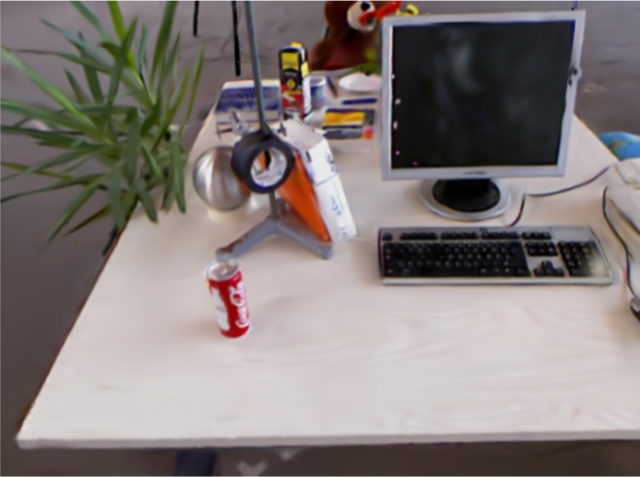} &
\includegraphics[width=\sz\textwidth]{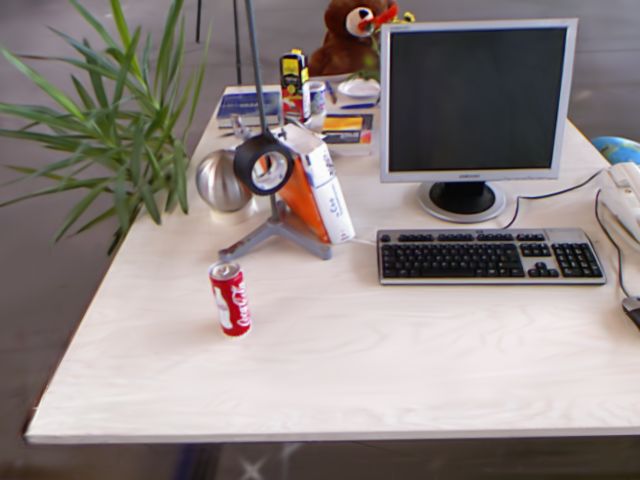} &
\includegraphics[width=\sz\textwidth]{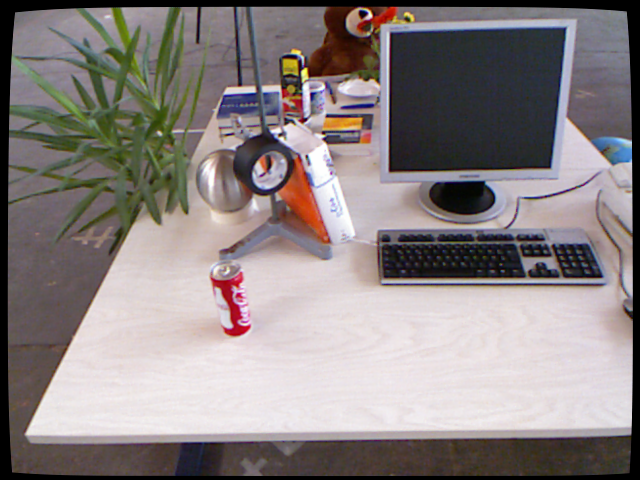} \\
[-2pt]
\includegraphics[width=\sz\textwidth]{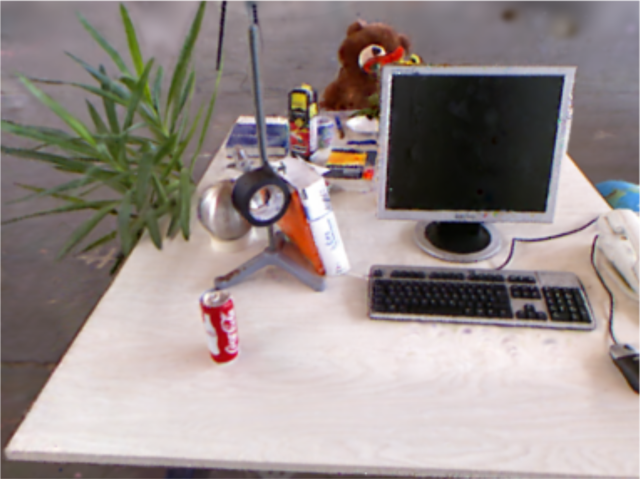} &
\includegraphics[width=\sz\textwidth]{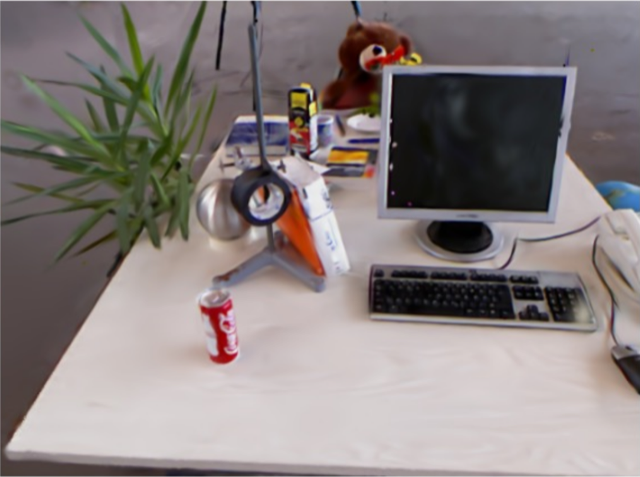} &
\includegraphics[width=\sz\textwidth]{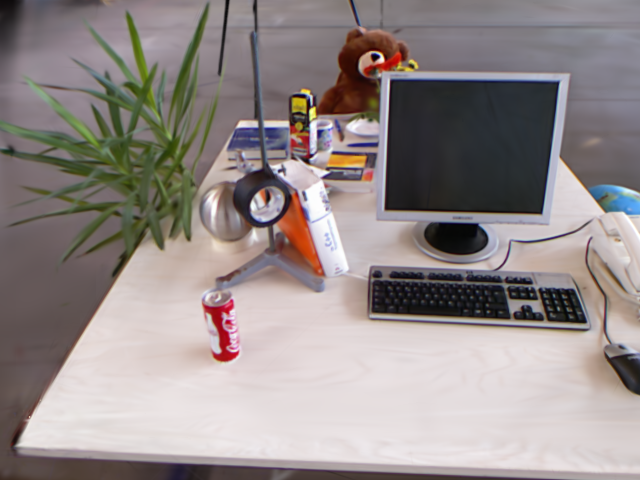} &
\includegraphics[width=\sz\textwidth]{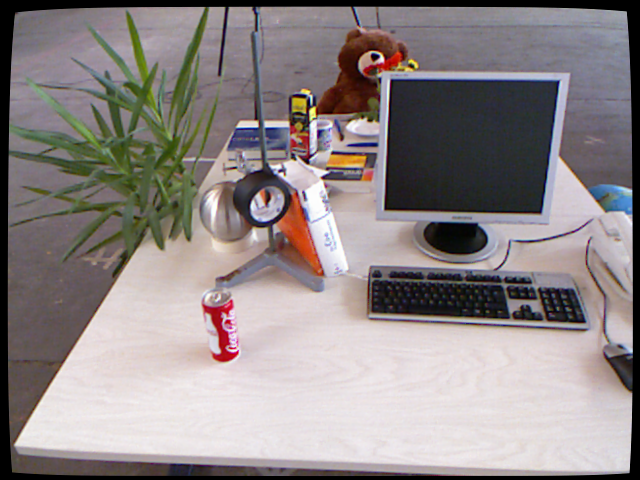} \\
[-2pt]
\includegraphics[width=\sz\textwidth]{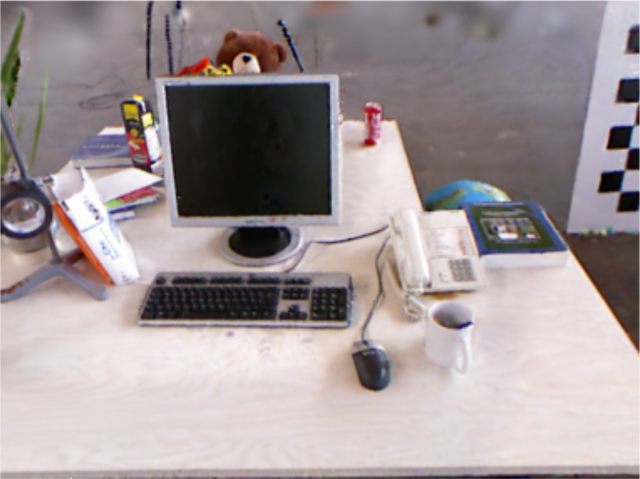} &
\includegraphics[width=\sz\textwidth]{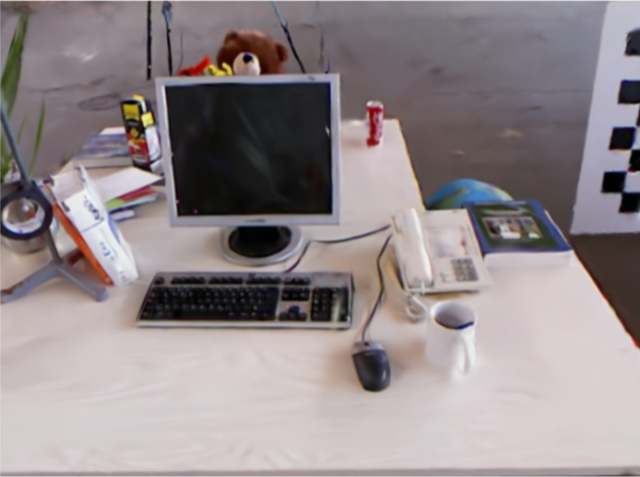} &
\includegraphics[width=\sz\textwidth]{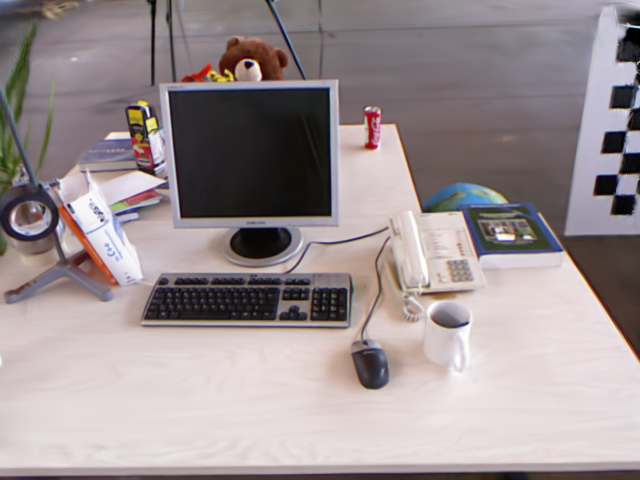} &
\includegraphics[width=\sz\textwidth]{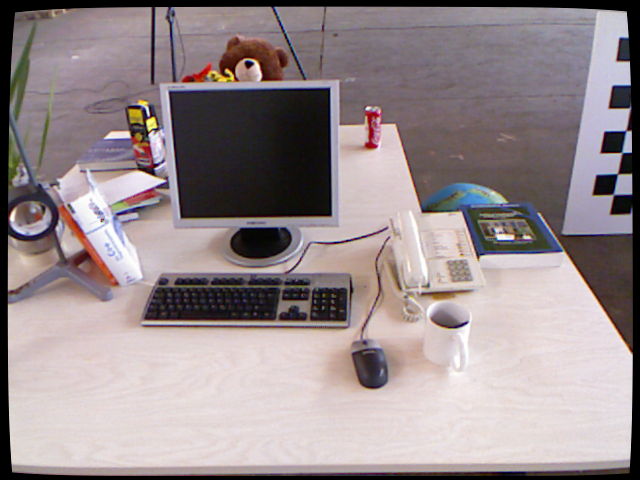} \\
[-2pt]

\end{tabular}
}
\caption{Rendering comparison on TUM \texttt{fr1/xyz}.}
\label{fig:rendering_comparison_xyz}
\end{figure*}

\begin{figure*}[h]
\centering
\setlength{\tabcolsep}{1pt}
\newcommand{\sz}{0.245}
\resizebox{\linewidth}{!}{
\begin{tabular}{c c c c}

\textbf{SplaTAM \cite{splatam}} & \textbf{MonoGS \cite{monogs}} & \textbf{Ours} & \textbf{GT} \\

\includegraphics[width=\sz\textwidth]{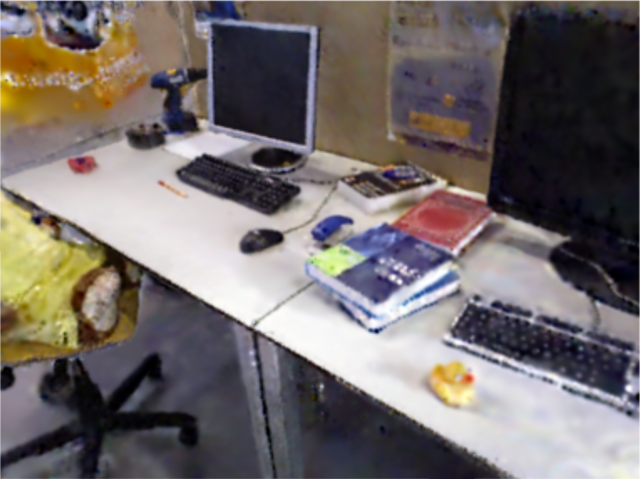} &
\includegraphics[width=\sz\textwidth]{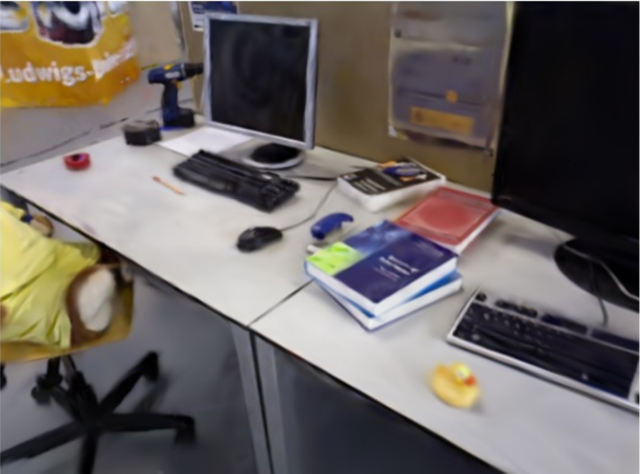} &
\includegraphics[width=\sz\textwidth]{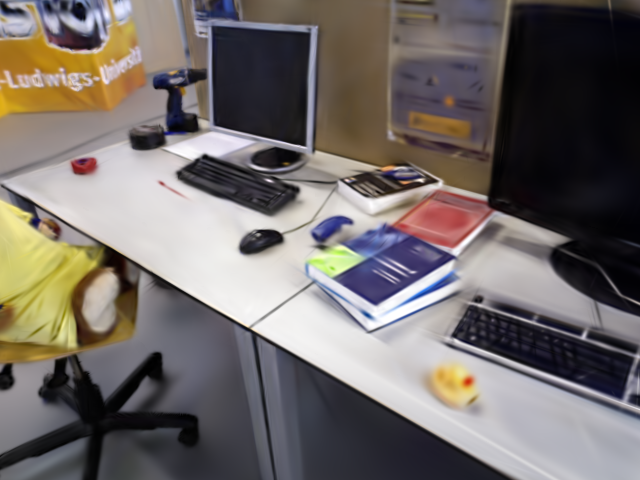} &
\includegraphics[width=\sz\textwidth]{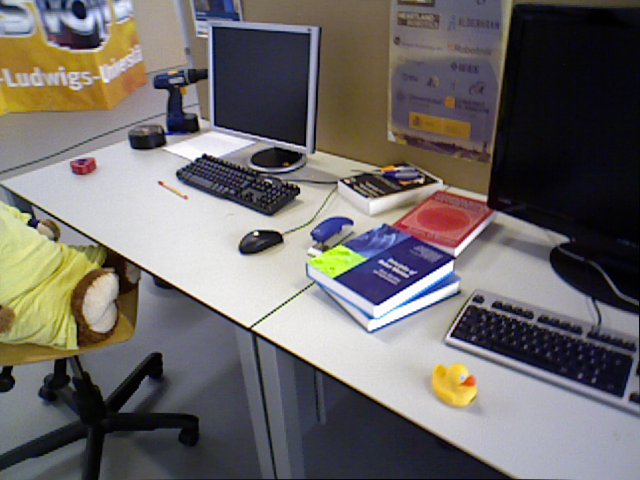} \\
[-2pt]
\includegraphics[width=\sz\textwidth]{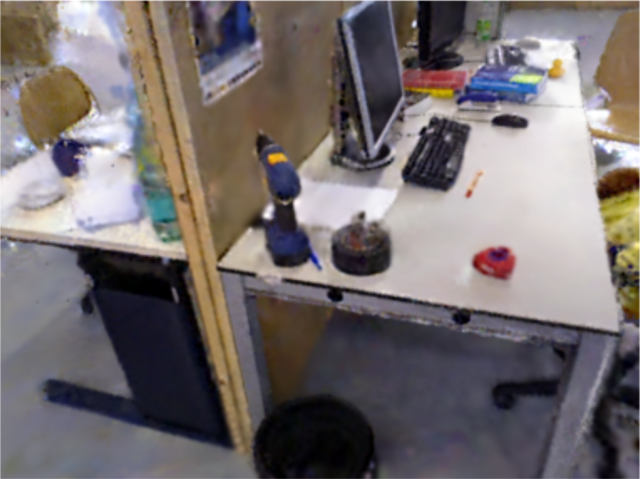} &
\includegraphics[width=\sz\textwidth]{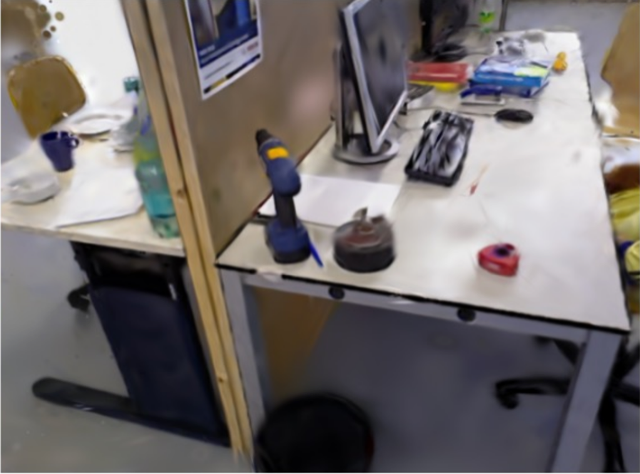} &
\includegraphics[width=\sz\textwidth]{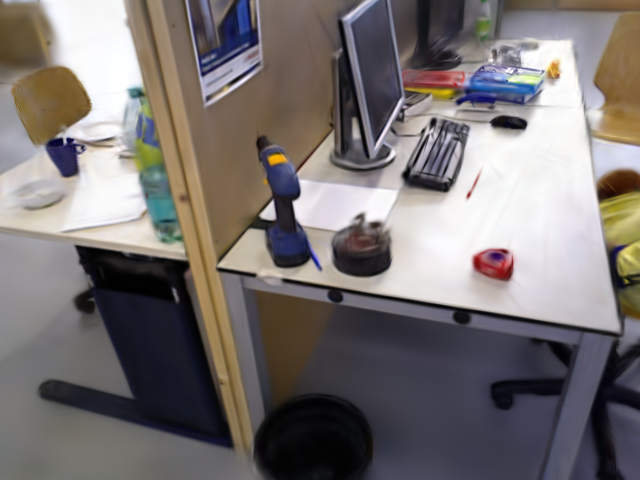} &
\includegraphics[width=\sz\textwidth]{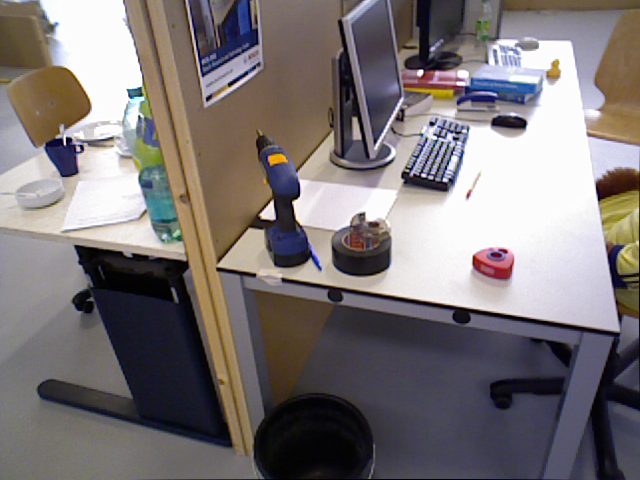} \\
[-2pt]
\includegraphics[width=\sz\textwidth]{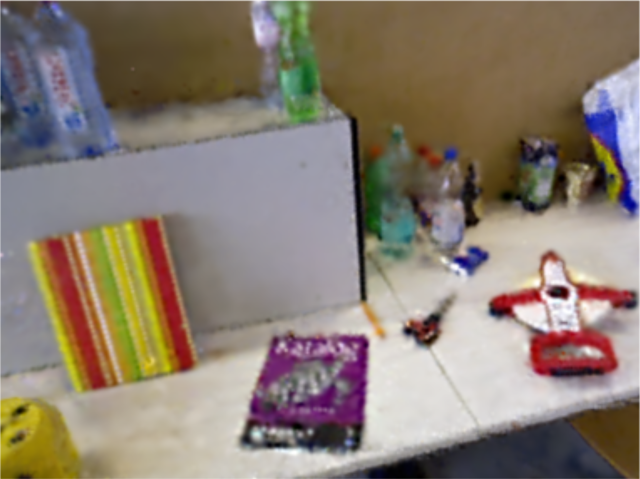} &
\includegraphics[width=\sz\textwidth]{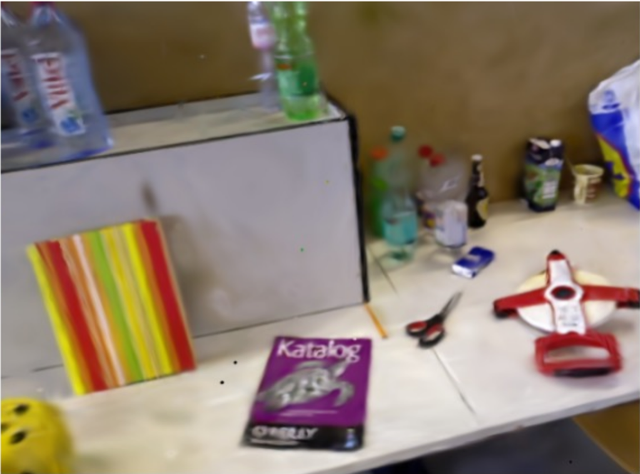} &
\includegraphics[width=\sz\textwidth]{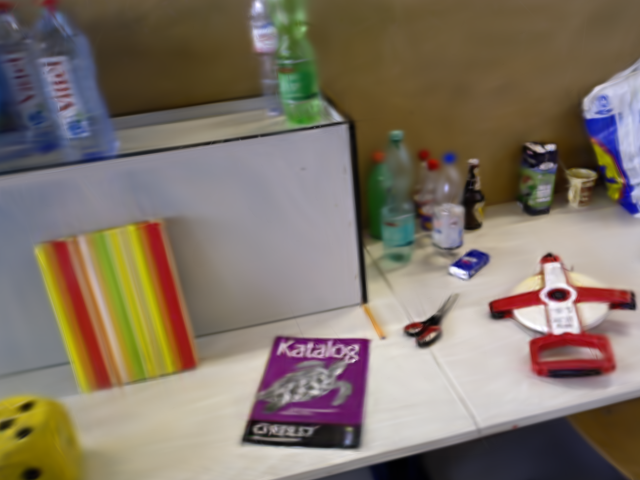} &
\includegraphics[width=\sz\textwidth]{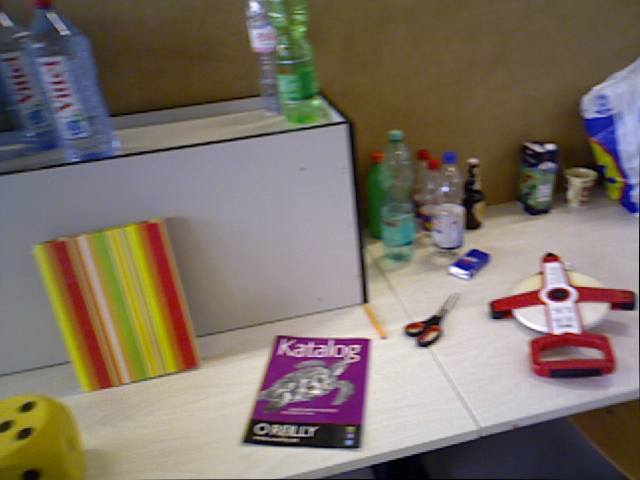} \\
[-2pt]

\end{tabular}
}
\caption{Rendering comparison on TUM \texttt{fr1/office}.}
\label{fig:rendering_comparison_office}
\end{figure*}

The additional visual results presented in Figures \ref{fig:rendering_comparison_desk}, \ref{fig:rendering_comparison_xyz}, and \ref{fig:rendering_comparison_office} illustrate the rendering quality of different methods on the TUM dataset, specifically on the \texttt{fr1/desk}, \texttt{fr1/xyz}, and \texttt{fr1/office} scenes. Each figure compares the outputs of SplaTAM \cite{splatam}, MonoGS \cite{monogs}, and our method against the ground truth (GT).

Our method consistently delivers sharper and more accurate reconstructions across all scenes. In Figure \ref{fig:rendering_comparison_desk} (\texttt{fr1/desk}), our approach captures fine details of objects such as the cables, laptop, and monitor, which are blurred or distorted in the outputs of SplaTAM and MonoGS. Similarly, in Figure \ref{fig:rendering_comparison_xyz} (\texttt{fr1/xyz}), our method preserves object boundaries and textures, such as the red soda can, the keyboard and desktop monitor, closely matching the GT. Finally, in Figure \ref{fig:rendering_comparison_office} (\texttt{fr1/office}), our method demonstrates higher detail retention, as seen in the books, objects on the desk, and smaller items in the scene, where other methods show noticeable blurring or inaccuracies.

\section{Evaluation on Self-Captured Dataset}  

We evaluate our method on a self-captured dataset consisting of 296 images taken with the depth camera of an iPhone 13 Pro Max. The images were captured individually rather than extracted from a video, making the dataset well-suited for sparse settings where the overlap between consecutive frames is limited. Notably, the depth measurements from the iPhone 13 Pro Max are only reliable within a range of up to 3.5 meters, as determined empirically. This limitation introduces significant challenges for reconstruction due to inherent depth noise in the sensor.

Table \ref{tab:custom_results} presents the performance of different methods on this custom dataset. Since both SplaTAM and MonoGS fail to track camera poses, their greyed-out results are invalid and included solely for reference. Figure \ref{fig:failed_splatam} provides an example of a failed reconstruction by SplaTAM, where the scene is heavily distorted, and the geometry is misaligned, emphasizing the challenges caused by depth noise and tracking failures. In contrast, our method successfully tracks the camera poses and delivers significantly better reconstruction metrics.

\begin{table}[htbp]
\centering
\scriptsize
\setlength{\tabcolsep}{5.5pt}
\begin{tabular}{ccccccc}
\toprule
\textbf{Methods} & \textbf{Success} & \textbf{ATE (cm)} \perflower & \textbf{PSNR} \higher & \textbf{SSIM} \higher & \textbf{LPIPS} \perflower \\
\midrule
SplaTAM & \redx & 1207 & \textcolor{gray}{16.15} & \textcolor{gray}{0.572} & \textcolor{gray}{0.622} \\
MonoGS  & \redx & 1970 & \textcolor{gray}{13.61} & \textcolor{gray}{0.584} & \textcolor{gray}{0.634} \\
Ours    & \greencheck & \cellcolor{tabfirst}\textbf{19.59} & \cellcolor{tabfirst}\textbf{23.73} & \cellcolor{tabfirst}\textbf{0.781} & \cellcolor{tabfirst}\textbf{0.277} \\

\bottomrule
\end{tabular}
\caption{}
\label{tab:custom_results}
\end{table}

\begin{figure*}[h]
\centering
\setlength{\tabcolsep}{1pt}
\newcommand{\sz}{0.245}
\resizebox{\linewidth}{!}{
\begin{tabular}{c c c c}

\textbf{SplaTAM \cite{splatam}} & \textbf{MonoGS \cite{monogs}} & \textbf{Ours} & \textbf{GT} \\

\includegraphics[width=\sz\textwidth]{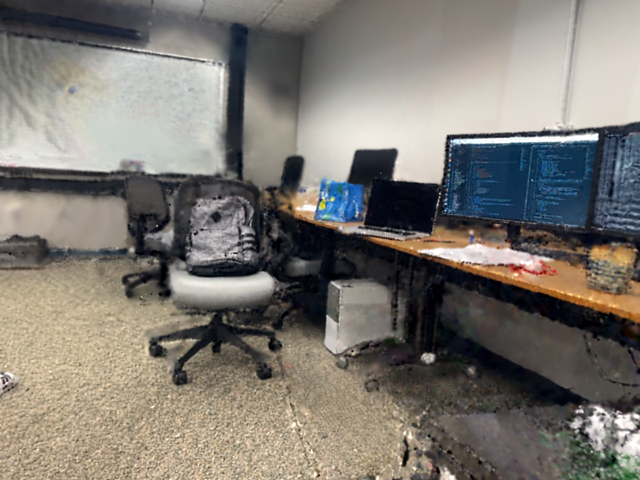} &
\includegraphics[width=\sz\textwidth]{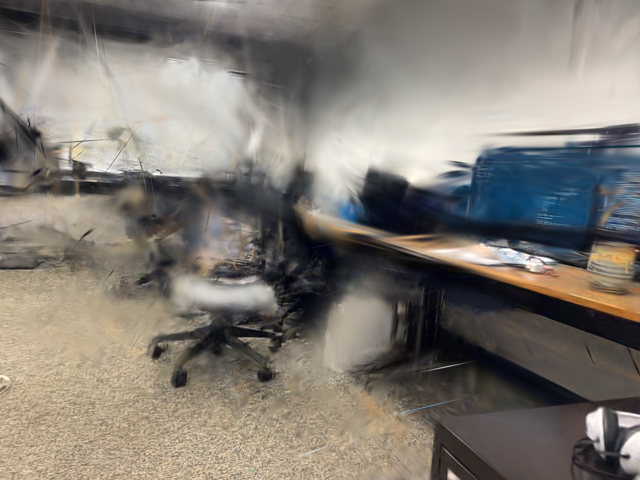} &
\includegraphics[width=\sz\textwidth]{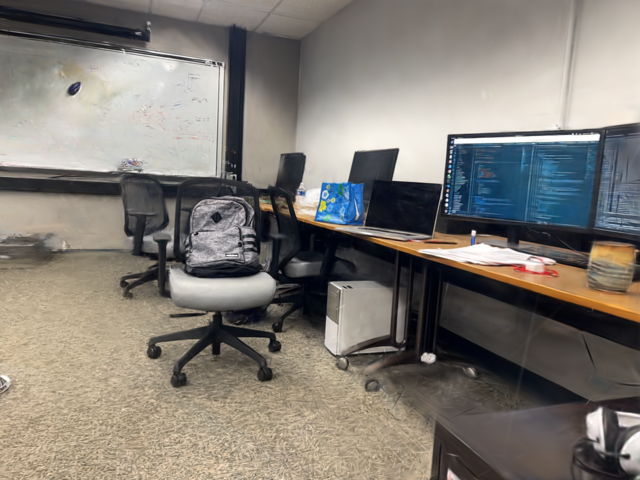} &
\includegraphics[width=\sz\textwidth]{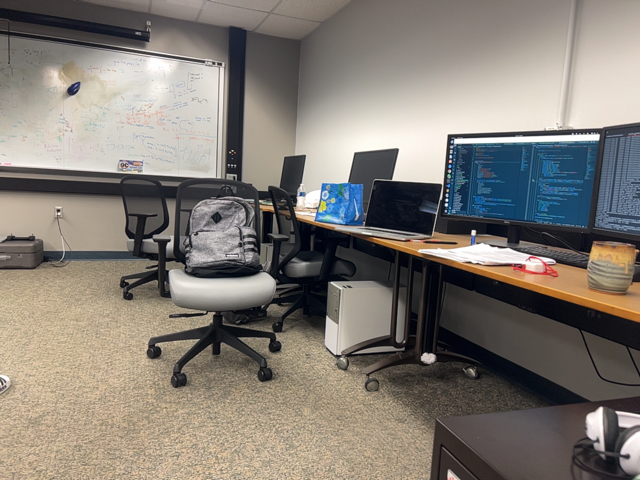} \\
[-2pt]
\includegraphics[width=\sz\textwidth]{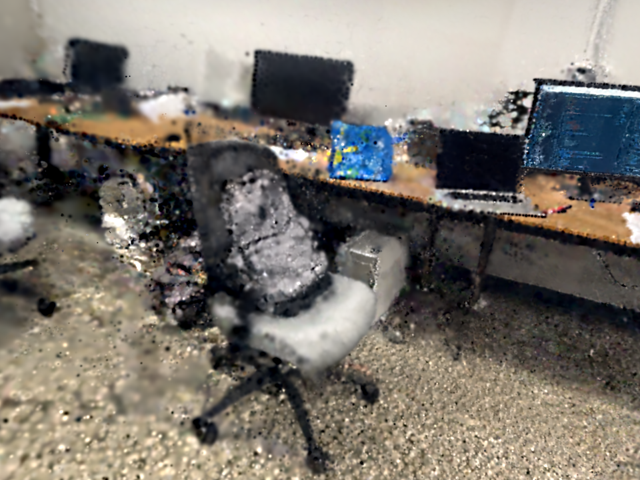} &
\includegraphics[width=\sz\textwidth]{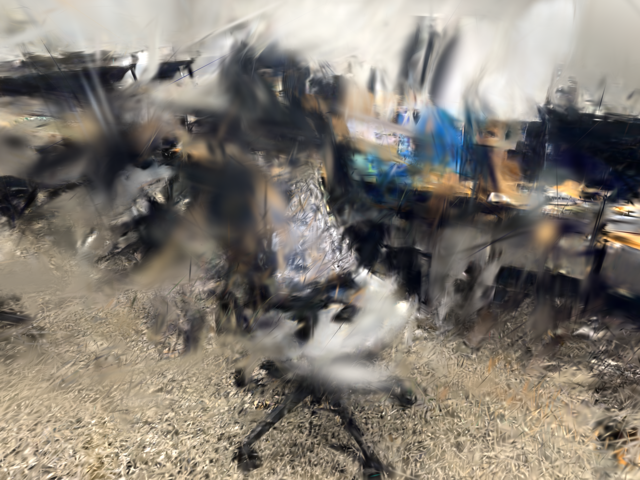} &
\includegraphics[width=\sz\textwidth]{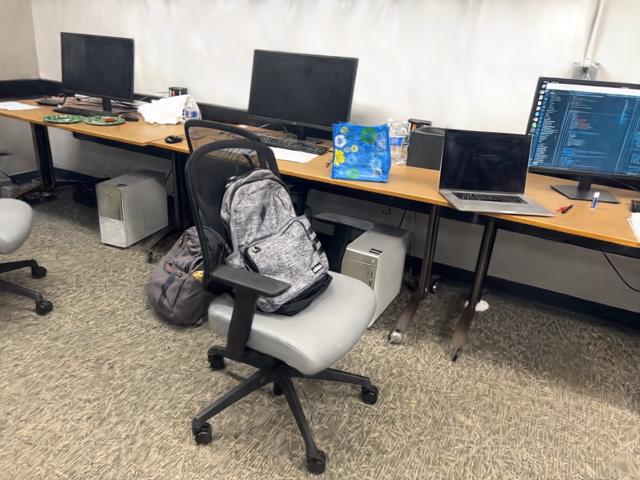} &
\includegraphics[width=\sz\textwidth]{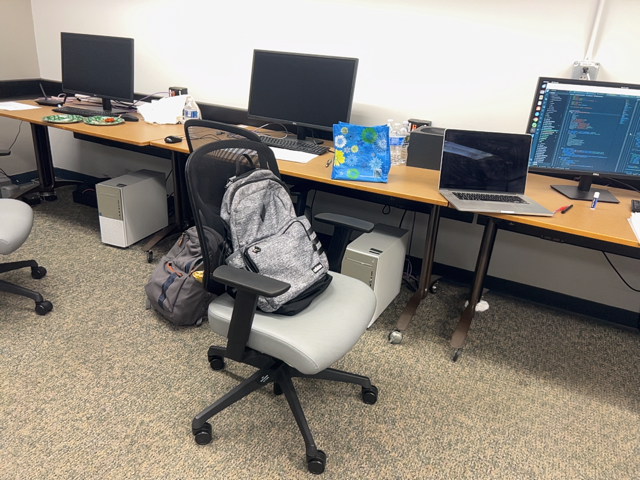} \\
[-2pt]
\includegraphics[width=\sz\textwidth]{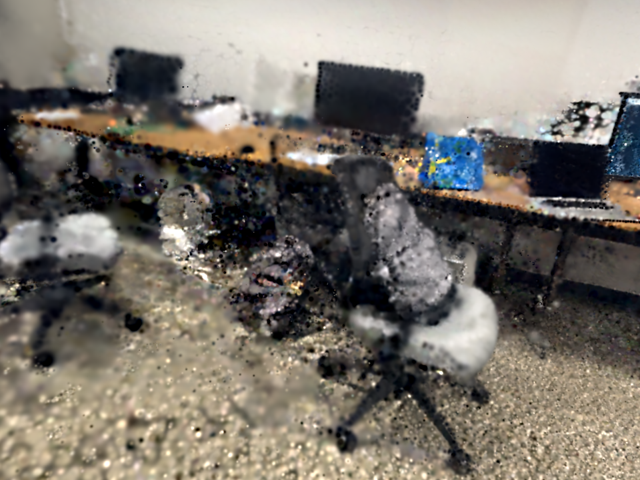} &
\includegraphics[width=\sz\textwidth]{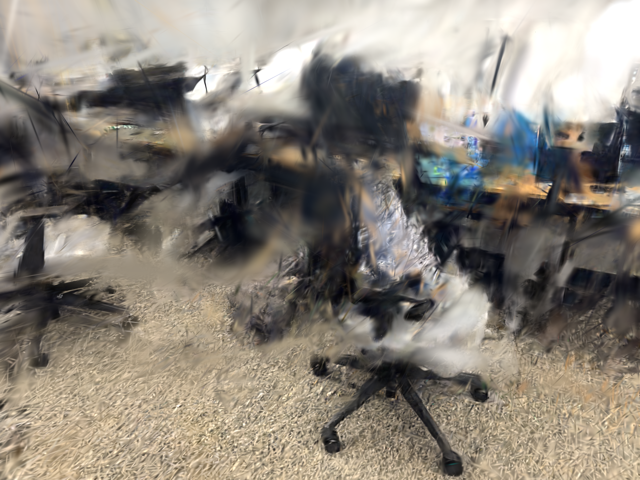} &
\includegraphics[width=\sz\textwidth]{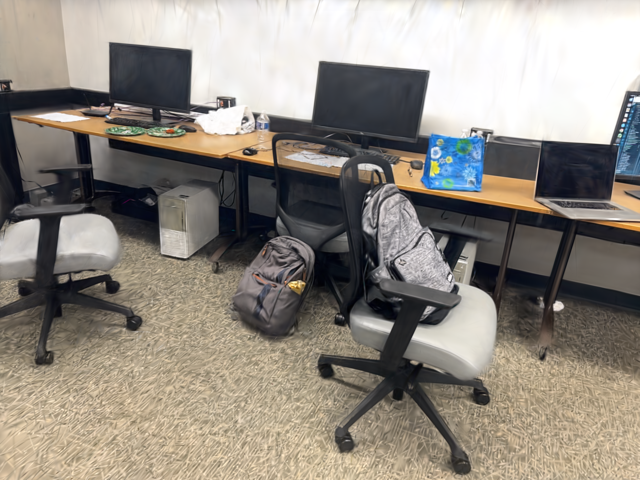} &
\includegraphics[width=\sz\textwidth]{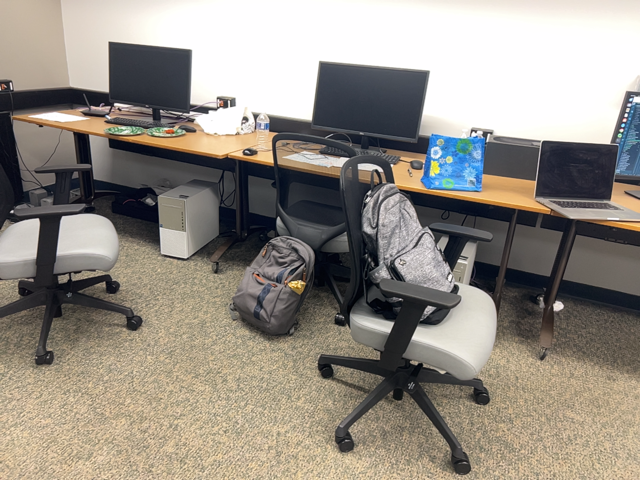} \\
[-2pt]
\includegraphics[width=\sz\textwidth]{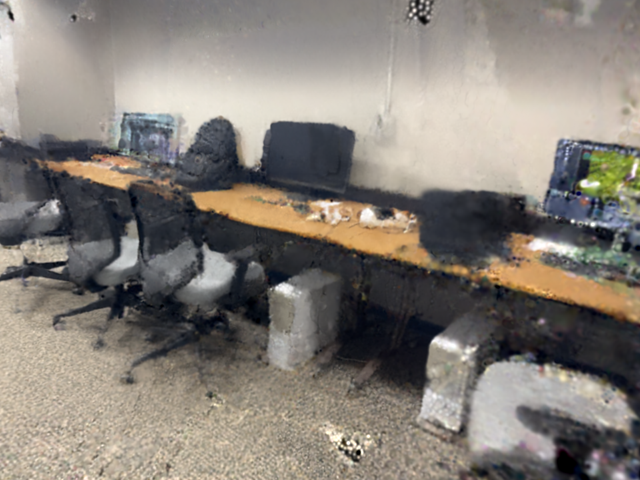} &
\includegraphics[width=\sz\textwidth]{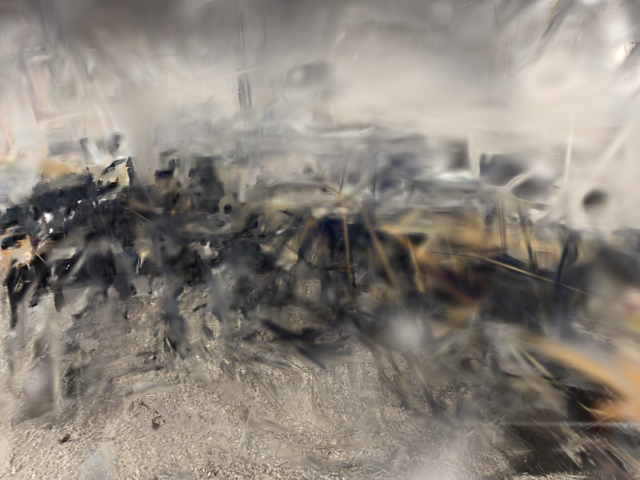} &
\includegraphics[width=\sz\textwidth]{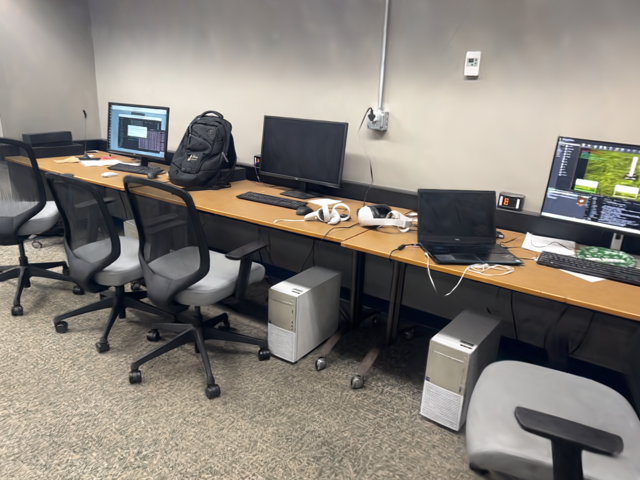} &
\includegraphics[width=\sz\textwidth]{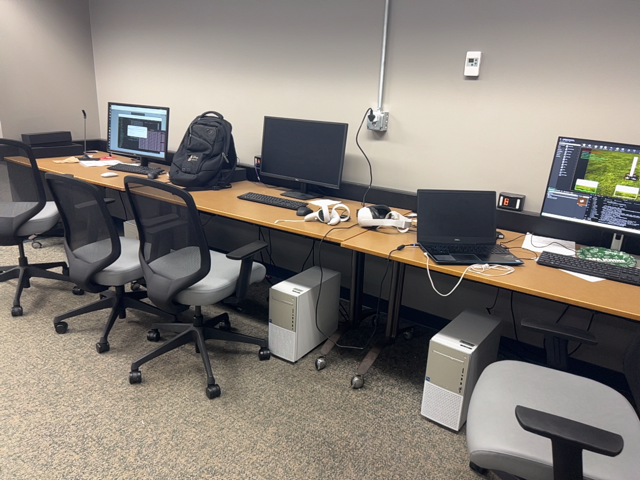} \\
[-2pt]

\includegraphics[width=\sz\textwidth]{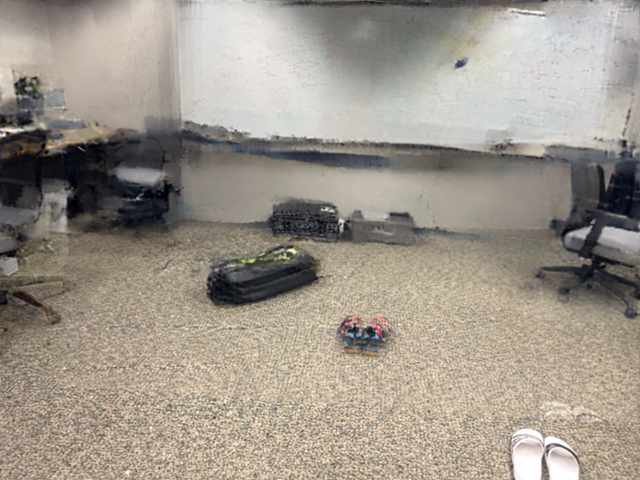} &
\includegraphics[width=\sz\textwidth]{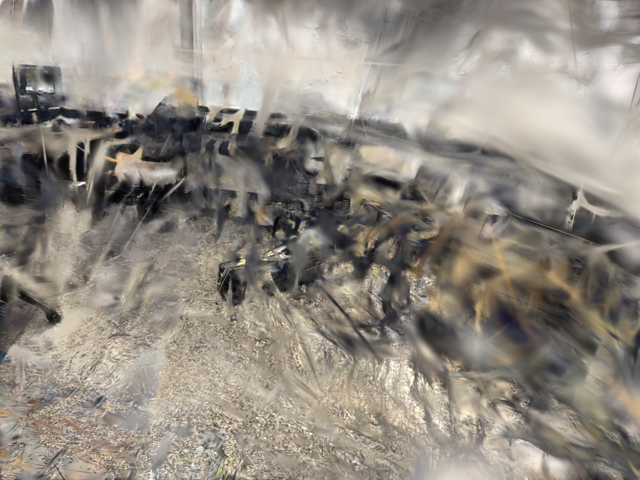} &
\includegraphics[width=\sz\textwidth]{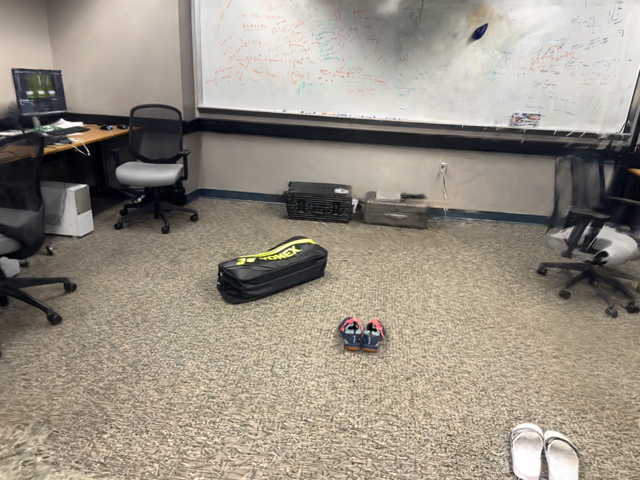} &
\includegraphics[width=\sz\textwidth]{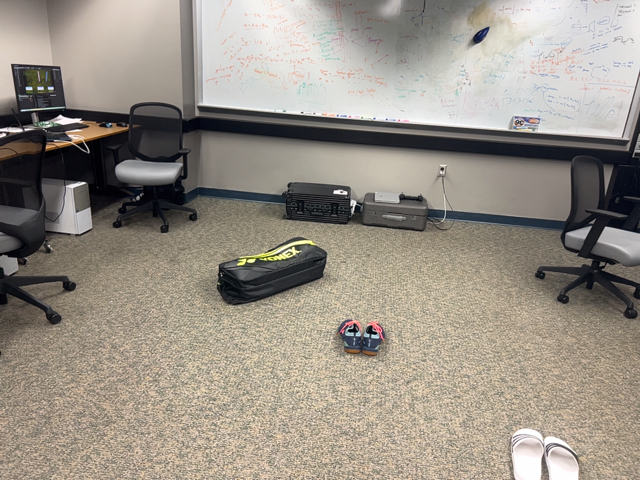} \\
[-2pt]
\includegraphics[width=\sz\textwidth]{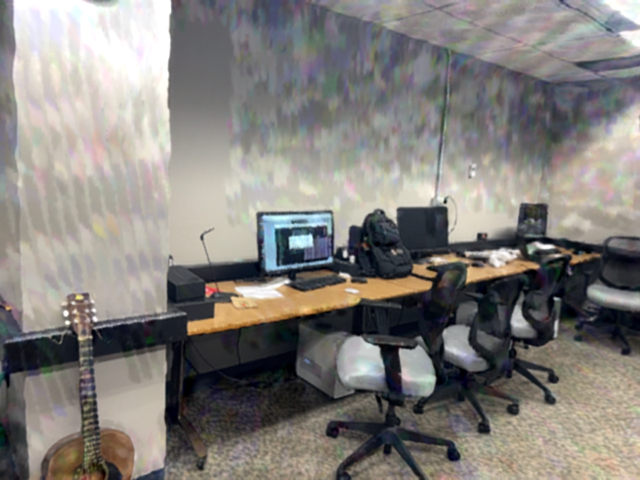} &
\includegraphics[width=\sz\textwidth]{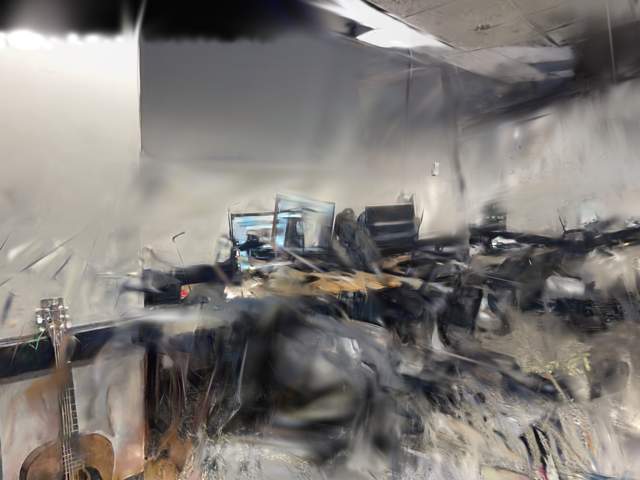} &
\includegraphics[width=\sz\textwidth]{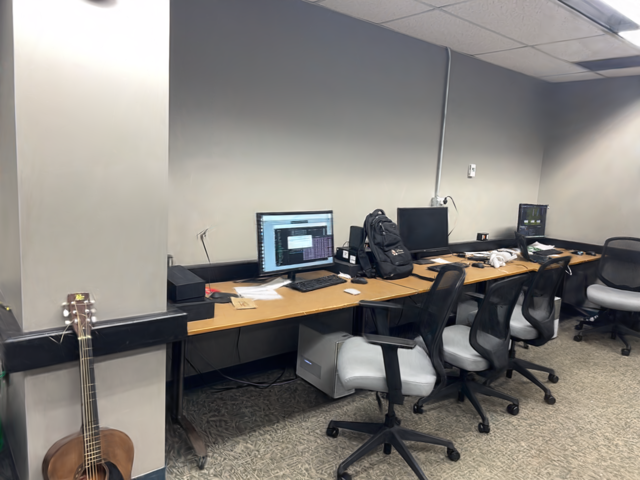} &
\includegraphics[width=\sz\textwidth]{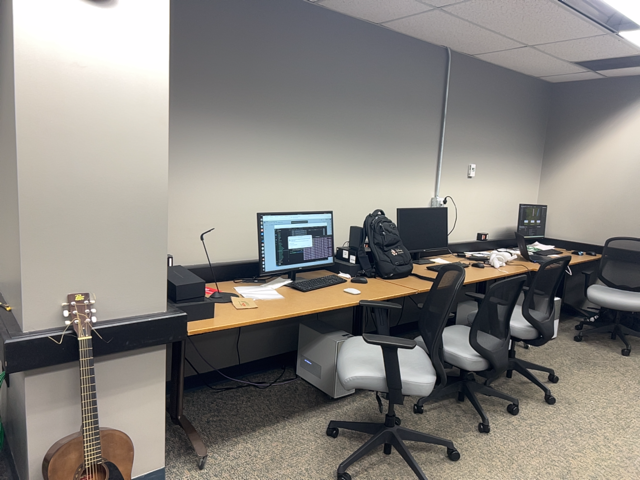} \\
% [-2pt]

\end{tabular}
}
\caption{Rendering comparison on a self-captured dataset captured with an iPhone camera.}
\label{fig:rendering_custom}
\end{figure*}

Figure \ref{fig:rendering_custom} highlights a rendering comparison on the self-captured dataset. Although SplaTAM and MonoGS produce rendered images that appear visually reasonable, their inability to accurately track the camera poses leads to completely incorrect scene reconstructions. Our method, however, achieves accurate camera pose tracking, resulting in a reconstruction that closely aligns with the ground truth. For more visual results and videos of the reconstructed scenes, please visit our project page at \href{https://flashslam.github.io}{flashslam.github.io}.

\end{document}